\RequirePackage{etex}
\documentclass{CUP-JNL-NLP}
\usepackage{amsmath}
\usepackage{graphicx}
\usepackage{natbib}
\usepackage{url}
\usepackage{linguex}

\begin{document}
\label{firstpage}

\lefttitle{Multilingual Web Register Identification}
\righttitle{}

\papertitle{Article}

\jnlPage{\pageref{firstpage}}{\pageref{lastpage}}
\jnlDoiYr{}
\doival{}

\begin{authgrp}
\author{Erik Henriksson}
\affiliation{University of Turku}
\author{Amanda Myntti}
\author{Saara Hellström}
\author{Anni Eskelinen}
\author{Selcen Erten-Johansson}
\author{ Veronika Laippala}
\end{authgrp}

\begin{abstract}
This article presents multilingual deep learning models for identifying web registers---text varieties such as news reports and discussion forums---across 16 languages. We introduce the Multilingual CORE corpora, which contain over 72,000 documents annotated with a hierarchical taxonomy of 25 registers designed to cover the entire open web. Using multi-label classification, our best model achieves 79\% F1 averaged across languages, matching or exceeding previous studies that used simpler classification schemes. This demonstrates that models can perform well even with a complex register scheme at multilingual scale. However, we observe a consistent performance ceiling across all models and configurations. When we remove documents with uncertain labels through data pruning, performance increases to over 90\% F1, suggesting that this ceiling stems from inherent ambiguity in web registers rather than model limitations. Analysis of hybrid texts (those combining multiple registers) reveals that the main challenge lies not in classifying hybrids themselves, but in distinguishing hybrid from non-hybrid documents. Multilingual models consistently outperform monolingual ones, particularly for languages with limited training data. Zero-shot performance on unseen languages drops by an average of 7\%, though this varies by language (3--8\%), indicating that while registers share features across languages, they also retain language-specific characteristics.
\end{abstract}

\title{Automatic register identification for the open web using multilingual deep learning}

\maketitle

\section{Introduction} 

The internet has produced vast amounts of text in numerous languages and contexts, from news and information sites to instructions, social media content, and machine-generated texts. This diversity is captured in large-scale web corpora, which have become essential for advancing Natural Language Processing (NLP), especially in training Large Language Models (LLMs) that require diverse, extensive training data. However, these datasets typically lack metadata about their text types, which makes it difficult to control what material models learn from. Adding such metadata helps balance these datasets to better represent different language use scenarios and addresses common challenges in web-crawled texts, including political and demographic biases, toxicity, and data privacy \citep[e.g.,][]{gehman-etal-2020-realtoxicityprompts, dodge2021documenting, feng-etal-2023-pretraining, gururangan2023scaling, kumar-etal-2023-language, mallen-etal-2023-trust, penedo2023refinedweb, tokpanov2024zyda}.

\textit{Automatic register identification} provides a systematic and linguistically well-founded approach to creating metadata and curating data. The concept of register---related to but distinct from \textit{genre}---refers to how language is used in different situational and communicative contexts. Research on register variation is well-established in corpus linguistics and has greatly improved our understanding of the connection between situational context and linguistic variation  \citep{biber1988variation,10.1093/oso/9780195083644.003.0015,biber2019register}. Within NLP, register and genre classification have expanded to cover online texts such as how-to pages, discussion forums, and blogs, helping to describe the contents of massive web datasets \citep{Santini2005, Biber_Egbert_2018, Kuzman.Ljubesic2023}.

Until recently, automatic web register identification has faced major challenges. One key issue has been finding register classes that could cover the entire web. Earlier studies used small hand-picked text corpora, and classifiers trained on these datasets failed to capture the full range of online linguistic variation \citep{eissen_stein2004, santini2007automatic,sharoff-etal-2010-web, asheghi2016}. Web text itself poses another challenge: Studies show that web documents can mix multiple registers, forming so-called \textit{hybrids}, display only vague features of a broad class, or fail to fit any known register category \citep{Biber_Egbert_2018,biber2023register-culture,sharoff-2019-dimensions}. NLP studies have not yet shown how to solve these problems while maintaining high classification performance across noisy web data. Low inter-annotator agreement in previous work has also raised doubts about whether web register classification is even feasible \citep[e.g.,][]{eissen_stein2004, asheghi-etal-2014-designing, asheghi2016}. Additionally, most register identification work has focused on English \citep{biber-egbert2016-using-grammatical-feats,Biber_Egbert_2018,BiberEgbertKeller+2020+581+616,laippala_ronnqvist2023} with only limited research extending to other languages or multilingual settings \citep[e.g.,][]{skantsi_laippala_2023, repo2021zeroshot, ronnqvist-etal-2021-multilingual,Kuzman.Ljubesic2023}. It remains an open question to what extent web register identification can benefit from multilingual deep learning models, and what level of performance is achievable when identifying registers across the entire noisy, multilingual web.

In this article, we address these issues by introducing the Multilingual CORE corpora, which provide register annotations for 16 typologically diverse languages using a detailed hierarchical 25-class scheme, and by training multilingual register classifiers on this dataset. To evaluate our classification system, we compare it with the simpler X-GENRE taxonomy \citep{Kuzman.Ljubesic2023}. We examine how common challenges in web register modeling---outliers, ambiguous cases, and hybrids---affect classifier performance, and explore strategies for handling them without sacrificing performance or classification granularity. Our main contributions are:

\begin{itemize}
    \item We release the Multilingual CORE corpora: manually annotated web register data covering 16 languages and 72,504 documents, available at \url{https://github.com/TurkuNLP/multilingual-CORE}. We also release a multilingual register classifier trained on these corpora at \url{https://huggingface.co/TurkuNLP/web-register-classification-multilingual}, with training code at \url{https://github.com/TurkuNLP/pytorch-registerlabeling}.
    \item  We demonstrate successful multilingual register identification using transformer-based models \citep{NIPS2017_3f5ee243}, achieving an average 77\% micro F1 score across 25 classes on the five main training languages (English, Finnish, French, Swedish, and Turkish).
    \item We show that zero-shot evaluation on the five main languages decreases performance by 7--9 percentage points compared to including them in multilingual training. For the 11 smaller evaluation language datasets, zero-shot performance ranges from 51\% to 82\% F1, with large variation across languages.
    \item We demonstrate that our hierarchical multi-label classification system performs as well as the simpler 9-class X-GENRE \citep{Kuzman.Ljubesic2023} while providing more detailed register classification.
    \item We quantify how web noise affects register identification: removing documents with uncertain labels (including ambiguous cases, outliers, and potential label errors) improves model performance by more than 10\%, reaching over 90\% F1 score.
    \item Interestingly, our best performance (79\% F1 score) is achieved by training on non-hybrids (documents belonging to a single register class) while testing on the standard test set that includes hybrids. This score matches or exceeds previous studies using simpler schemes \citep{ronnqvist-etal-2021-multilingual,Kuzman.Ljubesic2023} or monolingual data \citep{skantsi_laippala_2023,laippala_ronnqvist2023}.
    \item We compare several register classifier models, finding that XLM-R Large \citep{conneau-etal-2020-unsupervised} currently offers the best balance of speed and accuracy.
\end{itemize}

We highlight four unexpected findings here. First, the highest overall score comes from a model trained only on non-hybrid documents and evaluated on the full test set, which includes hybrids (Section \ref{subsubsec:hybrids}). This model outperforms a model trained on all documents, even though its training set is smaller. However, the margin is only 2 percentage points, and the non-hybrid model performs poorly on hybrids (55\% micro F1). We therefore release the model trained on all documents. Second, hybrid documents are not in themselves harder to classify than non-hybrid documents. Models trained and tested only on hybrids perform nearly as well as models trained and tested only on non-hybrids. The main difficulty lies in telling hybrid and non-hybrid documents apart (Section \ref{subsubsec:hybrids}). Third, removing documents with uncertain labels from the training data lowers performance on the full test set (Section \ref{subsubsec:pruning-experiments}). These noisy examples thus carry information that the classifier needs to handle real web data. Fourth, scores on the full 25-class taxonomy are only 1--2 percentage points below scores on the nine main registers alone (Section \ref{subsec:main_results}). The subregisters therefore add detail at little cost to classification performance.

\section{Background} \label{sec:background}

The automatic identification of registers and genres in web content has a long history in NLP and corpus linguistics \citep{kessler-etal-1997-automatic,dewe-etal-1998-assembling,FromtheBNCtowardtheCybercorpusAQuantumLeapintoChaos,kilgarriff2003introduction,MakingtheWebMoreUsefulasaSourceforLinguisticCorpora,Towardsataxonomyofwebregistersandtexttypesamultidimensionalanalysis}. Already in 1997, \citeauthor{kessler-etal-1997-automatic} recognized that ``the problems of genre classification don't become salient until we are confronted with large and heterogeneous search domains like the World-Wide Web.'' \citet{kilgarriff2003introduction} established the foundations of web register classification, and while the field has advanced significantly since then, handling the vast diversity of web content remains challenging \citep{Argamon2019,Kuzman.Ljubesic2023}. Below, we review key developments in this field.

\subsection{Terminology: register vs. genre}

Previous work on automatic classification of web texts often uses the terms \textit{register} and \textit{genre} interchangeably \citep{sharoff-2019-dimensions,Kuzman-genre-identification-2023, Kuzman.Ljubesic2023}, and in automatic classification tasks, this distinction has not been crucial in practice. However, register and genre originate from different disciplines with distinct theoretical foundations. Genre is rooted in discourse analysis \citep{swales1990genre} and the register approach emerged from corpus linguistics \citep{biber1988variation,biber2019register}.

In the approach we adopt for this study, registers are defined as distinct varieties of language associated with specific situations of use \citep{egbert2015developing}. Registers are characterized by both their linguistic features (which automatic classifiers can learn from) and their situational characteristics \citep[which are used to define the register classes;][Section 2]{Biber_Egbert_2018}. These two aspects are presumed to have a direct relationship, where linguistic features serve specific functions required by the situational context.

\subsection{Challenges in web register identification}

One fundamental challenge in web register identification is defining appropriate register categories. Register classification schemes have varied considerably between researchers and research traditions \citep[reviewed by][]{sharoff-etal-2010-web}. This has led to inconsistencies, where even similarly named categories have substantially different definitions across studies \citep{eissen_stein2004,santini2007automatic, sharoff-etal-2010-web}. Data collection approaches have also created limitations: most early datasets were built by first defining register categories and then selecting matching documents \citep{eissen_stein2004,santini2007automatic, asheghi2016}. This top-down approach captured only a selective view of web language and failed to generalize well to broader web-crawled datasets \citep{egbert2015developing}.

A major challenge when examining the entire web is the inherent variability of register categories. While some registers have clear, distinctive characteristics, others are ambiguous and difficult to classify \citep{laippala_ronnqvist2023, henriksson-etal-2024-discrete}. This variability causes human annotators to struggle with consistent identification, leading to low inter-annotator agreement in multiple studies and raising questions about whether automatic classification is feasible \citep{eissen_stein2004, asheghi-etal-2014-designing, egbert2015developing, asheghi2016}.

A further challenge stems from documents that do not fit neatly into single register categories. Hybrid documents combine features from multiple registers and make up to 25\% of web documents; others lack clear register characteristics altogether \citep{biber2023register-culture}. Because of these challenges, researchers often exclude documents with ambiguous or multiple register features from their analyses \citep{biber-egbert2016-using-grammatical-feats, Biber_Egbert_2018, Kuzman-genre-identification-2023}, resulting in a restricted view of web language.

\subsection{Methodological developments}

Web register classification advanced significantly with the recognition that hierarchical label taxonomies improve classification, as first implemented in the Hierarchical Genre Collection \citep{stubbe2007towards}, KRYS-1 \citep{berninger2008building}, and SANTINIS-ML \citep{santini2010cross}. As \citet{Madjarov.etal2019} note, online text naturally forms hierarchies---recipes, for instance, are a subtype of instructional text. Hierarchical taxonomies allow classification at varying levels of specificity, depending on how clear the document's features are \citep{egbert2015developing}. Another key advance was extending labeling systems to cover the entire web \citep{santini2009web,sharoff-etal-2010-web}. The Corpus of Online Registers of English (CORE) achieved this first, using near-random Google searches and a data-driven annotation scheme \citep{egbert2015developing}. The original CORE scheme (discussed in Section \ref{subsec:register-taxonomy}) has eight main categories and nearly 50 subregisters in its hierarchy.

\Citet{laippala_ronnqvist2023} tested various machine learning methods and classification approaches to improve CORE identification scores. They found that allowing multiple labels per instance works significantly better than forcing single labels \citep[see also][]{santini2007automatic, vidulin2007using, egbert2015developing, Madjarov.etal2019, Sharoff2021, kuzman-etal-2022-ginco}. Specifically, \citet{laippala_ronnqvist2023} used multi-label classification to assign both main and subregister labels where appropriate, with documents lacking clear subregister features receiving only main register labels. This hierarchical approach achieved a 68\% F1 score with their BERT model \citep{devlin-etal-2019-bert}. The study also showed that web registers vary greatly in how distinctive they are, which helps explain some of the challenges in their classification.

In another line of work, \citet{sharoff-2019-dimensions} developed the Functional Text Dimensions (FTD) approach. FTD models texts by how similar they are to prototypes, which helps handle both hybrid documents and varying register specificity. Using an ensemble classifier, this method reached an 82\% F1 score across 18 dimensions for English \citep{lepekhin2022estimating}. \citet{kuzman-etal-2022-ginco}, on the other hand, created GINCO, a web genre corpus for Slovenian, and achieved a 78\% F1 score using XLM-R \citep{conneau-etal-2020-unsupervised} for 12 genre classes that match CORE. To compare across languages, they trained models on machine-translated English versions of GINCO and tested them on CORE texts, reaching a 63\% F1 score---showing promise for multilingual research. \citet{Kuzman-genre-identification-2023} compared different register schemes and datasets, including CORE, FTD, and GINCO. They achieved their best results---80\% micro-average F1 using XLM-R Base with single labels---on X-GENRE, which combines all three corpora into nine classes.

Following \citet{sharoff-etal-2010-web}, \citet{Kuzman.Ljubesic2023} found that more classes typically lead to less reliable annotations and lower classifier accuracy. For GINCO, expert annotators achieved good agreement (Krippendorff's alpha of 0.71) through careful double-checking. The English CORE had lower reliability, with at least three of four MTurk coders agreeing on 69\% of texts \citep{egbert2015developing}. However, \citet{skantsi_laippala_2023} reached 80\% agreement using the CORE scheme for FinCORE, which has 9 main registers and 30 subregisters (see Section \ref{subsec:register_datasets}). These results show that consistent register annotation is achievable even with fine-grained taxonomies, though outcomes depend heavily on annotator expertise and methodology.

\subsection{Towards multilingual CORE}

Since the publication of the English CORE, several corpora and classifiers have been developed for different languages, using the CORE taxonomy to cover the unrestricted web. \Citet{skantsi_laippala_2023} introduced the Finnish FinCORE dataset and achieved a 79\% F1 score using XLM-R. \Citet{repo2021zeroshot} created smaller collections in French and Swedish and demonstrated that models trained on English could classify registers in these languages without language-specific training, achieving 61--69\% F1. Building on this work, \citet{ronnqvist-etal-2021-multilingual} investigated whether training on multiple languages simultaneously could improve performance, reporting significant gains over single-language training. \Citet{laippala-etal-2022-towards} then applied \citeauthor{ronnqvist-etal-2021-multilingual}'s best model to eight additional languages without any language-specific training data, reaching F1 scores between 58\% and 82\%, depending on the language. XLM-R Large performed the best in all tests. They also further developed the original, nearly 50-class CORE scheme by excluding the registers featuring the most challenges to both IAA and classification. This resulted in the 25-class version of the CORE taxonomy that is applied in the present article as well (see Section \ref{subsec:register_datasets}).

In summary, recent advances in web register identification show both progress and remaining challenges. Deep learning methods and new datasets---especially CORE, FinCORE, and the smaller GINCO and FTD---have significantly improved classification performance and cross-lingual generalization. However, three key challenges remain: classifying documents without clear register features, handling documents that combine multiple registers, and determining the appropriate number of classes to represent the full diversity of web language.

\section{Materials} \label{sec:materials}

This section introduces the Multilingual CORE Corpora, a collection of register-annotated datasets for training and testing register classifiers. We describe the corpora, explain the CORE taxonomy, and examine how documents are distributed across registers and languages. Finally, we discuss how CORE registers map to the X-GENRE taxonomy.

\subsection{Web register datasets} \label{subsec:register_datasets}

\begin{table}
  \caption{Multilingual CORE Corpora studied in this article.}\label{tbl:corpora}
  {\tablefont
  \begin{tabular}{@{\extracolsep{\fill}}lrl}
  \hline
  Language & $N$ Documents &  Publication \\
  \hline
  \textbf{Training sets} \\
  \hline
     English (CORE) & 47,853 &\citet{Biber_Egbert_2018, laippala_ronnqvist2023} \\\hdashline
     Finnish (FinCORE) & 10,751 & \citet{skantsi_laippala_2023} \\\hdashline
     Swedish (SweCORE) & 4,275 & New*%
     \\\hdashline
     French (FreCORE)  & 4,105 & New*%
     \\\hdashline
     Turkish (TurCORE) & 2,763 & Published in \citet{Erten-Johansson.etalForthcoming} but not used for register identification\\
    \hline
    \textbf{Evaluation sets} \\
    \hline
     Arabic & 92 & \citet{laippala-etal-2022-towards} \\\hdashline
     Catalan & 109 & \citet{laippala-etal-2022-towards} \\\hdashline
     Spanish & 100 & \citet{laippala-etal-2022-towards} \\\hdashline
     Farsi & 68 & New%
     \\\hdashline
     Hindi  & 160 & \citet{laippala-etal-2022-towards} \\\hdashline
     Indonesian & 1,186 & \citet{laippala-etal-2022-towards} \\\hdashline
     Japanese & 100 & New%
     \\\hdashline
     Norwegian & 133 & New%
     \\\hdashline
     Portuguese & 332 & \citet{laippala-etal-2022-towards} \\\hdashline
     Urdu & 160 & \citet{laippala-etal-2022-towards} \\\hdashline
     Chinese & 317 & \citet{laippala-etal-2022-towards} \\
    \hline
    \textbf{Total} & \textbf{72,504} & \\
    \hline
    \end{tabular}
  }
  {\begin{tabnote}
  *Significantly smaller versions of SweCORE (2,182 documents) and FreCORE (1,818 documents), featuring only main registers, have been published in \citet{repo2021zeroshot}.\\
  \end{tabnote}}
\end{table}

The datasets in our study cover 16 typologically diverse languages and use the 25-class CORE-based taxonomy introduced by \citet{laippala-etal-2022-towards} (see Section \ref{subsec:register-taxonomy}). All corpora are listed in Table \ref{tbl:corpora} and are available at \url{https://github.com/TurkuNLP/multilingual-CORE}.

The collection extends previously published datasets and adds substantial new material. It includes five large datasets (English, Finnish, Swedish, French, and Turkish) and smaller evaluation datasets for 11 additional languages. We have significantly expanded the French and Swedish corpora in both size and annotation granularity, with the annotation scheme now using 25 classes instead of the previous nine main CORE labels (nine main labels plus 16 subregister labels; see Section \ref{subsec:register-taxonomy} for details). We also introduce three entirely new datasets for Farsi, Japanese, and Norwegian.

Previous research with these corpora has been limited. The English CORE and FinCORE have only been used in monolingual experiments \citep{laippala_ronnqvist2023,skantsi_laippala_2023}, while multilingual experiments have been restricted to the English CORE and smaller versions of the Finnish, French, and Swedish corpora using only main register labels \citep{repo2021zeroshot,ronnqvist-etal-2021-multilingual}. \citet{laippala-etal-2022-towards} later applied the best multilingual model to several additional languages but did not develop the model further. The Turkish dataset \citep{Erten-Johansson.etalForthcoming}, originally published as a corpus linguistic study, is used here for register identification for the first time.

\begin{figure}
  \centering
  \begin{minipage}{0.75\textwidth}
   \includegraphics[width=\textwidth]{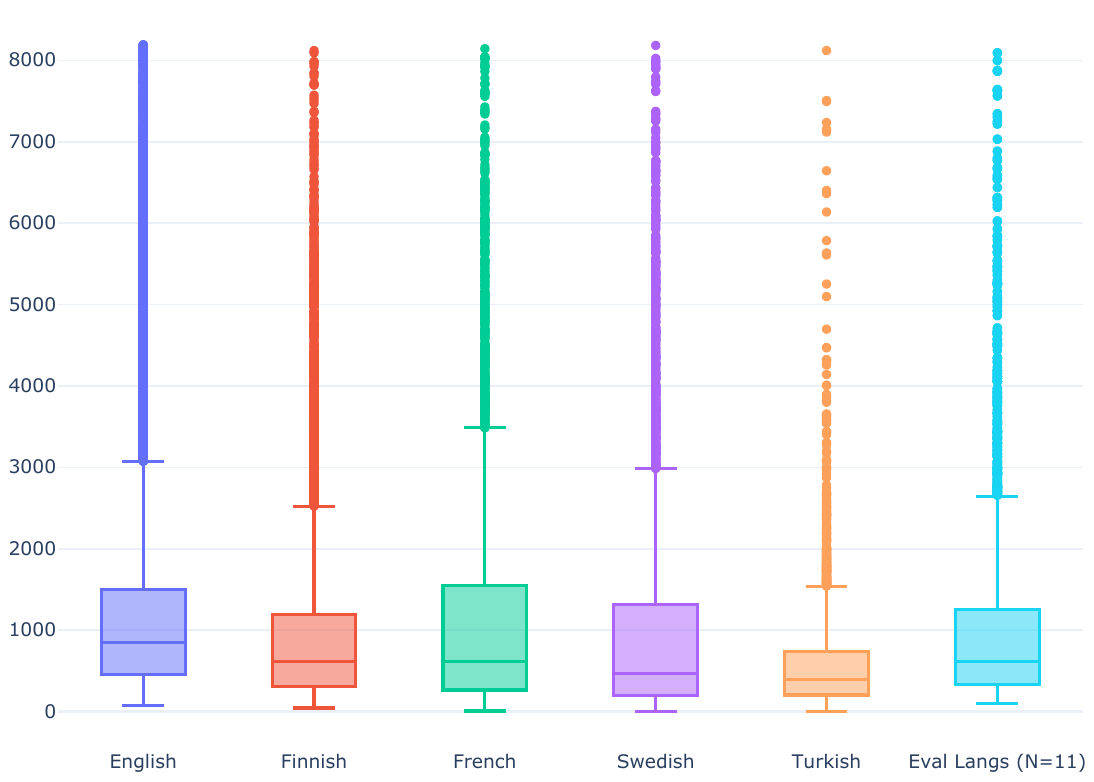}
   \vspace{0.5em}
  \caption{Distribution of text lengths across the Multilingual CORE Corpora, measured in SentencePiece tokens. Documents longer than 8,192 tokens (2\% of the dataset) are not shown for clarity.}
  \label{fig:document_lengths}
  \end{minipage}
\end{figure}

Figure \ref{fig:document_lengths} shows the document lengths in our corpora, counted using XLM-R's SentencePiece tokenizer \citep{kudo2018sentencepiecesimplelanguageindependent}. Most languages have median lengths between 400 and 600 tokens, while English documents are notably longer (median around 800 tokens). Since most encoder models in our experiments have a 512-token limit, we need to truncate longer documents. Following \citet{laippala_ronnqvist2023}, who found that initial text segments perform better for register classification, we use the first 512 tokens of each document in most experiments.

Our study provides the first comprehensive multilingual evaluation of the 25-class CORE-based scheme, including both main registers and subregisters. By combining expanded datasets with transformer-based encoder models across 16 languages, we aim to establish new benchmarks for multilingual register identification.

\subsection{Corpus compilation and annotation strategies} \label{subsec:corpus-compilation}

All Multilingual CORE Corpora datasets were collected from the unrestricted open web. As with other large multilingual data collections with standardized annotations, such as Universal Dependencies \citep{de-marneffe-etal-2021-universal}, some differences in compilation and annotation methods exist despite globally compatible strategies. For the English CORE, texts were collected through Google searches of highly frequent English 3-grams \citep{egbert2015developing}. Four Amazon Mechanical Turk (MTurk) annotators coded each text using a decision tree to classify documents into eight main registers and 44 subregisters. Documents that did not clearly fit any subregister received only a main register label. The final label for a document required agreement from at least two annotators \citep[for details, see][]{Biber_Egbert_2018}.

FinCORE \citep{skantsi_laippala_2023} was constructed from a random sample of the Finnish Internet Parsebank \citep{luotolahti-etal-2015-towards}, which was built by targeted crawling of Finnish web documents using language-detected seed URLs and by extracting Finnish content from Common Crawl. All remaining corpora were compiled directly from Common Crawl following \citet{laippala-etal-2022-towards}. All corpora underwent boilerplate removal, deduplication, and filtering of non-main text content.

For FinCORE and the Common Crawl-based corpora, trained experts with native-level language skills annotated texts using the CORE register scheme. Unlike the English CORE's multi-annotator approach, these corpora used single annotators who could assign up to two labels per text. Double-annotated samples showed strong inter-annotator agreement: 80\% for Finnish \citep{skantsi_laippala_2023}, 78\% for French, and 84\% for Swedish \citep{repo2021zeroshot}.

\subsection{Register taxonomy and distribution} \label{subsec:register-taxonomy}

\begin{table}

  \caption{Register labels and their abbreviations in the simplified CORE scheme adopted for this study. \\\hspace{\textwidth} Main register categories are in bold.}\label{tbl:scheme}
  {\tablefont
  \begin{tabular}{@{\extracolsep{\fill}}llll}
  \hline
 Name & Abbr. & Name & Abbr. \\
\hline
\textbf{Machine-translated} & MT & \textbf{Informational description} & IN \\\hdashline
\textbf{Lyrical} & LY &  Encyclopedia article & en \\\hdashline
\textbf{Spoken} & SP &   Research article & ra \\\hdashline
  Interview & it &  Description of a thing or person & dtp \\\hdashline
\textbf{Interactive discussion} & ID &  FAQ & fi \\\hdashline
\textbf{Narrative} & NA &  Legal terms \& conditions & lt \\\hdashline
  News report & ne & \textbf{Opinion} & OP \\\hdashline
  Sports report & sr &  Review & rv \\\hdashline
 Narrative blog & nb &  Opinion blog & ob \\\hdashline
\textbf{How-to or instructions} & HI &   Denominational religious blog or sermon & rs \\\hdashline
  Recipe & re &  Advice & av \\\hdashline
\textbf{Informational persuasion} & IP \\\hdashline
  Description with intent to sell & ds \\\hdashline
   News \& opinion blog or editorial & ed \\
    \hline
    \end{tabular}
}
  {}
\end{table}

The corpora studied here were annotated using the register taxonomy originally developed for the English CORE \citep{egbert2015developing}. 
We use an adapted, simplified version of the scheme with nine main registers and 16 subregisters \citep{laippala-etal-2022-towards}, as shown in Table \ref{tbl:scheme}. This version improves on the original CORE scheme by excluding infrequent and poorly defined categories \citep[see][for discussion]{laippala_ronnqvist2023}.

\begin{figure}
    \includegraphics[width=1\linewidth]{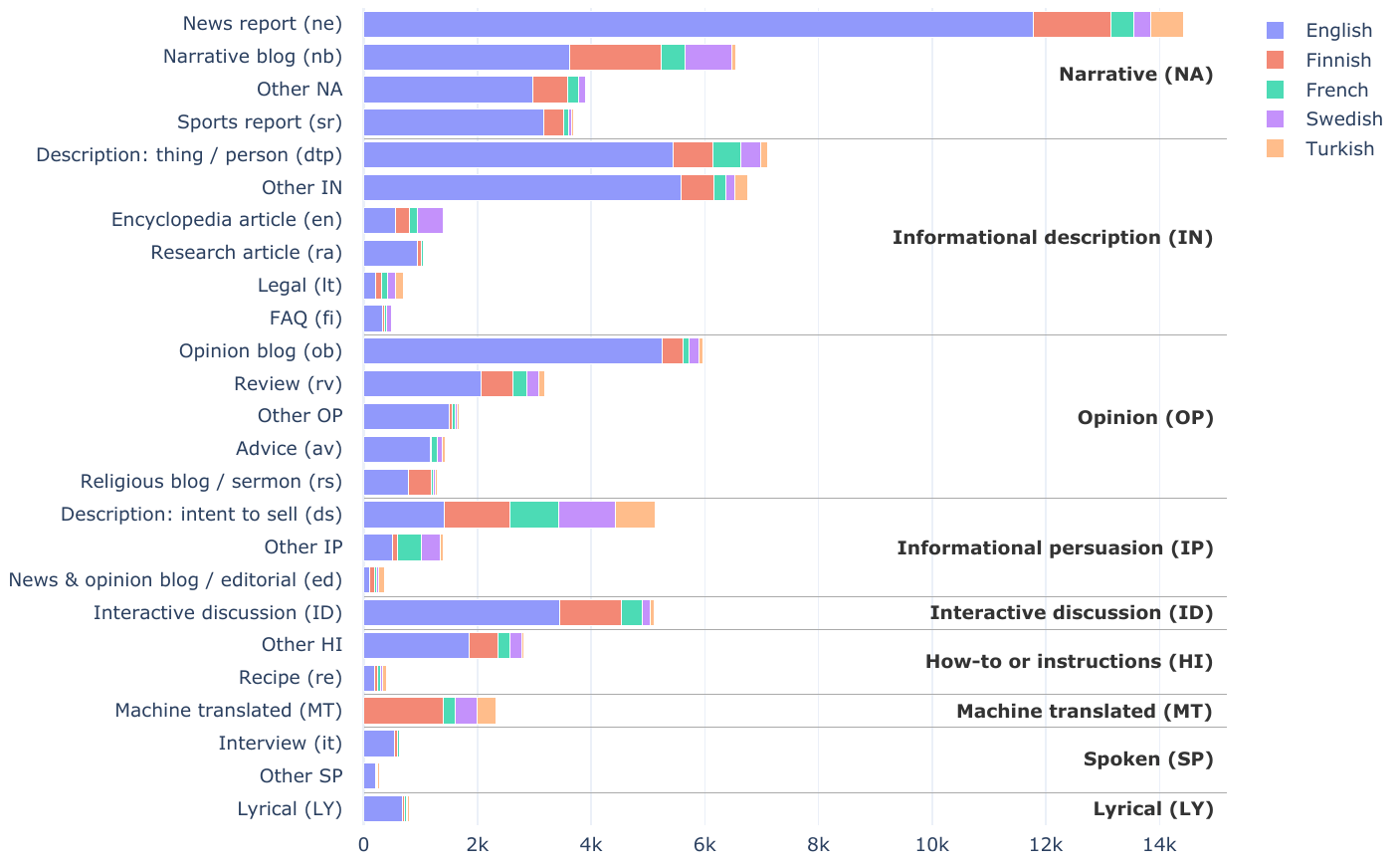}
    \vspace{0.5em}
    \caption{Main registers (right) and subregisters (left) with their distribution across the large corpora. The \textit{Other} categories include documents that could not be assigned to a subregister. In the classification process, these documents are assigned only the main label(s).}
    \label{fig:label_distributions_large}
\end{figure}

Figure \ref{fig:label_distributions_large} shows the registers and their distributions across the five large corpora. The distributions are highly imbalanced: \textit{News report}---a subregister of \textit{Narrative}---is by far the most common, followed by \textit{Description of a thing or person} within \textit{Informational description}. In contrast, \textit{Spoken} and \textit{Lyrical}, along with \textit{FAQ}, \textit{Recipe}, and \textit{News \& opinion blog or editorial}, appear much less frequently.

However, class size is not the only factor affecting classification performance. Linguistic distinctiveness also plays a role \citep{kumar-etal-2023-language}. For instance, despite their small sizes, \textit{Recipe} and \textit{Encyclopedia article} have clear linguistic features and have scored well in previous identification studies \citep{laippala2021exploring,laippala_ronnqvist2023}. We therefore kept these categories in our CORE version while removing some larger but linguistically unclear ones \citep[following][]{laippala_ronnqvist2023}. Our 25-class taxonomy balances adequate category sizes with distinctive linguistic features, considering the classification challenges noted above.

The languages also differ notably in their register distributions. \textit{Legal terms \& conditions}, \textit{Informational persuasion} and its subregisters appear much less often in English than in other languages, while \textit{Machine-translated} does not appear in the English CORE at all. Most differences likely result from different sampling methods (see Section \ref{subsec:corpus-compilation}); for example, Common Crawl includes more commercial pages and legal notices than Google searches. Other differences reflect language patterns and sample sizes. English lacks \textit{Machine-translated} texts because English is often the source language for translations or translations occur between other languages. Conversely, almost all \textit{Spoken} and \textit{Lyrical} examples come from English, probably because the English corpus is much larger and thus more likely to include these relatively rare web documents.

Figure \ref{fig:label_distributions_large} also shows how many documents received only main register labels, without a specific subregister. We assigned these general-level labels to documents that did not fit any subregister (see Section \ref{subsec:corpus-compilation}). These cases are marked in the Figure as \textit{Other [register abbreviation]} (e.g., \textit{Other NA}), though the datasets use only the main labels (e.g., \textit{NA}). Such general-label documents are common: \textit{Other IN} and \textit{Other NA} are the second and third largest categories within \textit{Informational description} and \textit{Narrative}, respectively.

\begin{figure}
  \centering
  \begin{minipage}{1\textwidth}
    \includegraphics[width=\textwidth]{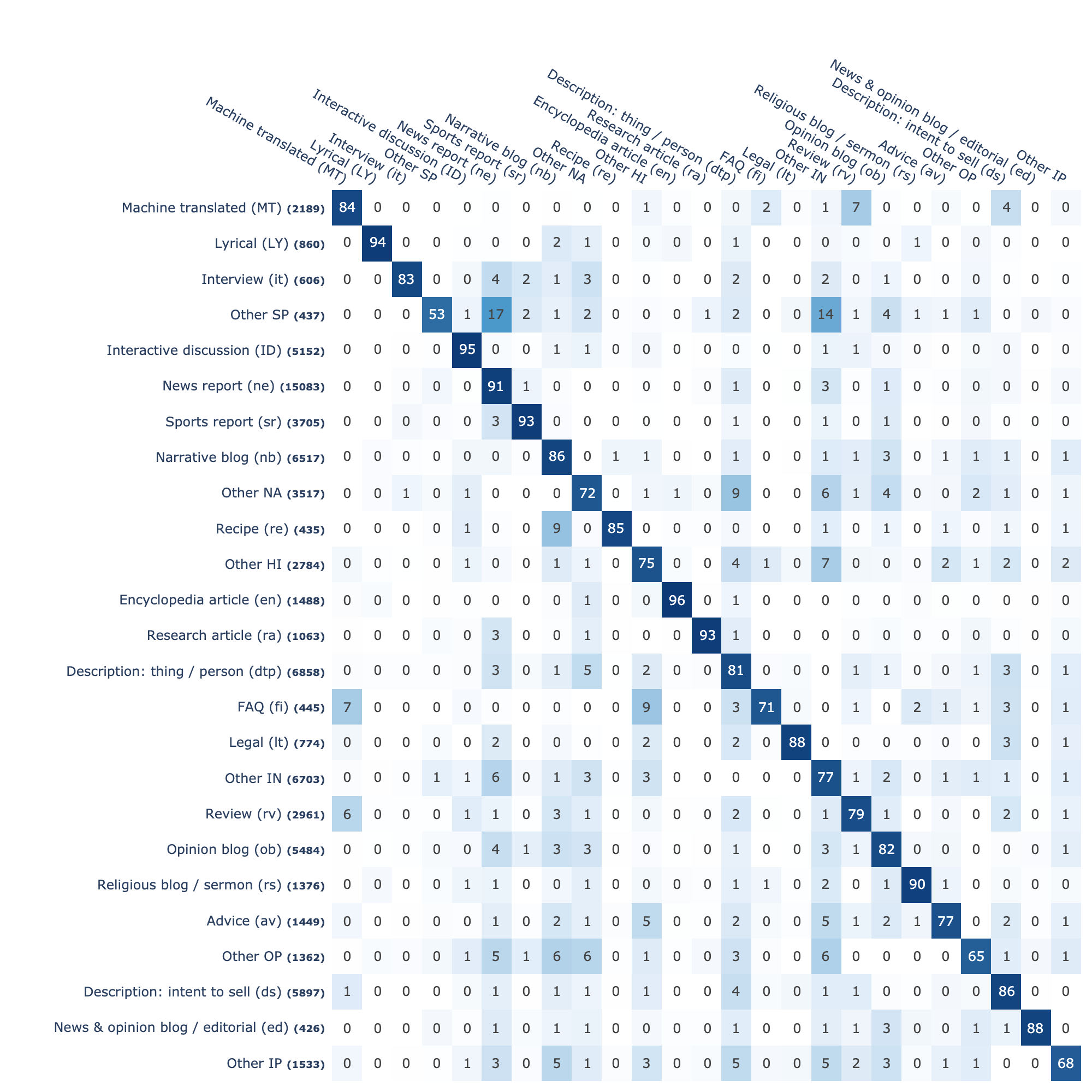}
   \vspace{0.5em}
    \caption{Percentage heatmap of class co-occurrences across 16 languages. Totals (in parentheses) show overall class occurrences. Diagonal cells display percentages of singly appearing labels, and off-diagonal cells show co-occurrence (i.e. hybrid) proportions.}
    \label{fig:label_co_occurrences}
\end{minipage}
\end{figure}

Hybrid documents form another important category. Figure \ref{fig:label_co_occurrences} shows how registers combine in these documents, revealing several patterns. First, combinations of only main register labels, such as \textit{Other IN} + \textit{Other NA}, likely represent documents that lack specific characteristics of any single register (cf. \citet{biber2023register-culture}). Second, certain combinations indicate different registers appearing in separate parts of a document. For example, texts labeled \textit{Recipe} + \textit{Narrative blog} often occur where the blogger tells a story associated with the recipe, a common characteristic of food blogging. Third, other combinations suggest multiple communicative purposes mixed throughout the text. These include \textit{Other OP} co-occurring with narrative subregisters such as \textit{Narrative blog}, and the opinionated \textit{Advice} subregister appearing alongside the more neutral \textit{FAQ} and \textit{Other HI}. Such patterns suggest that distinguishing between opinionated and neutral discourse can sometimes be challenging.

Example \ref{example1} illustrates the second type of hybrid document, where two registers (\textit{Narrative blog} and \textit{Recipe}) appear in clearly distinct parts.

\ex.
    Äggröra med färska Tomater och Basilika till rökt Fläsk och Paprika\\\\
    Den här frukosten åt jag häromdagen innan jag gav mig ut i skogen på bärplockning. Den stod jag mig länge på. Testa den en söndag. Paprika och fläsk blev en riktigt god kombination till äggröran. Mycket bättre än sladdrigt bacon.\\\\
    Äggröra med färska Tomater och Basilika\\\\
    ½ dl Grädde\\
    1 msk hackad färsk Basilika\\
    Havssalt och Svartpeppar\\
    2 tjocka skivor Varmrökt Sidfläsk\\
    1 tunn skiva Surdegsbröd\\\\
    Vispa ihop äggen, grädden och kryddorna lätt. Hacka tomater och basilika. Smält smör i en kastrull, rör i tomaterna och låt fräsa en liten stund. Häll i äggsmeten, rör runt lite. [...]\\\\
    \textit{[Scrambled eggs with fresh tomatoes and basil, smoked pork and bell pepper.\\\\
    The other day I ate this breakfast before I went to the forest to pick berries. It kept me full for a long time. Try it on a Sunday. Bell pepper and pork was a good combination with scrambled eggs. Much better than flabby bacon.\\\\
    Scrambled eggs with fresh tomatoes and basil\\\\
    ½ dl cream\\
    1 tbsp chopped fresh basil\\
    seasalt and black pepper\\
    2 thick slices smoked pork belly\\
    1 thin slice sourdough bread\\\\
    Whip the eggs, cream and spices lightly. Chop tomatoes and basil. Melt butter in a saucepan, add the tomatoes and fry for a while. Pour in the egg batter, mix a little. […]]}\label{example1}

Example \ref{example1} starts with a personal narrative about the author's day, moves to comments about the dish, and ends with a standard recipe format listing ingredients and preparation steps. The recipe section is clearly distinguishable from the personal story that frames it.

\subsection{CORE conversion to X-GENRE}
\label{subsec:core2xgenre}

\begin{figure}
  \centering
  \begin{minipage}{0.75\textwidth}
    \includegraphics[width=\textwidth]{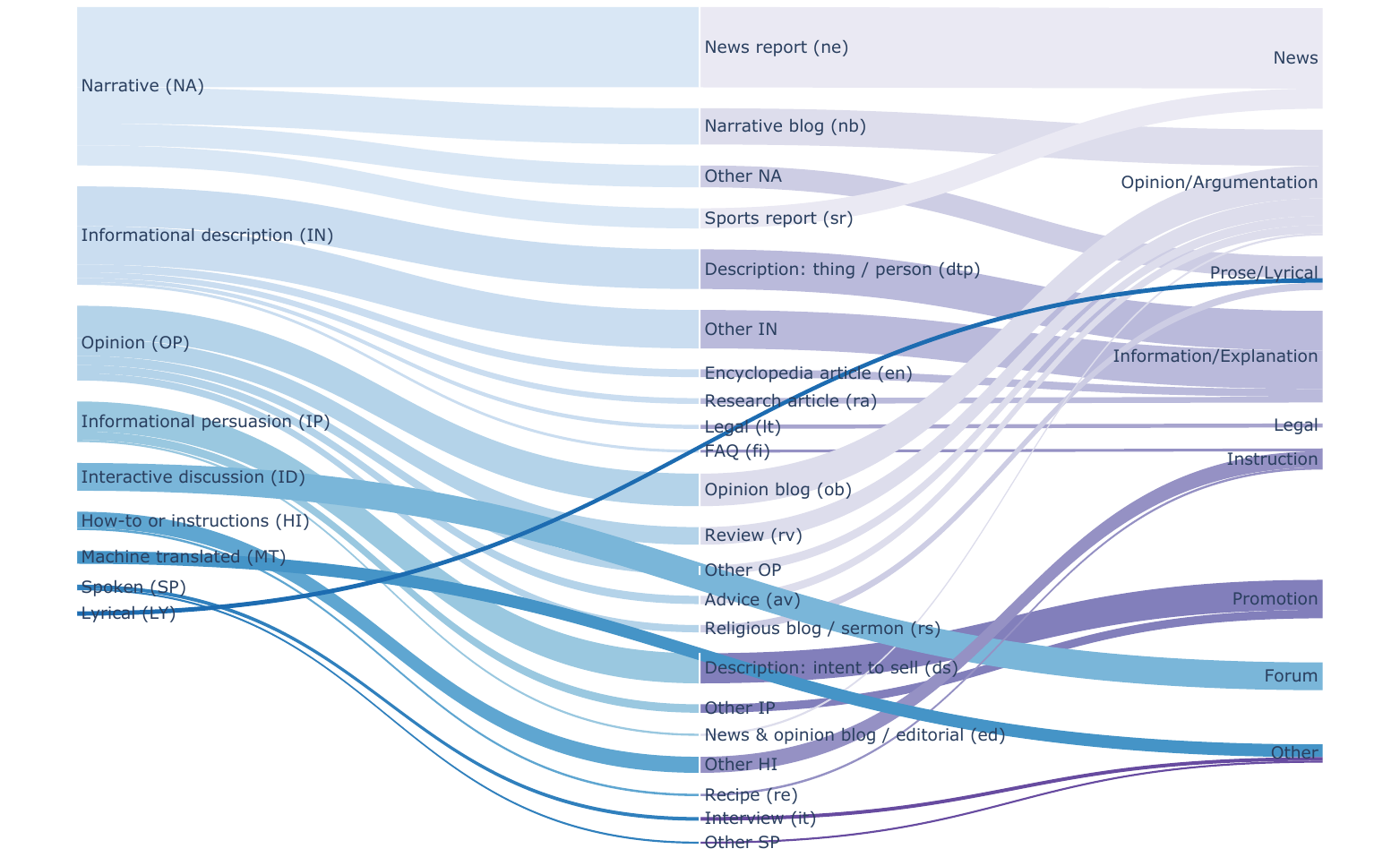}
    \vspace{0.5em}
    \caption{Mapping from CORE main registers (left) and subregisters (middle) to XGENRE (right).}
    \label{fig:xgenre_mapping}
\end{minipage}
\end{figure}

To explore different web register taxonomies, \citet{Kuzman-genre-identification-2023} trained classifiers on GINCO, FTD, CORE, and a new combined dataset called X-GENRE. X-GENRE uses a simpler non-hierarchical taxonomy with nine categories, where \textit{Other} covers hybrid cases. In their study, this taxonomy achieved 80\% F1, outperforming both the main CORE registers (75\% F1) and the full set of CORE subregisters (66\% F1). We compare our CORE-based scheme with X-GENRE by converting our labels following the mapping in \citet{Kuzman-genre-identification-2023} (see Figure \ref{fig:xgenre_mapping}). While the schemes mostly align, some conversions raise theoretical issues. X-GENRE maps \textit{Other Narrative} to \textit{Prose/Lyrical}, though linguistically it more closely resembles \textit{Informational description} \citep{doi:10.1177/0075424216628955}. It also assigns \textit{Narrative blog} to \textit{Opinion/Argumentation}, even though narratives do not typically express opinions. Nevertheless, we maintain these mappings for fair comparison with \citet{Kuzman-genre-identification-2023}.

\section{Methods}

In this section, we outline our methodology for multilingual web register classification. We start with dataset preprocessing, including data splits and label encoding for hierarchical multi-label classification. We then address language and class imbalance issues that arise from varying dataset sizes and register frequencies. Next, we describe our transformer-based classifier models and training procedures, followed by our model explanation approach using SACX to identify linguistically meaningful keywords. Finally, we present our data pruning methodology to evaluate whether removing ambiguous documents improves classification performance.

\subsection{Data splits} \label{preprocessing}

Following standard practice in supervised learning, we randomly divide our datasets into three splits: training (70\% of data), development (10\%), and test (20\%). We train our models on the training data, tune hyperparameters using the development data, and evaluate final performance on the held-out test data.

We create these splits separately for English, Finnish, French, Swedish, and Turkish (our training languages). We use stratified splitting to preserve the original distribution of classes in each split, which helps prevent model bias and improves performance on rare classes \citep{10.1007/978-3-642-23808-6_10}. The other 11 languages are used only to test cross-lingual zero-shot performance and therefore do not require splitting.

\subsection{Label encoding} \label{subsec:labelencoding}

We convert the CORE and X-GENRE register schemes into a numerical format suitable for multi-label classification using standard binary encoding. Each label is represented as a binary vector where positions correspond to specific labels, marked as 1 if present and 0 otherwise.

To preserve the hierarchical structure of the CORE taxonomy, we include both main labels and subregisters as separate positions in the label vector. When a text has a subregister label, we mark both that position and its parent main register as 1. Early tests showed this method captures hierarchical relationships well and rarely produces hierarchy-related prediction errors. We also tested an alternative flat encoding that treated each main register and its subregisters as completely separate categories, adding \textit{Other [register]} labels for documents without subregisters. This approach performed much worse and was abandoned.

\subsection{Language imbalance} \label{subsec:languageimbalance}

\begin{figure}
  \centering
  \begin{minipage}{0.5\textwidth}
  \includegraphics[width=\textwidth]{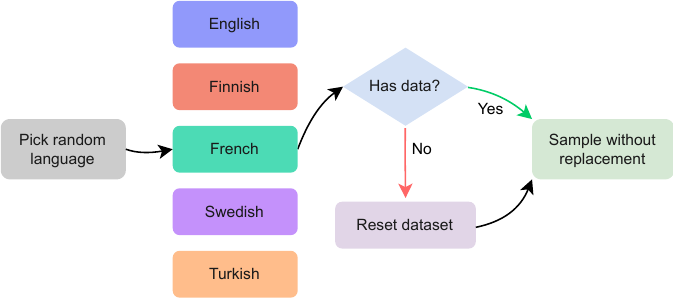}
  \vspace{0.5em}
  \caption{Data sampling process for multilingual fine-tuning, illustrating the repeated random selection and automatic replenishment of language datasets.} 
  \label{fig:sampler}
\end{minipage}
\end{figure}

As noted in Section \ref{subsec:register_datasets}, our datasets vary greatly in size, with English being the largest. In multilingual modeling, this size imbalance can bias models toward the largest language and reduce performance on others. To prevent this, we balance our training data using a custom sampling method.

We adapt the sampling method from \citet{ronnqvist-etal-2021-multilingual} with minor modifications. The sampler randomly selects a language, giving each language equal selection probability, then randomly draws a sample from the chosen language without replacement. Once all examples from a language are used, that language's dataset is replenished for the next round. This process is illustrated in Figure \ref{fig:sampler}. Unlike \citet{ronnqvist-etal-2021-multilingual}, who used single-language batches, we mix languages within each batch.

In preliminary experiments, we found that balanced sampling reduces bias toward larger language datasets. Using XLM-R Large \citep{conneau-etal-2020-unsupervised}, balanced sampling achieved a 0.4 percentage point improvement in micro F1 scores across languages, averaged over three training runs. The largest improvement was for Turkish (+3 points), our smallest dataset, followed by Swedish and French (+1 point each). Performance slightly decreased for the largest datasets: English (-2) and Finnish (-1). Based on these results, we adopt balanced sampling for all multilingual experiments in this article.

\subsection{Class imbalance} \label{subsec:classimbalance}

In initial experiments, we explored two methods to address class imbalance in our datasets (see Section \ref{subsec:register-taxonomy}). First, we tested class weighting, assigning weights inversely proportional to class frequency to increase penalties for misclassifying underrepresented classes. Second, we experimented with Focal Loss \citep{lin2018focal} to focus training on difficult examples. Neither approach improved model performance. We therefore use only stratified sampling to maintain the original class distribution (see Section \ref{preprocessing}), which likely reflects the natural frequency of different registers on the open web.

\subsection{Transformer-based classifiers} \label{subsec:classifiers}

Pre-trained transformer models have become the dominant approach in NLP, achieving state-of-the-art results across numerous tasks \citep{NIPS2017_3f5ee243,Wolf.etal2020}. In web register classification, transformer-based models have surpassed traditional methods such as SVM, fastText, and CNNs \citep[e.g.,][]{Kuzman.Ljubesic2023,laippala-etal-2019-toward,skantsi_laippala_2023}. In this study, we fine-tune several multilingual pre-trained models by adding classification heads and training them on register-labeled data.

\paragraph{XLM-RoBERTa} \label{subsubsec:xlm-roberta}

XLM-RoBERTa \citep{conneau-etal-2020-unsupervised} is a multilingual extension of the RoBERTa model \citep{liu2019roberta}, which builds on the original BERT architecture \citep{devlin-etal-2019-bert}. XLM-R is pre-trained on cleaned Common Crawl data in 100 languages. Previous studies have shown that XLM-R outperforms other multilingual models in web register identification and achieves comparable or superior results to monolingual models \citep{repo2021zeroshot, ronnqvist-etal-2021-multilingual, Kuzman.Pollak2022, Kuzman.Ljubesic2023}. We experiment with two variants: Large (561M parameters) and XL (3.5B parameters) \citep{goyal2021largerscale}. Both have a maximum token limit of 512, requiring truncation of longer texts. For the XL model, we employ Low-rank Adaptation (LoRA) \citep{hu2022lora}, a parameter-efficient fine-tuning method \citep{peft}, to reduce training time and computational requirements. We apply LoRA across all linear layers with $rank=128$ and $alpha=256$, chosen through early testing.

\paragraph{BGE-M3-RetroMAE} \label{subsubsec:bge}
The BGE M3-RetroMAE model \citep{bge-m3} is a multilingual embedding model initialized from XLM-RoBERTa with an extended maximum token length of 8,192. The model has undergone additional pretraining using the RetroMAE technique \citep{xiao2022retromae} on three datasets: the Pile \citep{Gao.etal2020}, Wudao \citep{Yuan.etal2021}, and mC4 \citep{2020t5}. Among the three BGE M3 variants, we select the RetroMAE variant for its general-purpose design and potential for enhanced cross-lingual transfer \citep{xiao2022retromae}. To evaluate the impact of sequence length, we fine-tune two versions with maximum token limits of 512 and 2,048, as most of our texts exceed 512 tokens (see Section \ref{subsec:register_datasets}).

\paragraph{Multilingual E5} \label{subsubsec:me5}
Like BGE M3, the Multilingual E5 (ME5) model \citep{wang2024multilingual} is initialized from XLM-RoBERTa. It has undergone additional pre-training on multilingual text pairs from diverse sources (Wikipedia, MC4, NLLB, Reddit) using contrastive learning, followed by supervised fine-tuning. This training aims to produce embeddings that better capture cross-lingual semantic similarities. We fine-tune the Large variant to test whether its specialized training improves register identification over XLM-R.

\paragraph{Mixtral 8x7B} \label{subsubsec:mixtral}

Finally, we experiment with Mixtral 8x7B \citep{jiang2024mixtral}, a causal (decoder-only) model designed primarily for text generation rather than natural language understanding. The model has 47B parameters total, but its mixture-of-experts architecture uses only 13B parameters in each forward pass. We fine-tune the model using a classification head and LoRA (see Section \ref{subsubsec:xlm-roberta}) with $rank=128$ and $alpha=256$ across all linear layers. To improve efficiency, we convert the model weights from 16-bit to 4-bit precision \citep{jacob2017quantization} and limit the context length to 512 tokens, despite its original 8,192-token context.

\subsection{Model parameters and training}

We fine-tune our models using default settings, adjusting hyperparameters only when necessary \citep{tuningplaybookgithub}. We use the Hugging Face Trainer class \citep{Wolf.etal2020}, modified for multi-label training with binary cross-entropy loss and a specialized dataloader for multilingual training (see Section \ref{subsec:languageimbalance}). All models use batches of 8 and train for up to 30 epochs, with early stopping if evaluation loss does not improve for 5 epochs. To prevent overfitting, we apply 0.01 weight decay and a 5\% warm-up ratio to stabilize early training.

Following \citet{Goodfellow-et-al-2016}, we focus on optimizing learning rates. We first use Ray Tune \citep{liaw2018tune} with HyperoptSearch \citep{pmlr-v28-bergstra13} to explore rates from $1 \times 10^{-8}$ to $1 \times 10^{-4}$ across selected models and datasets, using the ASHA scheduler \citep{li2020massively} to stop underperforming trials early. After this broad search, we run a focused grid search between $1 \times 10^{-6}$ and $1 \times 10^{-4}$. We also optimize the binary prediction threshold through grid search from 0.3 to 0.7 in 0.05 steps, using micro-averaged F1 score as the criterion. For efficiency, we load models in bfloat16 precision, except Mixtral, which uses 4-bit quantization (see Section \ref{subsubsec:mixtral}).

To evaluate generalization robustly \citep{mccoy-etal-2020-berts}, we run each experiment three times with different random seeds (42, 43, 44), keeping training parameters constant while varying only the models' initial weights and training sequence order. All experiments run on a single NVIDIA A100 Tensor Core GPU on the Mahti supercomputer at CSC --- IT Center for Science, Finland.

\subsection{Explaining the model decisions with keywords using SACX} \label{subsec:SACX}

To analyze the linguistic basis of classifier decisions, we examine keywords associated with each register class using the model explanation method SACX \citep{ronnqvist-etal-2021-multilingual}. SACX uses Integrated Gradients \citep{Sundararajan2017} to generate prediction-level explanations, which are then aggregated to form class-level keywords. The method involves training multiple models on different folds of the training data and averaging word-level prediction contributions across all trained models to identify words that consistently contribute to the classification of each class.

For this analysis, we combine the train and test splits of the dataset (see Section \ref{preprocessing}) and create 8 stratified folds from this combined set. We then run 50 iterations, each using 5 folds for training and 3 folds for the Integrated Gradients explanation step. Optimal model training parameters are determined using the development split, which is otherwise excluded from this analysis.

\subsection{Data pruning} \label{subsec:datacleaning}

As discussed in Section \ref{sec:background}, noisy web language poses a classic challenge for register classification: documents vary in how distinctive they are and how easily they can be labeled. We test whether removing hard-to-classify documents improves model performance. These include documents with ambiguous or unclear register features, outliers, and potentially mislabeled examples \citep{song2022learning}.

We use the Cleanlab tool and its \textit{confident learning} algorithm \citep{northcutt2021confidentlearning} to find documents with uncertain or unreliable labels. This model-agnostic approach has been shown to work well for finding labeling issues in multi-label and multi-annotator data \citep{goh2022crowdlab,thyagarajan2023multilabel,chen2024automated}. We evaluate different combinations of pruned and unpruned datasets during training and testing to see how this affects model robustness.

To generate the out-of-sample predicted probabilities that Cleanlab requires, we use five-fold cross-validation \citep{kohavi1995study}: we train the model on four folds and predict labels for the remaining fold. Cleanlab then uses these predictions to identify instances where the predicted and actual labels differ strongly \citep{northcutt2021confidentlearning}.

\section{Experiments and results} \label{sec:results}

This section presents our experimental results. We first evaluate how different models classify web registers across our 16-language dataset, comparing monolingual, multilingual, and zero-shot settings (Section \ref{subsec:main_results}), and apply SACX to verify that classification decisions are linguistically motivated. We then compare these approaches in detail and evaluate the CORE taxonomy against the simpler X-GENRE scheme (Section \ref{sec:comparing_results}). Section \ref{subsec:difficult-cases} examines two key challenges in web register classification: documents with uncertain labels and hybrids. Finally, Section \ref{subsec:inference-times} compares model inference times to assess which models are practical at scale.

\subsection{Classification results on the full dataset} 

\subsubsection{Numeric evaluation}
\label{subsec:main_results}

Table \ref{tab:main_results_core} shows classification results for our five main datasets (English, Finnish, French, Swedish, and Turkish). We present results for monolingual, multilingual, and zero-shot settings using different classifier models, evaluating both the full hierarchical taxonomy (N=25) and main categories only (N=9).

\begin{table}
  \caption{Micro ($\mu$) and macro ($M$) F1 scores for the five main language datasets (CORE full taxonomy vs. main labels only).}\label{tab:main_results_core}
  {\tablefont
  \setlength{\tabcolsep}{2.4pt}
  \renewcommand{\arraystretch}{1.3}
  \begin{tabular}{@{\extracolsep{\fill}}lp{8pt}llp{8pt}llp{8pt}llp{8pt}llp{8pt}llp{8pt}ll}
  \hline
   && \multicolumn{2}{c}{En} && \multicolumn{2}{c}{Fi} && \multicolumn{2}{c}{Fr} && \multicolumn{2}{c}{Sv} && \multicolumn{2}{c}{Tr} && \multicolumn{2}{c}{Avg} \\
   \cline{3-4} \cline{6-7} \cline{9-10} \cline{12-13} \cline{15-16} \cline{18-19}
   && All & Main && All & Main && All & Main && All & Main && All & Main && All & Main \\
   \hline
   Monolingual \\
   \hline
   XLM-R & $\mu$ & 74 \tiny{(0.02)} & 76 \tiny{(0.06)} && 79 \tiny{(0.18)} & 82 \tiny{(0.18)} && 72 \tiny{(0.57)} & 75 \tiny{(0.41)} && 79 \tiny{(0.35)} & 81 \tiny{(0.35)} && 75 \tiny{(0.37)} & 76 \tiny{(0.38)} && 76 \tiny{(0.30)} & 78 \tiny{(0.28)} \\
   \hdashline
   & $M$ & 69 \tiny{(0.53)} & 74 \tiny{(0.30)} && 75 \tiny{(0.34)} & 78 \tiny{(4.22)} && 67 \tiny{(1.89)} & 67 \tiny{(0.44)} && 73 \tiny{(0.78)} & 78 \tiny{(0.52)} && 62 \tiny{(2.69)} & 66 \tiny{(0.53)} && 69 \tiny{(1.25)} & 72 \tiny{(1.20)}\\
   \hdashline
   BGE-M3 & $\mu$ & 73 \tiny{(0.03)} & 76 \tiny{(0.08)} && 79 \tiny{(0.23)} & 81 \tiny{(0.33)} && 73 \tiny{(0.18)} & 76 \tiny{(0.05)} && 79 \tiny{(0.25)} & 81 \tiny{(0.08)} && 75 \tiny{(0.97)} & 76 \tiny{(0.75)} && 76 \tiny{(0.33)} & 78 \tiny{(0.26)} \\
   \hdashline
   & $M$ & 69 \tiny{(0.16)} & 74 \tiny{(0.12)} && 73 \tiny{(1.28)} & 73 \tiny{(1.76)} && 70 \tiny{(0.49)} & 71 \tiny{(0.37)} && 72 \tiny{(2.02)} & 75 \tiny{(3.80)} && 64 \tiny{(3.08)} & 69 \tiny{(3.70)} && 70 \tiny{(1.41)} & 72 \tiny{(1.95)} \\
   \hdashline
   BGE-M3$^{\text{2048}}$ & $\mu$ & 74 \tiny{(0.16)} & 76 \tiny{(0.06)} && 80 \tiny{(0.33)} & 82 \tiny{(0.21)} && 74 \tiny{(0.58)} & 77 \tiny{(0.40)} && 79 \tiny{(0.22)} & 81 \tiny{(0.23)} && 75 \tiny{(0.30)} & 75 \tiny{(0.47)} && 76 \tiny{(0.32)} & 78 \tiny{(0.27)} \\
   \hdashline
   & $M$ & 70 \tiny{(0.35)} & 74 \tiny{(0.30)} && 76 \tiny{(1.75)} & 78 \tiny{(2.46)} && 71 \tiny{(1.59)} & 74 \tiny{(1.54)} && 71 \tiny{(0.12)} & 70 \tiny{(1.32)} && 64 \tiny{(2.28)} & 67 \tiny{(0.80)} && 71 \tiny{(1.22)} & 73 \tiny{(1.28)} \\
   \hline
   Multilingual \\
   \hline
   XLM-R & $\mu$ & 72 \tiny{(0.14)} & 75 \tiny{(0.28)} && 79 \tiny{(0.42)} & 82 \tiny{(0.38)} && 75 \tiny{(0.16)} & 78 \tiny{(0.35)} && 81 \tiny{(0.26)} & 82 \tiny{(0.21)} && 78 \tiny{(0.63)} & 78 \tiny{(0.60)} && 77 \tiny{(0.32)} & 79 \tiny{(0.36)} \\
   \hdashline
   &$M$ & 67 \tiny{(0.85)} & 72 \tiny{(0.23)} && 72 \tiny{(0.98)} & 74 \tiny{(2.32)} && 73 \tiny{(0.31)} & 77 \tiny{(0.95)} && 75 \tiny{(0.37)} & 79 \tiny{(0.86)} && 66 \tiny{(1.52)} & 69 \tiny{(1.30)} && 71 \tiny{(0.80)} & 74 \tiny{(1.13)} \\
   \hdashline
   BGE-M3 & $\mu$ & 71 \tiny{(0.11)} & 74 \tiny{(0.13)} && 78 \tiny{(0.25)} & 81 \tiny{(0.26)} && 75 \tiny{(0.07)} & 77 \tiny{(0.06)} && 80 \tiny{(0.04)} & 81 \tiny{(0.17)} && 76 \tiny{(0.63)} & 77 \tiny{(0.50)} && 76 \tiny{(0.22)} & 78 \tiny{(0.22)} \\
   \hdashline
   &$M$ & 66 \tiny{(1.37)} & 70 \tiny{(0.72)} && 73 \tiny{(0.77)} & 73 \tiny{(2.77)} && 70 \tiny{(0.38)} & 74 \tiny{(0.52)} && 75 \tiny{(0.21)} & 77 \tiny{(0.14)} && 64 \tiny{(0.44)} & 65 \tiny{(1.14)} && 69 \tiny{(0.63)} & 72 \tiny{(1.06)} \\
   \hdashline
   BGE-M3$^{\text{2048}}$ & $\mu$ & 72 \tiny{(0.22)} & 74 \tiny{(0.25)} && 79 \tiny{(0.28)} & 82 \tiny{(0.33)} && 76 \tiny{(0.26)} & 79 \tiny{(0.02)} && 81 \tiny{(0.10)} & 82 \tiny{(0.16)} && 78 \tiny{(0.28)} & 78 \tiny{(0.46)} && 77 \tiny{(0.23)} & 79 \tiny{(0.25)} \\
   \hdashline
   &$M$ & 67 \tiny{(0.51)} & 71 \tiny{(0.64)} && 75 \tiny{(1.13)} & 79 \tiny{(0.58)} && 73 \tiny{(0.35)} & 77 \tiny{(0.68)} && 77 \tiny{(0.45)} & 77 \tiny{(0.60)} && 65 \tiny{(0.36)} & 67 \tiny{(0.43)} && 71 \tiny{(0.56)} & 74 \tiny{(0.59)} \\
   \hdashline
   ME5-L & $\mu$ & 71 \tiny{(0.58)} & 74 \tiny{(0.66)} && 78 \tiny{(0.29)} & 81 \tiny{(0.53)} && 76 \tiny{(0.40)} & 78 \tiny{(0.41)} && 80 \tiny{(0.55)} & 82 \tiny{(0.48)} && 76 \tiny{(0.35)} & 76 \tiny{(0.40)} && 76 \tiny{(0.44)} & 78 \tiny{(0.50)} \\
   \hdashline
   &$M$ & 68 \tiny{(0.90)} & 71 \tiny{(0.54)} && 73 \tiny{(2.39)} & 76 \tiny{(2.34)} && 72 \tiny{(1.26)} & 76 \tiny{(2.05)} && 73 \tiny{(2.44)} & 77 \tiny{(1.67)} && 64 \tiny{(0.09)} & 64 \tiny{(2.90)} && 70 \tiny{(1.41)} & 73 \tiny{(1.90)} \\
   \hdashline
   XLMR-XL & $\mu$ & 71 \tiny{(0.19)} & 74 \tiny{(0.34)} && 79 \tiny{(0.29)} & 81 \tiny{(0.18)} && 76 \tiny{(0.48)} & 78 \tiny{(0.71)} && 81 \tiny{(0.14)} & 82 \tiny{(0.30)} && 77 \tiny{(0.35)} & 77 \tiny{(0.69)} && 77 \tiny{(0.29)} & 79 \tiny{(0.44)} \\
   \hdashline
   &$M$ & 67 \tiny{(0.56)} & 70 \tiny{(1.15)} && 73 \tiny{(1.10)} & 75 \tiny{(0.48)} && 72 \tiny{(0.37)} & 75 \tiny{(2.44)} && 75 \tiny{(0.63)} & 78 \tiny{(0.37)} && 65 \tiny{(0.99)} & 67 \tiny{(0.73)} && 70 \tiny{(0.73)} & 73 \tiny{(1.03)} \\
   \hdashline
   Mixtral-8x7B & $\mu$ & 70 \tiny{(0.11)} & 73 \tiny{(0.14)} && 74 \tiny{(1.03)} & 77 \tiny{(1.06)} && 74 \tiny{(0.56)} & 76 \tiny{(0.51)} && 78 \tiny{(0.78)} & 80 \tiny{(0.45)} && 72 \tiny{(0.85)} & 73 \tiny{(0.73)} && 74 \tiny{(0.66)} & 76 \tiny{(0.58)} \\
   \hdashline
   &$M$ & 66 \tiny{(1.64)} & 69 \tiny{(0.31)} && 67 \tiny{(2.69)} & 70 \tiny{(1.57)} && 70 \tiny{(1.09)} & 74 \tiny{(1.02)} && 70 \tiny{(2.57)} & 73 \tiny{(2.09)} && 61 \tiny{(1.20)} & 63 \tiny{(2.08)} && 67 \tiny{(1.84)} & 70 \tiny{(1.41)} \\
   \hline
   Zero-shot \\
   \hline
   XLMR-L & $\mu$ & 64 \tiny{(0.81)} & 67 \tiny{(0.80)} && 72 \tiny{(0.14)} & 73 \tiny{(0.49)} && 72 \tiny{(0.31)} & 75 \tiny{(0.34)} && 78 \tiny{(0.05)} & 79 \tiny{(0.20)} && 68 \tiny{(0.46)} & 67 \tiny{(0.63)} && 70 \tiny{(0.35)} & 72 \tiny{(0.49)} \\
   \hdashline
   &$M$ & 60 \tiny{(1.29)} & 65 \tiny{(0.69)} && 64 \tiny{(1.06)} & 66 \tiny{(0.91)} && 68 \tiny{(0.75)} & 74 \tiny{(0.07)} && 72 \tiny{(1.09)} & 77 \tiny{(0.35)} && 57 \tiny{(0.83)} & 55 \tiny{(1.16)} && 64 \tiny{(1.00)} & 67 \tiny{(0.64)} \\
   \hdashline
   BGE-M3 & $\mu$ & 62 \tiny{(0.93)} & 66 \tiny{(0.66)} && 72 \tiny{(0.73)} & 74 \tiny{(0.45)} && 71 \tiny{(0.57)} & 74 \tiny{(0.60)} && 75 \tiny{(0.09)} & 77 \tiny{(0.16)} && 68 \tiny{(0.15)} & 67 \tiny{(0.21)} && 70 \tiny{(0.50)} & 72 \tiny{(0.41)} \\
   \hdashline
   &$M$ & 58 \tiny{(0.36)} & 62 \tiny{(1.90)} && 64 \tiny{(1.26)} & 67 \tiny{(1.88)} && 66 \tiny{(1.64)} & 73 \tiny{(1.89)} && 70 \tiny{(1.09)} & 73 \tiny{(0.06)} && 55 \tiny{(1.27)} & 52 \tiny{(1.07)} && 62 \tiny{(1.13)} & 66 \tiny{(1.36)} \\
   \hdashline
   BGE-M3$^{\text{2048}}$ & $\mu$ & 63 \tiny{(0.26)} & 66 \tiny{(0.17)} && 72 \tiny{(0.18)} & 75 \tiny{(0.27)} && 72 \tiny{(0.78)} & 75 \tiny{(0.75)} && 76 \tiny{(0.13)} & 79 \tiny{(0.26)} && 68 \tiny{(0.60)} & 68 \tiny{(1.15)} && 70 \tiny{(0.39)} & 73 \tiny{(0.52)} \\
   \hdashline
   &$M$ & 59 \tiny{(1.33)} & 64 \tiny{(0.61)} && 64 \tiny{(1.29)} & 67 \tiny{(2.22)} && 67 \tiny{(0.22)} & 74 \tiny{(1.51)} && 73 \tiny{(0.98)} & 75 \tiny{(1.46)} && 56 \tiny{(1.66)} & 58 \tiny{(3.07)} && 64 \tiny{(1.10)} & 68 \tiny{(1.77)} \\
   \hline
   \end{tabular}
  }{}
  
\end{table}

Monolingual models are trained and tested on individual languages. Multilingual models are trained on all five languages combined and tested separately on each language. For zero-shot evaluation, we train on four languages and test on the fifth excluded language. We evaluate all models from Section \ref{subsec:classifiers} in the multilingual setting, but focus on XLM-R Large and BGE M3 for monolingual and zero-shot experiments.

Monolingual and multilingual models achieve the strongest performance: monolingual models reach 76\% micro F1 averaged across the full label scheme, while multilingual models reach 77\%. Zero-shot models perform 6-7 percentage points lower, averaging 70\% micro F1 across languages. The performance gap between the full taxonomy and main categories is surprisingly small---the simpler scheme achieves only 1--2 percentage points higher micro F1 scores, reaching 76--79\%. This small gap suggests that the CORE subregister classes are sufficiently distinctive that adding them does not substantially increase classification difficulty.

Micro-averaged F1 scores consistently exceed macro-averaged scores by 5--7 percentage points. The highest macro scores are 71\% for the full scheme and 74\% for the main labels (multilingual XLM-R and BGE-M3\textsuperscript{2048}). The many small register classes in our 25-class CORE taxonomy make it difficult to achieve high macro F1 scores---which weights each register class equally---since poor performance on even a few small classes significantly impacts the macro average (see also Section \ref{subsec:register-taxonomy}). Our macro F1 scores of 71\% therefore demonstrate strong overall performance.

Most classifiers perform similarly, except Mixtral, which scores consistently lower. This likely stems from Mixtral being the only generative (decoder) model \citep{muennighoff2022sgpt} and our use of 4-bit quantization. BGE-M3, ME5, and both XLM-R models score very similarly, despite their architectural differences: longer token limits (BGE-M3), contrastive learning (ME5), and larger model size (XLM-R XL). This suggests that these architectural improvements do not enhance classifier performance, and that these datasets have a performance ceiling around 80\% F1.

Statistical significance testing via paired bootstrap (Appendix, Tables \ref{tab:app-monoling}--\ref{tab:multi-sv-tr}) confirms that Mixtral's lower performance is significant across nearly all languages and settings, whereas the differences among the top models (XLM-R Large, BGE-M3, ME5, XLM-R XL) are often negligible. For instance, in the multilingual setting, XLM-R Large significantly outperforms BGE-M3 for English but not for Finnish or Swedish, despite small numerical differences. This suggests that model selection among top performers can be based on practical considerations such as inference speed (Section \ref{subsec:inference-times}) rather than accuracy alone.

Table \ref{tab:zeroshot_results} shows zero-shot results for the 11 evaluation-only languages. The average micro F1 score across all labels is 66\%, slightly below the zero-shot scores for the main languages. Performance varies widely by language, from 50-55\% micro F1 for Japanese to 70--80\% for Farsi, Hindi, and Urdu. We observe the same patterns as in the main languages: Mixtral scores 2--4 percentage points below other models (significantly worse in most cases; see Appendix Tables \ref{tab:transfer_ar-ca-fa}--\ref{tab:transfer_ur-zh}), and the main labels perform best, reaching 70\% F1 across languages. Significance testing reveals that for evaluation-only languages, model rankings vary considerably by language, with no single model consistently and significantly outperforming others, except for Mixtral's consistently lower performance. 

While still competitive, zero-shot models score lower than mono- and multilingual approaches. These language differences stem from factors such as varying dataset sizes (see Table \ref{tbl:corpora}), annotator variations, linguistic relationships (e.g., Norwegian and Swedish are closely related), and how well each language is represented in pre-training (see Section \ref{subsubsec:crossling_learning}). The results also suggest that registers have language-specific features that cannot be fully captured without in-language training data.

\begin{table}
\scriptsize
\caption{Micro ($\mu$) and macro ($M$) zero-shot F1 scores for 11 evaluation languages, using a multilingual model fine-tuned with the five major language datasets, using different models.}\label{tab:zeroshot_results}
  {\tablefont
\setlength{\tabcolsep}{3.5pt}
\renewcommand{\arraystretch}{1.3}
\begin{tabular}{llllllllllllll}
\hline
& & Ar & Ca & Es & Fa & Hi & Id & Jp & No & Pt & Ur & Zh & Avg. \\
\hline
& & \multicolumn{12}{l}{All labels}  \\
\hline
 XLMR-L & $\mu$ &64 \tiny{(0.20)} & 61 \tiny{(0.75)} & 61 \tiny{(1.01)} & 71 \tiny{(0.75)} & 79 \tiny{(0.01)} & 61 \tiny{(0.24)} & 51 \tiny{(3.06)} & 65 \tiny{(0.07)} & 67 \tiny{(0.70)} & 82 \tiny{(0.39)} & 66 \tiny{(0.63)} & 66 \tiny{(0.71)} \\
 \hdashline
 &$M$ & 64 \tiny{(1.06)} & 67 \tiny{(1.68)} & 62 \tiny{(0.24)} & 69 \tiny{(2.66)} & 70 \tiny{(3.79)} & 52 \tiny{(2.15)} & 60 \tiny{(0.70)} & 64 \tiny{(3.53)} & 58 \tiny{(0.33)} & 62 \tiny{(0.89)} & 53 \tiny{(1.89)} & 62 \tiny{(1.72)} \\
\hdashline
 BGE-M3 & $\mu$ &64 \tiny{(0.50)} & 62 \tiny{(0.63)} & 62 \tiny{(0.85)} & 70 \tiny{(1.07)} & 78 \tiny{(0.31)} & 61 \tiny{(0.33)} & 50 \tiny{(2.04)} & 65 \tiny{(0.98)} & 68 \tiny{(0.85)} & 80 \tiny{(0.76)} & 66 \tiny{(0.53)} & 66 \tiny{(0.80)} \\
 \hdashline
 &$M$ & 63 \tiny{(0.45)} & 66 \tiny{(1.63)} & 63 \tiny{(0.75)} & 70 \tiny{(1.10)} & 66 \tiny{(1.00)} & 53 \tiny{(0.63)} & 60 \tiny{(2.05)} & 65 \tiny{(0.18)} & 56 \tiny{(2.06)} & 62 \tiny{(2.22)} & 54 \tiny{(0.39)} & 62 \tiny{(1.13)} \\
\hdashline
 BGE-M3$^{\text{2048}}$ & $\mu$ &65 \tiny{(0.34)} & 62 \tiny{(0.40)} & 62 \tiny{(0.36)} & 70 \tiny{(0.27)} & 79 \tiny{(0.31)} & 61 \tiny{(0.06)} & 54 \tiny{(0.96)} & 64 \tiny{(0.16)} & 68 \tiny{(0.21)} & 81 \tiny{(0.36)} & 67 \tiny{(0.80)} & 67 \tiny{(0.38)} \\
 \hdashline
 &$M$ & 66 \tiny{(1.47)} & 69 \tiny{(0.77)} & 61 \tiny{(1.58)} & 72 \tiny{(0.86)} & 66 \tiny{(1.41)} & 55 \tiny{(0.75)} & 62 \tiny{(2.01)} & 65 \tiny{(0.91)} & 56 \tiny{(1.30)} & 61 \tiny{(2.15)} & 57 \tiny{(0.78)} & 63 \tiny{(1.27)} \\
\hdashline
 ME5-L & $\mu$ &65 \tiny{(1.19)} & 63 \tiny{(0.72)} & 60 \tiny{(1.22)} & 71 \tiny{(1.43)} & 80 \tiny{(1.18)} & 60 \tiny{(0.24)} & 51 \tiny{(2.62)} & 65 \tiny{(1.09)} & 66 \tiny{(0.75)} & 83 \tiny{(0.74)} & 65 \tiny{(0.31)} & 66 \tiny{(1.04)} \\
 \hdashline
 &$M$ & 62 \tiny{(1.16)} & 68 \tiny{(0.79)} & 59 \tiny{(2.84)} & 70 \tiny{(1.50)} & 68 \tiny{(1.91)} & 54 \tiny{(0.94)} & 63 \tiny{(2.90)} & 62 \tiny{(0.42)} & 58 \tiny{(0.75)} & 64 \tiny{(0.56)} & 57 \tiny{(1.64)} & 62 \tiny{(1.40)} \\
\hdashline
 XLMR-XL & $\mu$ &64 \tiny{(0.38)} & 62 \tiny{(1.10)} & 62 \tiny{(0.70)} & 72 \tiny{(0.96)} & 77 \tiny{(0.90)} & 60 \tiny{(0.25)} & 55 \tiny{(1.00)} & 64 \tiny{(0.43)} & 66 \tiny{(0.81)} & 81 \tiny{(0.45)} & 67 \tiny{(0.45)} & 66 \tiny{(0.67)} \\
 \hdashline
 &$M$ & 64 \tiny{(3.46)} & 67 \tiny{(1.55)} & 62 \tiny{(1.79)} & 72 \tiny{(3.65)} & 66 \tiny{(2.32)} & 53 \tiny{(2.04)} & 62 \tiny{(1.37)} & 64 \tiny{(1.58)} & 57 \tiny{(1.20)} & 61 \tiny{(1.76)} & 54 \tiny{(1.09)} & 62 \tiny{(1.98)} \\
\hdashline
 Mixtral-8x7B & $\mu$ &59 \tiny{(1.42)} & 62 \tiny{(0.15)} & 61 \tiny{(0.34)} & 67 \tiny{(2.73)} & 74 \tiny{(1.38)} & 58 \tiny{(1.61)} & 54 \tiny{(3.87)} & 65 \tiny{(0.61)} & 64 \tiny{(1.01)} & 76 \tiny{(3.32)} & 61 \tiny{(0.69)} & 64 \tiny{(1.56)} \\
 \hdashline
 &$M$ & 60 \tiny{(4.49)} & 67 \tiny{(0.22)} & 63 \tiny{(1.96)} & 64 \tiny{(3.24)} & 64 \tiny{(1.13)} & 50 \tiny{(1.99)} & 62 \tiny{(5.99)} & 65 \tiny{(1.99)} & 55 \tiny{(0.28)} & 61 \tiny{(3.33)} & 50 \tiny{(0.64)} & 60 \tiny{(2.30)} \\
\hline
& & \multicolumn{12}{l}{Main labels}  \\
\hline
 XLMR-L & $\mu$ &66 \tiny{(0.42)} & 63 \tiny{(0.71)} & 67 \tiny{(0.37)} & 71 \tiny{(1.62)} & 79 \tiny{(0.03)} & 62 \tiny{(0.10)} & 61 \tiny{(1.73)} & 70 \tiny{(0.12)} & 69 \tiny{(0.33)} & 84 \tiny{(0.19)} & 70 \tiny{(0.81)} & 69 \tiny{(0.59)} \\
 \hdashline
 &$M$ & 59 \tiny{(0.38)} & 67 \tiny{(4.06)} & 50 \tiny{(2.34)} & 69 \tiny{(4.06)} & 65 \tiny{(1.12)} & 53 \tiny{(3.32)} & 59 \tiny{(3.11)} & 66 \tiny{(0.86)} & 57 \tiny{(0.82)} & 65 \tiny{(0.21)} & 53 \tiny{(1.45)} & 60 \tiny{(1.98)} \\
\hdashline
 BGE-M3 & $\mu$ &67 \tiny{(1.04)} & 64 \tiny{(0.74)} & 69 \tiny{(0.52)} & 70 \tiny{(1.24)} & 77 \tiny{(0.70)} & 62 \tiny{(0.41)} & 62 \tiny{(1.42)} & 69 \tiny{(0.94)} & 70 \tiny{(0.91)} & 82 \tiny{(1.12)} & 70 \tiny{(0.67)} & 69 \tiny{(0.88)} \\
 \hdashline
 &$M$ & 59 \tiny{(1.00)} & 64 \tiny{(0.77)} & 50 \tiny{(1.54)} & 69 \tiny{(0.96)} & 62 \tiny{(2.77)} & 52 \tiny{(3.15)} & 59 \tiny{(0.71)} & 66 \tiny{(0.39)} & 60 \tiny{(1.01)} & 67 \tiny{(1.47)} & 51 \tiny{(0.38)} & 60 \tiny{(1.29)} \\
\hdashline
 BGE-M3$^{\text{2048}}$ & $\mu$ &68 \tiny{(0.56)} & 63 \tiny{(0.56)} & 69 \tiny{(0.76)} & 70 \tiny{(0.98)} & 79 \tiny{(0.67)} & 62 \tiny{(0.12)} & 66 \tiny{(0.50)} & 68 \tiny{(0.48)} & 71 \tiny{(0.35)} & 82 \tiny{(0.19)} & 71 \tiny{(0.81)} & 70 \tiny{(0.54)} \\
 \hdashline
 &$M$ & 60 \tiny{(1.11)} & 66 \tiny{(2.70)} & 49 \tiny{(2.44)} & 69 \tiny{(1.41)} & 62 \tiny{(0.56)} & 57 \tiny{(1.05)} & 64 \tiny{(1.85)} & 66 \tiny{(1.29)} & 58 \tiny{(2.62)} & 62 \tiny{(0.31)} & 54 \tiny{(1.61)} & 61 \tiny{(1.54)} \\
\hdashline
 ME5-L & $\mu$ &68 \tiny{(1.44)} & 64 \tiny{(0.42)} & 67 \tiny{(1.59)} & 72 \tiny{(0.68)} & 81 \tiny{(1.35)} & 61 \tiny{(0.52)} & 63 \tiny{(2.62)} & 70 \tiny{(1.37)} & 68 \tiny{(0.96)} & 84 \tiny{(0.71)} & 69 \tiny{(0.37)} & 70 \tiny{(1.09)} \\
 \hdashline
 &$M$ & 59 \tiny{(2.21)} & 65 \tiny{(1.67)} & 54 \tiny{(1.90)} & 68 \tiny{(1.82)} & 72 \tiny{(0.55)} & 56 \tiny{(0.73)} & 62 \tiny{(3.55)} & 66 \tiny{(1.36)} & 59 \tiny{(1.49)} & 64 \tiny{(2.69)} & 59 \tiny{(3.25)} & 62 \tiny{(1.93)} \\
\hdashline
 XLMR-XL & $\mu$ &68 \tiny{(1.02)} & 63 \tiny{(0.87)} & 69 \tiny{(0.93)} & 71 \tiny{(1.35)} & 77 \tiny{(0.98)} & 60 \tiny{(0.07)} & 66 \tiny{(1.60)} & 68 \tiny{(0.19)} & 68 \tiny{(0.55)} & 81 \tiny{(0.30)} & 71 \tiny{(0.39)} & 69 \tiny{(0.75)} \\
 \hdashline
 &$M$ & 59 \tiny{(1.31)} & 64 \tiny{(1.76)} & 53 \tiny{(1.57)} & 68 \tiny{(0.75)} & 64 \tiny{(6.15)} & 52 \tiny{(2.76)} & 63 \tiny{(1.19)} & 64 \tiny{(0.41)} & 59 \tiny{(1.91)} & 59 \tiny{(2.64)} & 54 \tiny{(1.03)} & 60 \tiny{(1.95)} \\
\hdashline
 Mixtral-8x7B & $\mu$ &60 \tiny{(2.69)} & 64 \tiny{(0.71)} & 67 \tiny{(0.76)} & 68 \tiny{(2.46)} & 73 \tiny{(1.31)} & 59 \tiny{(2.09)} & 64 \tiny{(3.72)} & 69 \tiny{(1.21)} & 67 \tiny{(1.26)} & 76 \tiny{(4.13)} & 66 \tiny{(1.25)} & 67 \tiny{(1.96)} \\
 \hdashline
 &$M$ & 51 \tiny{(2.61)} & 68 \tiny{(2.57)} & 50 \tiny{(3.25)} & 62 \tiny{(3.63)} & 59 \tiny{(1.76)} & 47 \tiny{(1.16)} & 61 \tiny{(9.58)} & 63 \tiny{(1.33)} & 57 \tiny{(1.95)} & 60 \tiny{(2.15)} & 51 \tiny{(2.26)} & 57 \tiny{(2.93)} \\
\hline
\end{tabular}
}{}
\end{table}

Overall, our multilingual models achieve 77\% average micro F1 across the five main languages---matching the best results for Finnish CORE \citep{skantsi_laippala_2023} and significantly outperforming previous state-of-the-art English CORE results of 68\% F1 \citep{laippala-etal-2022-towards}. These results compare well with the 80\% F1 that \citet{Kuzman.Ljubesic2023} achieved on the simpler X-GENRE scheme, which uses nine classes and excludes hybrids. Although the detailed CORE taxonomy makes classification of small classes more challenging, our macro F1 of 71\% with the full scheme shows robust performance even with 25 classes. Importantly, this robust performance makes the classifier more valuable for practical applications---for example, \citet{eskelinen-etal-2024-building} leveraged this capability when building question-answer datasets from web crawls.

\subsubsection{Linguistic motivation behind the classification}

\label{subsubsec:sacx_application}

A common challenge in deep learning is the lack of transparency in classifier decisions \citep{sogaard-explainable}. In web register and genre identification, understanding the linguistic basis of classifications is critical to ensure that models have learned the intended task \citep{petrenz-webber,lepekhin-sharoff-adversarial,laippala2021exploring}. To this end, we apply SACX to analyze the top 10 keywords learned by XLM-R Large for two registers across three languages: \textit{Interactive Discussion} and \textit{Spoken} in English, Finnish, and French (see Section \ref{subsec:SACX} for implementation details).

Table \ref{tab:sacx} presents the keywords. For \textit{Interactive Discussion}, the keywords reflect typical forum and question-answer activities: \textit{answers, question, keskustelut} `discussions', \textit{ketjun} `of the thread', and \textit{forum(s)}. For \textit{Spoken}, which includes interviews and official speeches, the keywords refer to interviews, questions, and parliamentary settings where such speeches occur.

For both registers, most top keywords relate strongly to their respective classes, aligning with previous findings \citep{Biber_Egbert_2018}. Although these keywords represent only a small fraction of what the classifier has learned, they corroborate our numeric evaluation by showing that the models focus on linguistically appropriate features. However, some keywords appear less clearly related to their registers, such as \textit{cycling, jpg} in \textit{Interactive Discussion} and \textit{philosophy, turvallisuuden} `of-safety' in \textit{Spoken}. These likely reflect document-specific topics that do not generalize to entire classes. This pattern is consistent with SACX's known behavior of identifying proper nouns and topic-specific terms that distinguish document subsets within classes \citep{ronnqvist-etal-2022-SACX}.

\begin{table}[]
\footnotesize
\caption{Keywords extracted with the SACX-method for English, Finnish, and French. Translations by us.}
{\tablefont
\setlength{\tabcolsep}{3.5pt}
\renewcommand{\arraystretch}{0.8}
\begin{tabular}{@{}lll@{}}
\toprule

En                            & Fi                                & Fr                             \\ \midrule
Interactive discussion \\ \midrule
answers           & kysyjä (\scriptsize asker)                  & blog                          \\
re               & keskustelut (\scriptsize discussions)        & discussion                    \\
question         & ketjuun (\scriptsize to thread)             & forums                        \\
wyoo             & keskustelupari (\scriptsize forum name) & a-t-il (\scriptsize has he/she/it) \\
thread           & ketjussa (\scriptsize in thread)            & forum                         \\
rescue          & re                                & re                             \\
forums;         & ketjun (\scriptsize of thread)              & topic                          \\
ow              & kommentista (\scriptsize of comment)        & jpg                            \\
cycling         & faq                               & acheter (\scriptsize to buy)  \\
topic           & keskustelu (\scriptsize discussion)         & membre (\scriptsize member)   \\ \midrule
Spoken                            &                                    &                                \\ \midrule
q\&a            & puhemies (\scriptsize chairman)             & interview                     \\
speech          & herra (\scriptsize sir/mr.)                 & quels (\scriptsize which)     \\
interview       & turvallisuuden (\scriptsize of safety)         & as-tu (\scriptsize do you have) \\
communicated    & arvoisa (\scriptsize honorable)             & entretien (\scriptsize interview) \\
interviews      & olet (\scriptsize you are)                  & a (\scriptsize he/she/it has) \\
interviewed     & esitys (\scriptsize presentation)           & monsieur (\scriptsize mr.)    \\
communicate     & me (\scriptsize we)                         & comment (\scriptsize how)     \\
presenter       & en (\scriptsize I do not)                   & qu’il (\scriptsize that he/it) \\
philosophy      & miksi (\scriptsize why)                     & oui (\scriptsize yes)         \\
architecture    & eduskunnassa (\scriptsize in the parliament) & où (\scriptsize where)        \\ \midrule
\end{tabular}
}
\label{tab:sacx}
\end{table}

\subsection{Comparing training settings and schemes}
\label{sec:comparing_results}

As shown in the previous section, our models achieve competitive results but appear to hit a performance ceiling: multilingual models show only marginal improvements over monolingual ones despite additional training data, and various architectural improvements yield similar results. In this section, we investigate the reasons behind this ceiling by comparing monolingual and multilingual models in detail. We also test whether the X-GENRE taxonomy, which achieves higher reported results than the CORE scheme \citep{Kuzman.Ljubesic2023}, can break through this ceiling.

\subsubsection{Comparing monolingual, multilingual and zero-shot models across registers}
\label{subsubsec:crossling_learning}

A closer examination of the language-specific scores in Table \ref{tab:main_results_core} shows that the effect of multilingual learning varies by language, even though average performance improves by only 1--2 percentage points. Scores range from a 2-point decrease (XLM-R on English, micro and macro F1) to a 6-point improvement (XLM-R on Turkish, macro F1). We find that this variation stems from two factors: how the language-specific datasets were annotated and how large they are.

English is the only language that experiences a performance decrease with multilingual training, dropping from 74\% to 72\% micro F1. For Finnish, micro F1 remains 79\% in both settings, while for French, Swedish, and Turkish, the multilingual models outperform the monolingual models by 1–3 points (XLM-R scores increasing from 72--79\% to 75--81\% micro F1).

The small decrease in English performance in the multilingual setting likely results from differences in annotation practices between English and the other languages. Although \citet{repo2021zeroshot} reported promising zero-shot transfer results from English to French, Finnish, and Swedish, suggesting good corpus compatibility, the use of MTurk for the English CORE corpus created annotation inconsistencies (see Section \ref{subsec:corpus-compilation}).

\begin{figure}
  \centering
  \begin{minipage}{0.8\textwidth}
   \includegraphics[width=\textwidth]{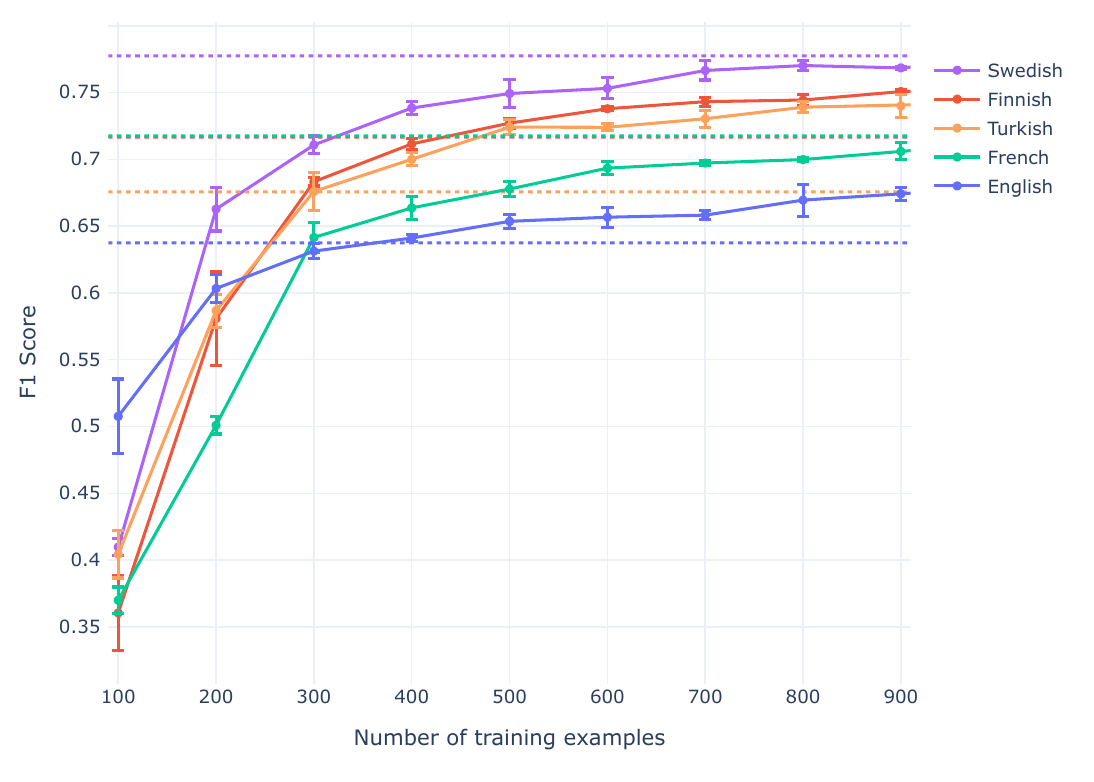}
  \vspace{0.5em}
  \caption{F1 scores as a function of training steps. Standard deviations presented as error bars and zero-shot performances as dotted lines.}
  \label{fig:learning_curves}
  \end{minipage}
\end{figure}

The improvements for Turkish, French, and Swedish in the multilingual setting result from the additional training examples provided by other languages. In contrast, Finnish shows no improvement because its dataset is already large enough. This becomes clear in the learning curve in Figure \ref{fig:learning_curves}, where the model reaches high performance after just 1,000 documents. Since FinCORE contains 10,000 documents, Finnish does not benefit from multilingual training while the smaller French, Swedish, and Turkish datasets do. \citet{laippala_ronnqvist2023} found similar results when studying how CORE classifier performance relates to training data size---learning slows after 5,000 documents and stops after 10,000.

Data saturation thus explains why mono- and multilingual models score similarly overall. Languages with less training data benefit more from multilingual training, especially for their smaller classes. This is evident in the macro-average F1 scores when comparing monolingual and multilingual results in Table \ref{tab:main_results_core}. For example, with multilingual XLM-R, the French macro-average F1 score rises from 67\% to 73\%, and Turkish from 62\% to 66\%.

\begin{figure}
  \centering
  \begin{minipage}{0.75\textwidth}
  \includegraphics[width=\textwidth]{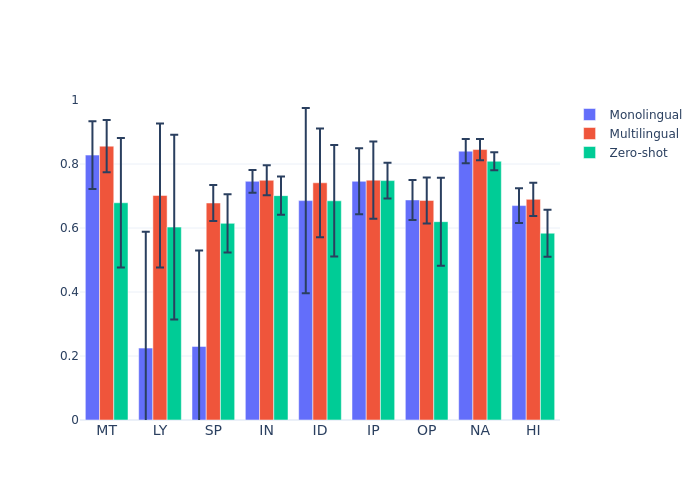}
  \caption{Register-specific micro F1 scores in monolingual, multilingual and zero-shot settings for \textit{Machine-translated} (MT), \textit{Lyrical} (LY), \textit{Spoken} (SP), \textit{Informational description} (IN), \textit{Interactive discussion} (ID), \textit{Informational Persuasion} (IP),\textit{Opinion} (OP), \textit{Narrative} (NA) and \textit{How-to or instructions} (HI).}
  \label{fig:register_bar_chart}
\end{minipage} 
\end{figure}

Figure \ref{fig:register_bar_chart} compares monolingual, multilingual, and zero-shot performance across the five major languages, broken down by main register. Performance varies considerably across classes: \textit{Machine translation} and \textit{Narrative} achieve the highest scores, while \textit{Lyrical} and \textit{Spoken} achieve the lowest. \textit{How-to Instruction} and \textit{Opinion} also perform below average. This variation mirrors findings from previous web register identification studies \citep{laippala-etal-2019-toward,biber-egbert2016-using-grammatical-feats} and reflects differences in register distinctiveness. Some registers like \textit{Machine translation} or \textit{Encyclopedia article} have obvious, well-defined characteristics, while others are less distinctive and harder to identify.

The three training settings also affect registers differently. Most registers---especially the larger ones---score slightly higher with multilingual training than monolingual, while zero-shot scores slightly lower. \textit{Lyrical} and \textit{Spoken} show a different pattern: monolingual models perform very poorly, while multilingual and zero-shot models perform much better. Smaller languages have too few training examples for these classes to train effective monolingual models. As a result, multilingual or even zero-shot approaches work particularly well for these registers, even though the models perform similarly overall.

\subsubsection{Comparing the CORE scheme with X-GENRE}

\label{subsec:coreVSxgenre}

\citet{Kuzman-genre-identification-2023} compare versions of the CORE and X-GENRE taxonomies and find that X-GENRE performs better (see Section \ref{subsec:core2xgenre}). Specifically, using XLM-R Base and a CORE-based scheme with 47 labels, they achieve a micro F1 score of 66\%, compared to 80\% F1 when they use X-GENRE with nine labels. The relatively low score for CORE likely stems from three main factors. First, \citet{Kuzman-genre-identification-2023} use the Base variant of XLM-R, which performs worse on register identification than the Large variant \citep{laippala_ronnqvist2023}. Second, they treat register identification as a multi-class task, assigning each text to a single category; however, as shown by \citet{Madjarov.etal2019} and \citet{laippala_ronnqvist2023}, a multi-label setting better captures register representations. Third, \citet{Kuzman-genre-identification-2023} use a CORE version with 47 registers (vs. our 25), making the classification task more challenging.

\begin{table}
  \caption{Micro ($\mu$) and macro ($M$) F1 scores for the five main language datasets using the X-GENRE labels.}\label{tab:main_results_xgenre}
  {\tablefont
  \setlength{\tabcolsep}{2.4pt}
  \renewcommand{\arraystretch}{1.3}
  \begin{tabular}{@{\extracolsep{\fill}}lllllllll}
  \hline
   && \multicolumn{6}{l}{Monolingual} \\
   \hline
   && En & Fi & Fr & Sv & Tr & Avg. \\
   \hline
 XLMR-L & $\mu$ &74 \tiny{(0.22)} & 78 \tiny{(0.23)} & 73 \tiny{(0.92)} & 78 \tiny{(0.27)} & 74 \tiny{(0.95)} & 75 \tiny{(0.52)} \\
  \hdashline
 &$M$ & 68 \tiny{(0.64)} & 75 \tiny{(0.80)} & 69 \tiny{(1.81)} & 70 \tiny{(0.39)} & 62 \tiny{(2.08)} & 69 \tiny{(1.14)} \\
 \hdashline
 BGE-M3 & $\mu$ &73 \tiny{(0.08)} & 78 \tiny{(0.18)} & 73 \tiny{(0.55)} & 79 \tiny{(0.13)} & 74 \tiny{(1.60)} & 75 \tiny{(0.51)} \\
  \hdashline
 &$M$ & 69 \tiny{(0.60)} & 75 \tiny{(0.08)} & 71 \tiny{(1.08)} & 71 \tiny{(0.25)} & 63 \tiny{(3.27)} & 70 \tiny{(1.06)} \\
 \hdashline
 BGE-M3$^{\text{2048}}$ & $\mu$ &74 \tiny{(0.18)} & 79 \tiny{(0.48)} & 75 \tiny{(0.45)} & 79 \tiny{(0.44)} & 74 \tiny{(0.33)} & 76 \tiny{(0.38)} \\
  \hdashline
 &$M$ & 70 \tiny{(0.04)} & 76 \tiny{(0.58)} & 72 \tiny{(0.47)} & 73 \tiny{(0.58)} & 65 \tiny{(1.34)} & 71 \tiny{(0.60)} \\
 \hline
 && \multicolumn{6}{l}{Multilingual} \\
 \hline
 XLMR-L & $\mu$ &71 \tiny{(0.36)} & 78 \tiny{(0.59)} & 76 \tiny{(0.72)} & 81 \tiny{(0.25)} & 77 \tiny{(0.76)} & 77 \tiny{(0.53)} \\
  \hdashline
 &$M$ & 67 \tiny{(0.21)} & 75 \tiny{(0.57)} & 74 \tiny{(1.25)} & 76 \tiny{(0.44)} & 71 \tiny{(1.57)} & 73 \tiny{(0.81)} \\
 \hdashline
 BGE-M3 & $\mu$ &70 \tiny{(0.22)} & 77 \tiny{(0.56)} & 74 \tiny{(0.32)} & 80 \tiny{(0.04)} & 76 \tiny{(0.33)} & 75 \tiny{(0.29)} \\
 \hdashline
 &$M$ & 65 \tiny{(0.24)} & 73 \tiny{(1.16)} & 71 \tiny{(0.47)} & 75 \tiny{(0.30)} & 67 \tiny{(0.41)} & 70 \tiny{(0.52)} \\
 \hdashline
 BGE-M3$^{\text{2048}}$ & $\mu$ &71 \tiny{(0.41)} & 78 \tiny{(0.24)} & 76 \tiny{(0.18)} & 80 \tiny{(0.21)} & 77 \tiny{(0.04)} & 77 \tiny{(0.22)} \\
 \hdashline
 &$M$ & 66 \tiny{(0.71)} & 75 \tiny{(0.76)} & 74 \tiny{(0.13)} & 76 \tiny{(0.41)} & 69 \tiny{(0.49)} & 72 \tiny{(0.50)} \\
 \hdashline
 ME5-L & $\mu$ &71 \tiny{(0.46)} & 77 \tiny{(0.53)} & 76 \tiny{(0.47)} & 80 \tiny{(0.59)} & 76 \tiny{(0.51)} & 76 \tiny{(0.51)} \\
 \hdashline
 &$M$ & 67 \tiny{(1.10)} & 74 \tiny{(0.51)} & 73 \tiny{(1.18)} & 76 \tiny{(0.77)} & 69 \tiny{(0.09)} & 72 \tiny{(0.73)} \\
 \hdashline
 XLMR-XL & $\mu$ &71 \tiny{(0.23)} & 78 \tiny{(0.52)} & 76 \tiny{(0.60)} & 81 \tiny{(0.21)} & 76 \tiny{(0.59)} & 76 \tiny{(0.43)} \\
 \hdashline
 &$M$ & 65 \tiny{(0.35)} & 75 \tiny{(0.58)} & 73 \tiny{(1.03)} & 75 \tiny{(0.45)} & 69 \tiny{(0.32)} & 72 \tiny{(0.55)} \\
 \hdashline
 Mixtral-8x7B & $\mu$ &69 \tiny{(0.67)} & 72 \tiny{(0.64)} & 73 \tiny{(0.95)} & 77 \tiny{(0.72)} & 71 \tiny{(0.89)} & 73 \tiny{(0.77)} \\
 \hdashline
 &$M$ & 64 \tiny{(0.10)} & 70 \tiny{(0.91)} & 71 \tiny{(0.43)} & 72 \tiny{(0.80)} & 62 \tiny{(2.86)} & 68 \tiny{(1.02)} \\
 \hline
 && \multicolumn{6}{l}{Zero-shot} \\
 \hline
 XLMR-L & $\mu$ &64 \tiny{(0.42)} & 70 \tiny{(0.28)} & 71 \tiny{(0.47)} & 78 \tiny{(0.09)} & 67 \tiny{(0.54)} & 70 \tiny{(0.36)} \\
  \hdashline
 &$M$ & 60 \tiny{(0.34)} & 67 \tiny{(0.84)} & 70 \tiny{(0.34)} & 73 \tiny{(0.33)} & 57 \tiny{(0.75)} & 65 \tiny{(0.52)} \\
 \hdashline
 BGE-M3 & $\mu$ &62 \tiny{(1.50)} & 71 \tiny{(1.11)} & 71 \tiny{(0.49)} & 74 \tiny{(0.38)} & 67 \tiny{(0.46)} & 69 \tiny{(0.79)} \\
 \hdashline
 &$M$ & 57 \tiny{(1.53)} & 67 \tiny{(1.13)} & 68 \tiny{(0.41)} & 70 \tiny{(0.46)} & 56 \tiny{(1.20)} & 64 \tiny{(0.95)} \\
 \hdashline
 BGE-M3$^{\text{2048}}$ & $\mu$ &63 \tiny{(0.22)} & 71 \tiny{(0.25)} & 71 \tiny{(0.99)} & 75 \tiny{(0.27)} & 68 \tiny{(1.28)} & 70 \tiny{(0.60)} \\
 \hdashline
 &$M$ & 59 \tiny{(0.30)} & 67 \tiny{(0.51)} & 69 \tiny{(0.93)} & 71 \tiny{(1.01)} & 57 \tiny{(2.07)} & 65 \tiny{(0.96)} \\
   \hline
   \end{tabular}
  }
\end{table}

We evaluate the performance of the X-GENRE scheme by mapping our CORE labels to X-GENRE labels (see Section \ref{subsec:core2xgenre} for the mapping). For fair comparison, we fine-tune the same set of models as in the CORE evaluations (Section \ref{subsec:main_results}) using the same multi-label setting across monolingual, multilingual, and zero-shot configurations.

Table \ref{tab:main_results_xgenre} shows the results. The X-GENRE labels achieve scores very similar to our 25-label CORE scheme, reaching 77\% micro F1 across languages in the multilingual setting, 75\% for monolingual training, and 70\% for zero-shot. Macro F1 scores differ by only 0--1 percentage points from our CORE evaluations. These results suggest that X-GENRE is a viable classification scheme that performs on par with our full hierarchical CORE taxonomy. However, X-GENRE contains fewer labels and is thus less expressive for analyzing register variation on the web. Moreover, it performs slightly worse than the main nine-label CORE scheme while providing similar granularity.

\subsection{Breaking the performance ceiling by targeting uncertainty and hybrids} \label{subsec:difficult-cases} 

In the previous section, we showed that multilingual training is particularly beneficial for smaller languages and that the detailed hierarchical CORE taxonomy with multi-label classification outperforms the multi-class X-GENRE. However, neither finding helped us break the performance ceiling identified in Section \ref{subsec:main_results}.

In this section, we attempt to improve classification scores by focusing on two key challenges in web register classification: documents with uncertain labels and hybrid documents. First, we experiment with data pruning by removing examples with potential labeling issues. Then, we analyze hybrids by running classification experiments on hybrid and non-hybrid data separately. Throughout these experiments, we use the multilingual XLM-R classifier, which combines strong performance with fast inference (see Section \ref{subsec:inference-times}).

\subsubsection{Data pruning} \label{subsubsec:pruning-experiments}

As discussed in Section \ref{sec:background}, one of the challenges of web register identification is that many documents lack clear register characteristics. Some documents mix features from multiple registers to form hybrids, show only vague characteristics of a broad class, or do not fit any register at all. Here, we test how removing documents with uncertain labels affects classifier performance. These uncertain labels stem from various issues: ambiguous register assignments, outliers, and potential mislabeling.

We use the Cleanlab tool (see Section \ref{subsec:datacleaning}) to identify and remove documents with uncertain labels, expecting this pruning to improve model performance. We then compare models trained on the full unfiltered data with those trained on pruned data. We evaluate using three datasets: (a) the original full dataset without pruning; (b) instances Cleanlab marked as uncertain (24\% of the original); and (c) a pruned set without uncertain instances (76\% of the original). For this analysis, we train models on the full (a) and pruned (c) sets, splitting them as described in Section \ref{preprocessing}, and evaluate these models on all three datasets.

\begin{figure}
  \begin{center}
    \includegraphics[height=160pt]{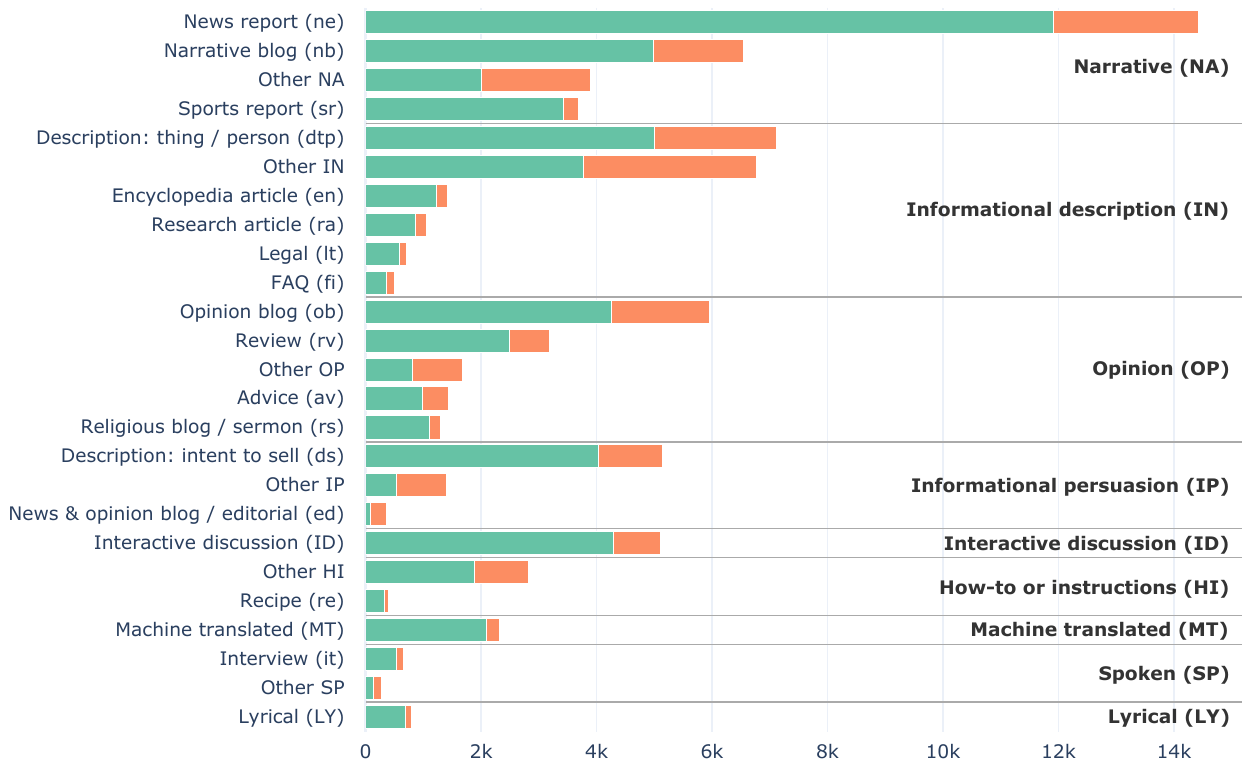}
    \includegraphics[height=160pt]{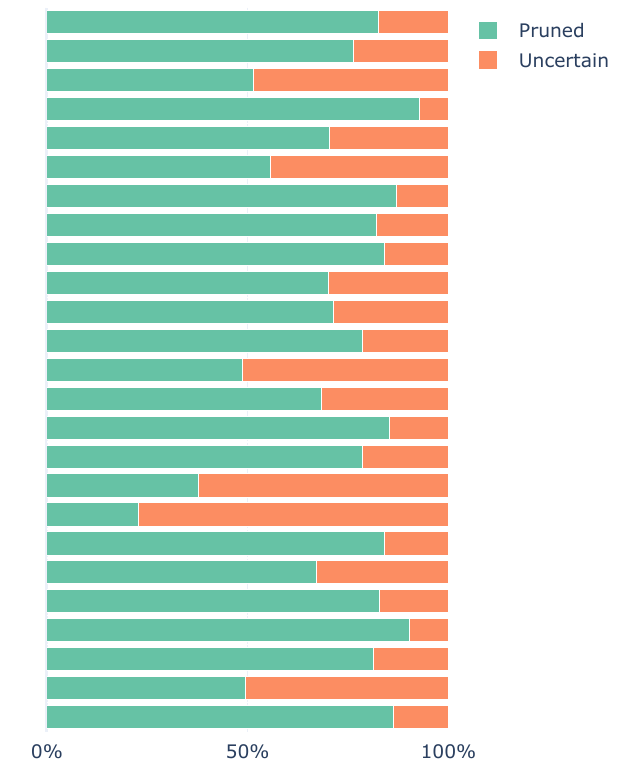}
  \end{center}
  \vspace{0.5em}
  \caption{Label distributions comparison between pruned and uncertain data partitions as identified by Cleanlab. Left: Main and subregister distributions, with the \textit{Other} class denoting labels without a specified subregister. Right: Percentages of pruned vs. uncertain labels.} 
  \label{fig:cleaned_stats}
\end{figure}

\begin{figure}
    \begin{center}
    \includegraphics[width=0.7\linewidth]{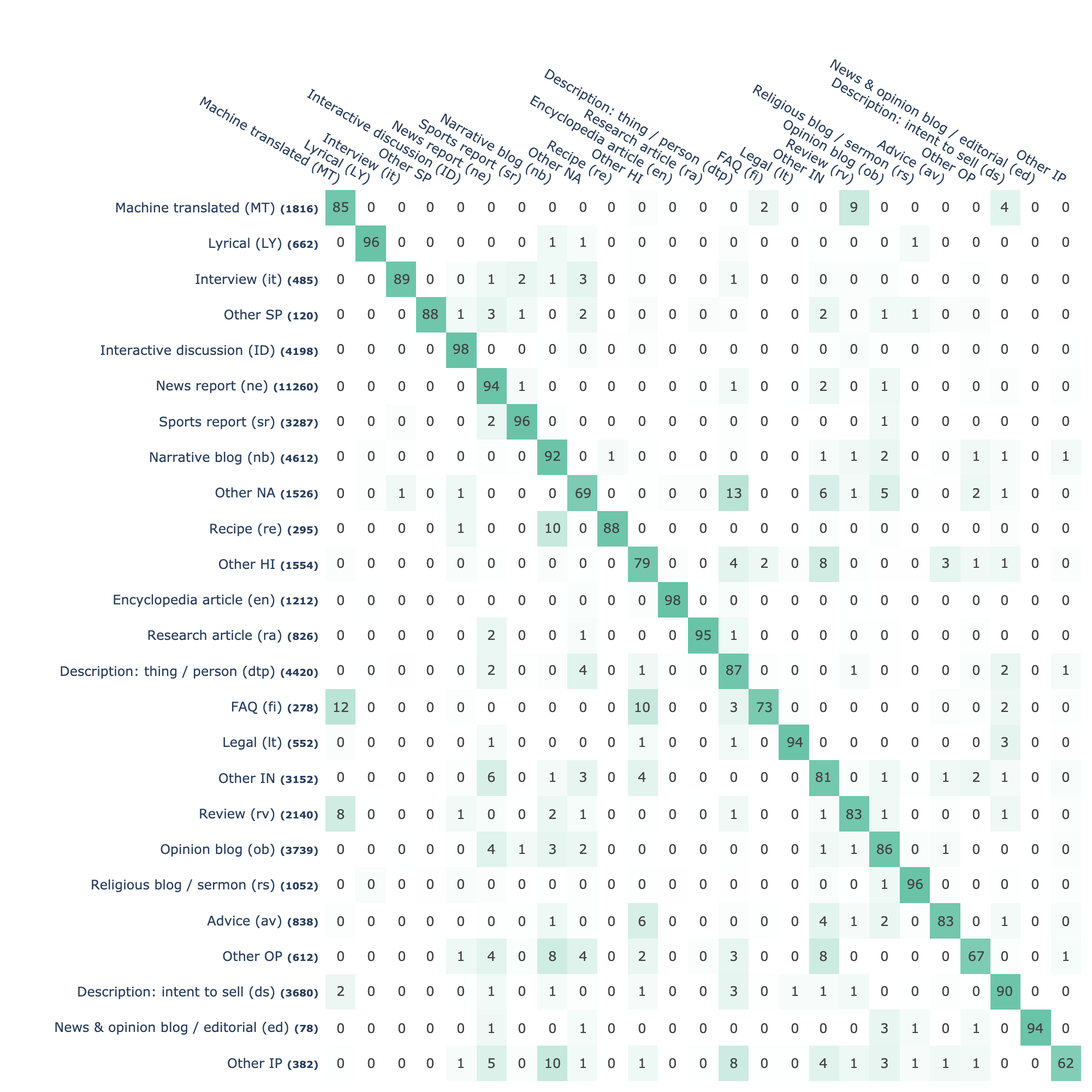}
    \includegraphics[width=0.7\linewidth]{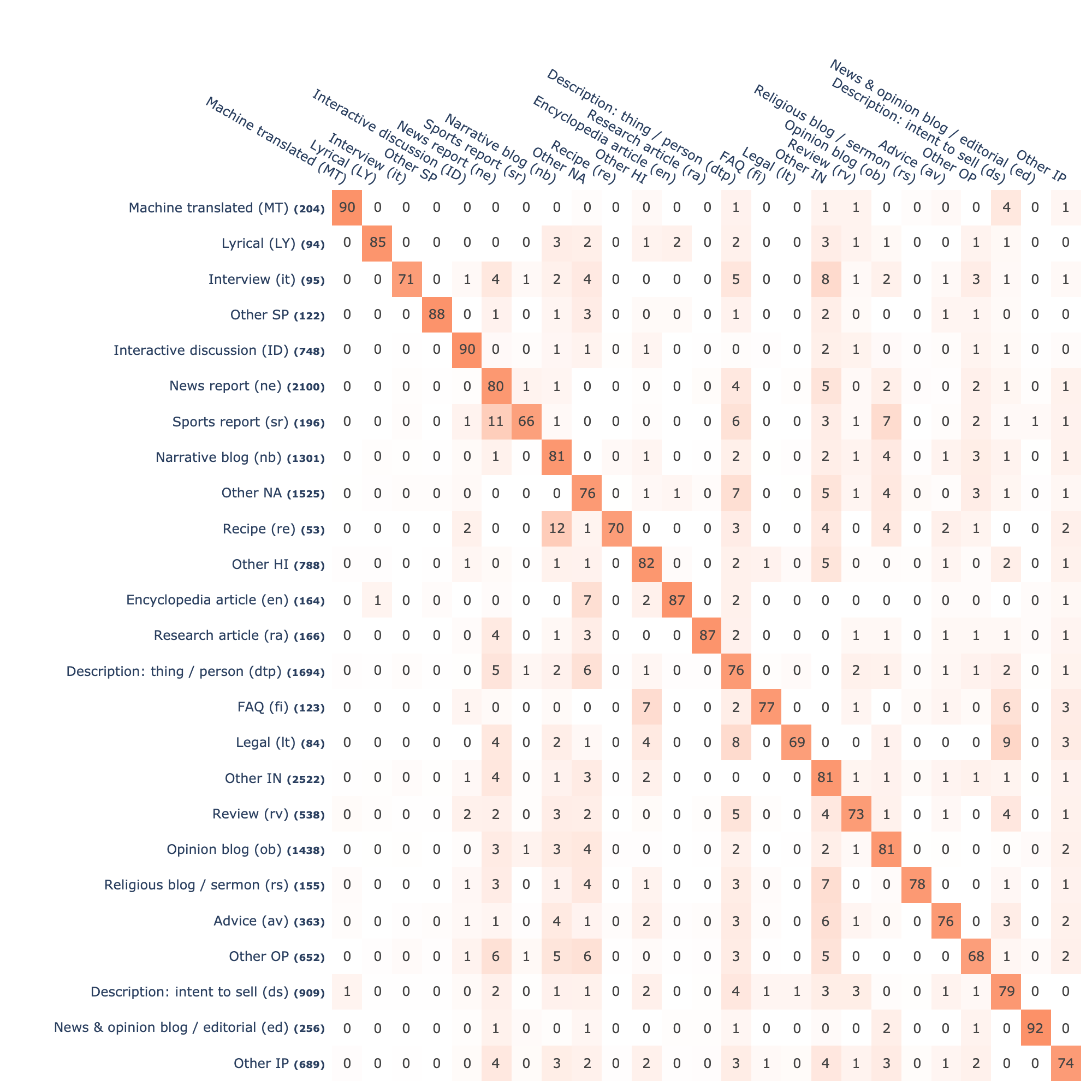}
  \end{center}
  \vspace{0.5em}
  \caption{Heatmaps showing class co-occurrences in pruned (left) and uncertain (right) datasets; diagonals represent single-label percentages, off-diagonals show co-occurrences.} 
  \label{fig:cleaned_heatmaps}
\end{figure}

Figure \ref{fig:cleaned_stats} compares class distributions between the pruned and uncertain partitions of the datasets. Documents without subregisters (the \textit{Other} classes in the figure) are most often marked as uncertain. The exception is \textit{News \& Opinion Blog or editorial}, where most documents were dropped, likely due to the small size of this class. This pattern matches expectations, since more specific categories typically imply more certainty. It also reflects a known challenge in web register identification: not all documents show clear register characteristics. Example \ref{example2} illustrates this, where the text was classified as \textit{Other IP}.

\begin{samepage}
\ex.
    \textit{Get Involved \\
    It' s easy, just devote at least one day a week to eating meat-free. The benefits will be greatest if you also avoid, or keep to a minimum, your use of dairy as well. We have materials you can use to promote MFM and run your own meat-free Mondays, along with ideas and recipes for you to enjoy. If you're vegetarian or vegan already, please use the campaign as an opportunity to get others involved by sharing your delicious dishes with them. And you can tell them how much cheaper it is to be vegetarian! Sign up Send me regular email updates How to... DIY - Just remember not to eat meat on a Monday and for inspiration and ideas see our Meat Free Monday recipes SHARE - The more people that take part the greater the impact, so get your family, friends, workplace, school, restaurant, shop involved.} \label{example2}

\end{samepage}

The text clearly aims to persuade but lacks specific features of any subregister within \textit{Informational persuasion}: it is neither a \textit{News \& opinion blog or editorial}, nor does it show features of \textit{Description with intent to sell}, as it is not selling anything. It simply aims to persuade readers to eat less meat.

For hybrid documents specifically, the label co-occurrence heatmaps in Figure \ref{fig:cleaned_heatmaps} (bottom panels) show that pruning does not significantly change the distribution of register combinations. This suggests that register ambiguity appears throughout the dataset rather than being concentrated in specific hybrids.

Table \ref{tbl:cleaned_results} presents classification results on pruned data, comparing models trained on pruned data (left) with those trained on the full dataset (right). Both models perform worst when tested on uncertain data (F1 scores of 30–46\%). This confirms that our data pruning method effectively identifies examples that are difficult to classify.

Interestingly, when tested on the full dataset, models trained on the full data outperform those trained on pruned data. The full-trained model achieves notably higher micro-average (77\%) and macro-average (71\%) F1 scores, while the pruned model reaches only 59\% and 52\%, respectively. This suggests that noisy examples provide valuable information for the classifier, exposing it to hard-to-classify cases and edge cases common in real-world data. Manual inspection of a sample of removed examples (not detailed here) supports this finding.

Finally, both models perform exceptionally well on the pruned dataset, exceeding the performance ceiling by a large margin. The model trained on pruned data reaches 91\% F1 on the pruned test set, while the model trained on full data achieves 89\%. These results demonstrate how noisy web data affects register identification and help quantify the impact of uncertain labels on model evaluation. Removing ambiguous, outlier, or otherwise difficult cases from evaluation boosts performance by over 10 percentage points. This level of performance also helps explain why neither additional training data nor more sophisticated models improve classification performance (see Section \ref{sec:comparing_results}).

\begin{table}
  
  \caption{Micro ($\mu$) and macro ($M$) F1 scores for experiments using different combinations of the pruned and full datasets, using the XLM-R model and the hierarchical CORE scheme. The pruned dataset consists of 76\% of the original full dataset, with the remaining 24\% of data identified as having uncertain labels.}\label{tbl:cleaned_results}
  {\tablefont
\setlength{\tabcolsep}{3.5pt}
\renewcommand{\arraystretch}{1.3}
    \footnotesize
  \begin{tabular}{@{\extracolsep{\fill}}lllllllp{15pt}llllll}
  \hline
& \multicolumn{6}{l}{Train: Pruned data} & & \multicolumn{5}{l}{Train: Full data}\\
  \hline
& En & Fi & Fr & Sv & Tr & Avg. & & En & Fi & Fr & Sv & Tr & Avg. \\
  \hline
& \multicolumn{6}{l}{Evaluate: Uncertain data} & & \multicolumn{5}{l}{Evaluate: Uncertain data} \\
  \hline
 $\mu$ &35 \tiny{(0.85)} & 36 \tiny{(0.18)} & 43 \tiny{(0.35)} & 46 \tiny{(0.12)} & 34 \tiny{(0.62)} & 39 \tiny{(0.42)} & & 39 \tiny{(0.37)} & 40 \tiny{(0.40)} & 45 \tiny{(0.57)} & 48 \tiny{(0.16)} & 38 \tiny{(0.86)} & 42 \tiny{(0.47)}  \\
  \hdashline
 $M$ & 24 \tiny{(0.25)} & 30 \tiny{(1.44)} & 37 \tiny{(0.26)} & 37 \tiny{(0.83)} & 30 \tiny{(0.52)} & 31 \tiny{(0.66)} & & 27 \tiny{(0.83)} & 32 \tiny{(1.10)} & 40 \tiny{(0.40)} & 39 \tiny{(0.65)} & 35 \tiny{(1.34)} & 35 \tiny{(0.86)} \\
  \hline
& \multicolumn{6}{l}{Evaluate: Full data} & & \multicolumn{5}{l}{Evaluate: Full data} \\
  \hline
$\mu$ &54 \tiny{(0.54)} & 61 \tiny{(0.07)} & 59 \tiny{(0.27)} & 64 \tiny{(0.06)} & 56 \tiny{(0.27)} & 59 \tiny{(0.24)} & & 72 \tiny{(0.14)} & 79 \tiny{(0.42)} & 75 \tiny{(0.16)} & 81 \tiny{(0.26)} & 78 \tiny{(0.63)} & 77 \tiny{(0.32)} \\
 \hdashline
 $M$ & 48 \tiny{(0.44)} & 55 \tiny{(1.16)} & 56 \tiny{(0.36)} & 58 \tiny{(0.55)} & 46 \tiny{(1.04)} & 52 \tiny{(0.71)} & & 67 \tiny{(0.85)} & 72 \tiny{(0.98)} & 73 \tiny{(0.31)} & 75 \tiny{(0.37)} & 66 \tiny{(1.52)} & 71 \tiny{(0.80)} \\
  \hline
& \multicolumn{6}{l}{Evaluate: Pruned data} & & \multicolumn{5}{l}{Evaluate: Pruned data} \\
  \hline
 $\mu$ &85 \tiny{(0.36)} & 92 \tiny{(0.27)} & 91 \tiny{(0.08)} & 93 \tiny{(0.03)} & 92 \tiny{(0.73)} & 91 \tiny{(0.29)} & &84 \tiny{(0.34)} & 91 \tiny{(0.47)} & 89 \tiny{(0.62)} & 91 \tiny{(0.15)} & 88 \tiny{(0.94)} & 89 \tiny{(0.50)} \\
  \hdashline
 $M$ & 81 \tiny{(0.21)} & 89 \tiny{(0.89)} & 90 \tiny{(0.92)} & 87 \tiny{(0.42)} & 85 \tiny{(2.19)} & 86 \tiny{(0.93)} & & 79 \tiny{(0.68)} & 85 \tiny{(0.17)} & 83 \tiny{(0.04)} & 88 \tiny{(0.60)} & 80 \tiny{(0.88)} & 83 \tiny{(0.48)} \\
\hline
\end{tabular}}{}
\end{table}

\subsubsection{Hybrid documents} \label{subsubsec:hybrids}

Hybrids represent another challenging aspect of web register identification, as shown for English by \citet{laippala_ronnqvist2023} and for Slovenian by \citet{kuzman-etal-2022-ginco}. We extend this analysis to our multilingual setting by comparing classifier performance across three dataset segments: (a) hybrid documents only, (b) non-hybrid documents only, and (c) the complete dataset. Table \ref{tab:hybrids} shows results from training and evaluating models on all combinations of these sets.

\begin{table}
  \caption{Micro ($\mu$) and macro ($M$) F1 scores across combinations of full, hybrid and non-hybrid labels in dataset and predictions. Percentages in parentheses show the proportion of included labels from the total dataset. All experiments use the XLM-R model and the full 25-class CORE taxonomy.}\label{tab:hybrids} 
  {\tablefont
\setlength{\tabcolsep}{3.5pt}
\renewcommand{\arraystretch}{1.3}
\footnotesize
\begin{tabular}{lllllllp{15pt}llllll}
\hline
& En & Fi & Fr & Sv & Tr & Avg. & & En & Fi & Fr & Sv & Tr & Avg. \\
\hline
& \multicolumn{6}{l}{Train: Full data, Evaluate: Non-hybrids (72\%)} & & \multicolumn{6}{l}{Train: Non-Hybrids, Evaluate: Hybrids (17\%)}  \\
\hline
 $\mu$ & 77 \tiny{(1.01)} & 81 \tiny{(0.25)} & 80 \tiny{(1.30)} & 83 \tiny{(0.36)} & 80 \tiny{(0.52)} & 80 \tiny{(0.69)} & & 55 \tiny{(0.70)} & 51 \tiny{(4.71)} & 54 \tiny{(0.26)} & 56 \tiny{(1.14)} & 56 \tiny{(2.36)} & 55 \tiny{(1.83)} \\
 \hdashline
 $M$ & 70 \tiny{(1.27)} & 73 \tiny{(0.51)} & 75 \tiny{(1.16)} & 74 \tiny{(1.12)} & 69 \tiny{(0.53)} & 72 \tiny{(0.92)} & & 49 \tiny{(1.78)} & 53 \tiny{(4.50)} & 58 \tiny{(2.81)} & 62 \tiny{(2.16)} & 58 \tiny{(2.83)} & 56 \tiny{(2.82)} \\
\hline
& \multicolumn{6}{l}{Train: Full data, Evaluate: Hybrids (28\%)} & & \multicolumn{6}{l}{Train: Hybrids, Evaluate: Full data (22\%)} \\
\hline
 $\mu$ & 60 \tiny{(0.44)} & 69 \tiny{(1.06)} & 65 \tiny{(1.72)} & 75 \tiny{(1.14)} & 62 \tiny{(0.99)} & 66 \tiny{(1.07)} & & 66 \tiny{(1.65)} & 78 \tiny{(0.87)} & 68 \tiny{(0.51)} & 72 \tiny{(0.16)} & 68 \tiny{(1.28)} & 71 \tiny{(0.90)}  \\
 \hdashline
 $M$ & 52 \tiny{(1.02)} & 62 \tiny{(0.25)} & 66 \tiny{(2.44)} & 69 \tiny{(2.05)} & 62 \tiny{(1.45)} & 62 \tiny{(1.44)} & & 55 \tiny{(1.55)} & 76 \tiny{(1.43)} & 67 \tiny{(0.95)} & 66 \tiny{(0.56)} & 70 \tiny{(0.91)} & 67 \tiny{(1.08)} \\
\hline
& \multicolumn{6}{l}{Train: Non-hybrids, Evaluate: Full data (78\%)} & & \multicolumn{6}{l}{Train: Hybrids, Evaluate: Non-hybrids (12\%)}\\
\hline
 $\mu$ & 73 \tiny{(0.54)} & 79 \tiny{(0.43)} & 78 \tiny{(0.35)} & 84 \tiny{(0.40)} & 79 \tiny{(0.75)} & 79 \tiny{(0.49)} & & 60 \tiny{(1.09)} & 57 \tiny{(0.92)} & 59 \tiny{(1.90)} & 59 \tiny{(0.46)} & 62 \tiny{(1.13)} & 59 \tiny{(1.10)} \\
 \hdashline
 $M$ & 68 \tiny{(0.72)} & 72 \tiny{(1.04)} & 76 \tiny{(0.47)} & 79 \tiny{(0.74)} & 69 \tiny{(1.44)} & 73 \tiny{(0.88)} & & 54 \tiny{(1.65)} & 64 \tiny{(0.20)} & 60 \tiny{(2.07)} & 54 \tiny{(1.28)} & 64 \tiny{(1.12)} & 59 \tiny{(1.26)} \\
\hline
& \multicolumn{6}{l}{Train: Non-hybrids, Evaluate: Non-hybrids (60\%)} & & \multicolumn{6}{l}{Train: Hybrids, Evaluate: Hybrids (10\%)}\\
\hline
 $\mu$ & 80 \tiny{(0.92)} & 82 \tiny{(0.14)} & 84 \tiny{(1.13)} & 89 \tiny{(0.41)} & 82 \tiny{(0.59)} & 83 \tiny{(0.64)} & & 75 \tiny{(0.17)} & 93 \tiny{(2.03)} & 77 \tiny{(2.71)} & 84 \tiny{(0.67)} & 80 \tiny{(2.35)} & 82 \tiny{(1.58)}  \\
 \hdashline
 $M$ & 72 \tiny{(1.14)} & 75 \tiny{(0.63)} & 83 \tiny{(0.27)} & 83 \tiny{(2.17)} & 72 \tiny{(0.76)} & 77 \tiny{(0.99)} & & 63 \tiny{(2.46)} & 84 \tiny{(3.22)} & 76 \tiny{(2.27)} & 79 \tiny{(1.61)} & 86 \tiny{(2.09)} & 78 \tiny{(2.33)} \\
\hline
\end{tabular}}{}
\end{table}

Model performance varies considerably across these combinations. As expected, training and evaluating on non-hybrids gives the highest scores (83\%), and models trained on full data and tested on non-hybrids also perform well (80\%).

Surprisingly, models trained and tested only on hybrids perform nearly as well (82\%), contradicting previous studies that describe hybrids as particularly challenging to classify. Our results suggest that the main challenge is not classifying hybrids themselves, but distinguishing hybrids from non-hybrids. This is evident in Table \ref{tab:hybrids}: models trained on non-hybrids perform poorly on hybrids (55\%), and models trained on hybrids perform poorly on non-hybrids (59\%). The finding also aligns with our data pruning results (see Section \ref{subsubsec:pruning-experiments}), where algorithmic cleaning did not disproportionately remove hybrids or systematically alter their distribution—if hybrids were inherently problematic, we would expect them to be flagged more often.

Finally, Table \ref{tab:hybrids} reveals an unexpected finding: training on non-hybrids and testing on the full dataset yields our highest overall performance, 79\% micro and 73\% macro F1 averaged across languages. Removing multi-label documents from training, despite reducing the dataset size, actually improves model performance. This suggests that hybrids may introduce conflicting signals during training, making it harder for models to learn clear class boundaries. However, we did not analyze whether this improvement occurs mainly in non-hybrid predictions while harming hybrid predictions, or whether performance improves for both document types. Future work should investigate the source of these performance gains.

\subsection{Inference time comparison} \label{subsec:inference-times}

\begin{figure}
  \centering
  \begin{minipage}{0.75\textwidth}
  \includegraphics[width=\textwidth]{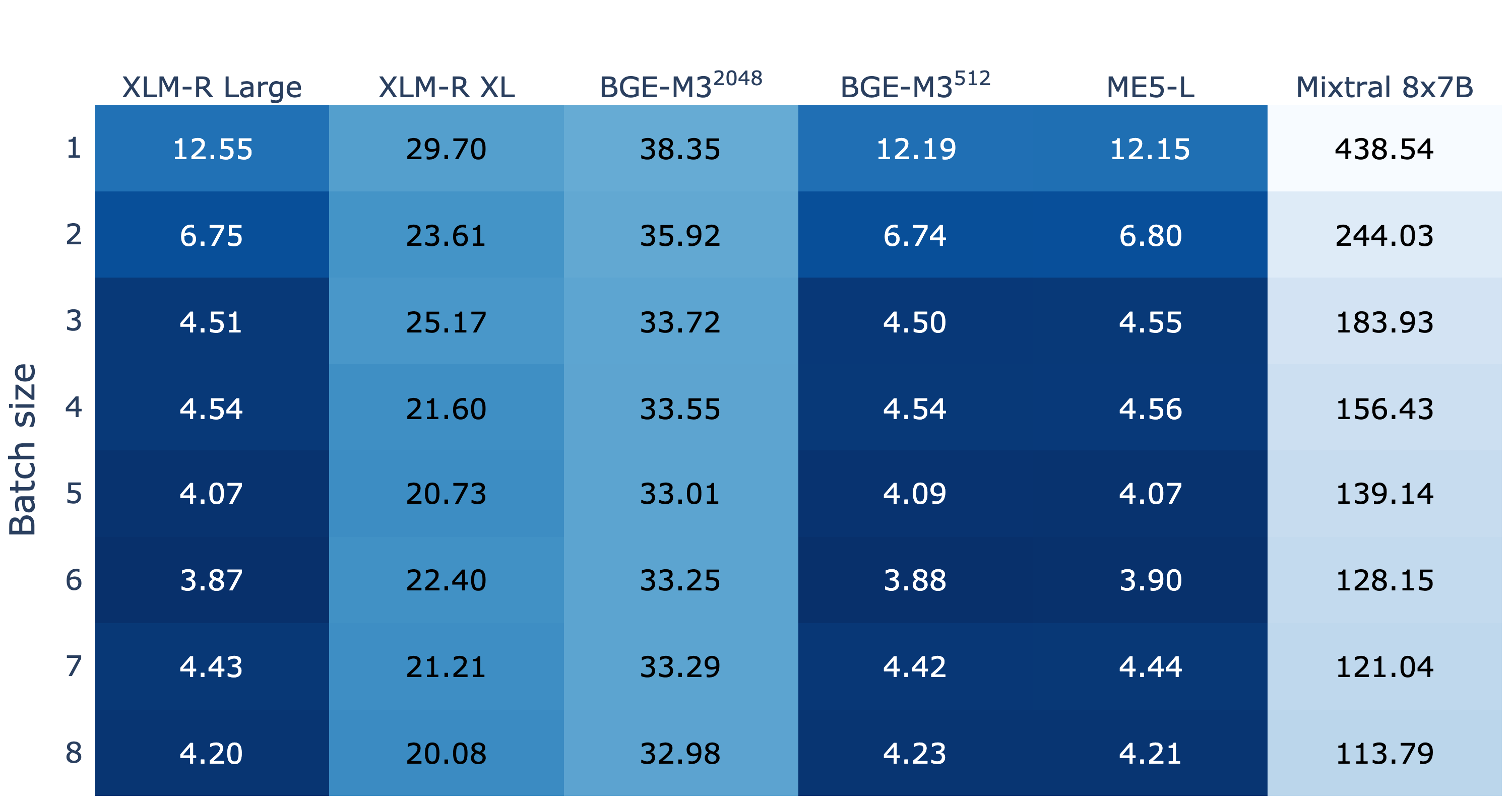}
  \vspace{0.5em}
  \caption{Inference times (latencies) for different models across different batch sizes.}
    \label{fig:inference_times}
    \end{minipage}
\end{figure}

We measure inference times for our six fine-tuned models: XLM-R Large, XLM-R XL, ME5, BGE-M3\textsuperscript{512}, BGE-M3\textsuperscript{2048}, and Mixtral 8x7B. To ensure accurate timing, we use CUDA events \citep{Atkins.MacLeod2023} and warm up the GPU to reduce initial overhead before measuring. All experiments run in the same software environment on a single NVIDIA A100 GPU, testing across all test sets with ten iterations per model and batch sizes from 1 to 8.

The results, shown in Figure \ref{fig:inference_times}, report latencies in milliseconds using the main label taxonomy. XLM-R Large, BGE-M3\textsuperscript{512}, and ME5 consistently process data fastest, making them the best choices for large-scale web register labeling. As expected, the larger models run more slowly, with Mixtral being the slowest by far. BGE-M3\textsuperscript{2048} is notably slower than BGE-M3\textsuperscript{512} due to its longer token limit.

\section{Discussion and conclusions}

\subsection{Model performance and multilinguality}

In this article, we have created and released the Multilingual CORE corpora, covering 16 languages with 72,504 documents annotated using a detailed 25-class hierarchical taxonomy. We trained and compared multiple deep learning architectures, finding a consistent performance ceiling around 77--80\% F1, and demonstrated successful zero-shot transfer to 11 additional languages. We also compared our CORE taxonomy with the simpler X-GENRE classification system, showed through SACX keyword analysis that model decisions are linguistically motivated, and conducted systematic experiments with data partitions to examine the effects of challenging cases such as documents with uncertain labels and hybrids.

Our best multilingual models achieve a 79\% micro F1 score when trained on non-hybrid documents and tested on the full dataset across the five training languages. In the standard multilingual setting (training and testing on all document types), we reach 77\% micro F1. These results represent substantial improvements over previous work: for English CORE, we improve from the previous best of 68\% F1 \citep{laippala-etal-2022-towards} to 72\% in both our monolingual and multilingual settings. Our multilingual approach also compares favorably with the 80\% F1 that \citet{Kuzman.Ljubesic2023} achieved on X-GENRE, which uses only nine classes and excludes hybrid documents entirely. We achieve competitive performance while using a more complex 25-class taxonomy that includes hybrids. Our macro F1 of 73\% demonstrates that performance remains strong even with fine-grained classification, which is important for applications requiring detailed metadata \citep[e.g.,][]{eskelinen-etal-2024-building}.

Multilingual training provides substantial benefits for smaller datasets and rare register classes, but these benefits depend heavily on training data size. Languages with limited data (Turkish, French, Swedish) improve by 1–3 percentage points with multilingual training, while Finnish shows no improvement because its 10,751 documents already provide sufficient training examples. Multilingual training particularly helps rare registers like \textit{Lyrical} and \textit{Spoken}, where individual languages have too few examples for effective monolingual training. However, cross-lingual transfer has clear limits: zero-shot performance on our five main languages drops by 7\% on average, with variation across languages (4--8\% decreases). For the 11 evaluation-only languages, zero-shot performance ranges from 50\% F1 (Japanese) to 82\% F1 (Farsi). These results suggest that while registers share cross-lingual features, they also maintain language-specific characteristics that require in-language training data.

\subsection{Register taxonomy and web data}

Our results support using the detailed, hierarchical CORE taxonomy for web register modeling. Direct comparison with the simpler X-GENRE scheme shows that CORE achieves equivalent performance (77\% micro F1) while providing much finer-grained classification: 25 classes compared to X-GENRE's 9. The additional granularity does not come at the cost of classification quality.

This detailed classification offers practical advantages. Specific register information enables applications such as those demonstrated by \cite{eskelinen-etal-2024-building}, who leveraged specific subregisters to extract question-answer pairs from web data. Our multi-label treatment of hybrid documents also preserves all register information rather than collapsing diverse combinations into a single ``Other'' category. This is particularly important because hybrids constitute up to 25\% of web documents \citep{biber2023register-culture}, and collapsing them into a single category discards valuable information about their register composition.

\subsection{Understanding web register complexity}

Our experiments with different data partitions reveal the extent to which web data noise affects register identification. When evaluating on pruned data (excluding documents with uncertain labels), performance improves to over 90\%. This demonstrates how much outliers and ambiguous cases affect model performance and suggests that many web documents are inherently difficult to assign to specific registers.

Our analysis of hybrid documents shows an interesting pattern: models trained and tested exclusively on hybrids achieve strong performance (82\% micro F1), but models trained on non-hybrids perform poorly when predicting hybrids (55\%), and vice versa (59\%). This suggests that the main challenge is distinguishing between hybrid and non-hybrid documents rather than classifying hybrids themselves. Surprisingly, training without hybrids and evaluating on the full test set yields our best overall results (79\% micro F1, compared to 77\% when training with hybrids).

These findings connect to recent research on the nature of web registers. Studies by Doug Biber and colleagues have shown that English web registers form a continuum rather than discrete categories \citep{BiberEgbertKeller+2020+581+616,biber2023register-culture,egbert2024_dual}, an observation that extends to other languages as well, according to a recent study \citep{henriksson-etal-2024-discrete}. This view is further supported by research on document-internal structure, where individual web documents often contain passages from different registers \citep{henriksson2025analyzing}, complicating classification at the document level.

\section{Limitations and further work}

Several aspects of this study could be extended in future work. Our SACX analysis demonstrated that model decisions are linguistically motivated, but was limited to two registers across three languages. Future work could expand this interpretability analysis to cover all 25 register classes and investigate whether keyword patterns differ systematically across languages. Examining misclassified documents could also reveal why certain register boundaries are difficult to detect and help explain the performance ceiling we observed \citep{Santini2005,Gries.etal2011}. Additionally, analyzing the semantic embeddings of web documents through clustering techniques could reveal register relationships not captured in the CORE taxonomy.

Our methodology has limitations that could be addressed in future work. The different sampling methods used for the language-specific collections may affect cross-lingual modeling results; creating a new English register dataset from Common Crawl (the same source used for all other languages) or from a large-scale cleaned and deduplicated web corpus such as HPLT \citep{burchell2025HPLT2} would help mitigate this issue. We also use only document beginnings (512 or 2,048 tokens) due to model token limits. Future research should examine document endings, beginning-end combinations, and various windowing methods \citep{sun2020finetune}. This is especially important for hybrid documents, where such analysis could reveal whether different segments contain distinct registers or have hybrid-specific characteristics \citep{henriksson2025analyzing}.

We see potential in several other research directions. We did not explore generative LLMs for web register identification, though recent research suggests that LLMs may outperform humans in many annotation tasks \citep[e.g.,][]{bermejo2024llms, Gilardi_2023, tornberg_2024}, and some studies have demonstrated promising results for zero-shot register classification using LLMs \citep{kuzman2023chatgpt,törnberg2023chatgpt4outperformsexpertscrowd,henriksson-etal-2024-discrete}. Current LLMs are computationally expensive for processing large-scale web data, but one direction would be using them to generate training annotations for fine-tuning smaller, faster encoder models. Finally, large-scale web register annotation using our classifier opens new possibilities for corpus-linguistic research \citep{2025arXiv250310267B}, and exploring how register composition in training data affects LLM quality \citep{amanda2025register} represents another promising direction.

\section*{Declaration of AI use}

We used ChatGPT \citep{openai2024gpt4o} and Claude 
\citep{anthropic2024claude3} for language checking, formatting, 
and generating visualizations. These tools were not used for data 
collection, analysis, or interpretation of results. The authors 
take full responsibility for all content.

\bibliography{references}

@article{bge-m3,
  title = {{{BGE}} M3-Embedding: {{Multi-lingual}}, Multi-Functionality, Multi-Granularity Text Embeddings through Self-Knowledge Distillation},
  author = {Chen, Jianlv and Xiao, Shitao and Zhang, Peitian and Luo, Kun and Lian, Defu and Liu, Zheng},
  year = {2024},
  eprint = {2402.03216},
  primaryclass = {cs.CL},
  archiveprefix = {arxiv},
journal={arXiv preprint arXiv:2402.03216}
}

@inproceedings{eskelinen-etal-2024-building,
    title = "Building Question-Answer Data Using Web Register Identification",
    author = "Eskelinen, Anni  and
      Myntti, Amanda  and
      Henriksson, Erik  and
      Pyysalo, Sampo  and
      Laippala, Veronika",
    editor = "Calzolari, Nicoletta  and
      Kan, Min-Yen  and
      Hoste, Veronique  and
      Lenci, Alessandro  and
      Sakti, Sakriani  and
      Xue, Nianwen",
    booktitle = "Proceedings of the 2024 Joint International Conference on Computational Linguistics, Language Resources and Evaluation (LREC-COLING 2024)",
    month = may,
    year = "2024",
    address = "Torino, Italia",
    publisher = "ELRA and ICCL",
    url = "https://aclanthology.org/2024.lrec-main.234",
    pages = "2595--2611",
}

@inproceedings{conneau-etal-2020-unsupervised,
    title = "Unsupervised Cross-lingual Representation Learning at Scale",
    author = "Conneau, Alexis  and
      Khandelwal, Kartikay  and
      Goyal, Naman  and
      Chaudhary, Vishrav  and
      Wenzek, Guillaume  and
      Guzm{\'a}n, Francisco  and
      Grave, Edouard  and
      Ott, Myle  and
      Zettlemoyer, Luke  and
      Stoyanov, Veselin",
    booktitle = "Proceedings of the 58th Annual Meeting of the Association for Computational Linguistics",
    month = jul,
    year = "2020",
    address = "Online",
    publisher = "Association for Computational Linguistics",
   url = "https://aclanthology.org/2020.acl-main.747",
    doi = "10.18653/v1/2020.acl-main.747",
    pages = "8440--8451",}

@article{Gao.etal2020,
  title={The pile: An 800gb dataset of diverse text for language modeling},
  author={Gao, Leo and Biderman, Stella and Black, Sid and Golding, Laurence and Hoppe, Travis and Foster, Charles and Phang, Jason and He, Horace and Thite, Anish and Nabeshima, Noa and others},
  journal={arXiv:2101.00027},
  year={2020}
}

@article{biber-egbert2016-using-grammatical-feats,
 author = {Biber, Douglas and Egbert, Jesse},
 doi = {10.1558/jrds.v2i1.27637},
 journal = {Journal of Research Design and Statistics in Linguistics and Communication Science},
 number = {1},
 pages = {3-36},
 title = {Using Grammatical Features for Automatic Register Identification in an Unrestricted Corpus of Documents from the Open Web},
 url = {https://app.dimensions.ai/details/publication/pub.1067980703},
 volume = {2},
 year = {2016},
}

@book{swales1990genre,
  title={Genre Analysis: English in Academic and Research Settings},
  author={Swales, John},
  year={1990},
  publisher={Cambridge University Press}
}

@article{BiberEgbertKeller+2020+581+616,
url = {https://doi.org/10.1515/cllt-2018-0086},
title = {Reconceptualizing register in a continuous situational space},
author = {Douglas Biber and Jesse Egbert and Daniel Keller},
pages = {581--616},
volume = {16},
number = {3},
journal = {Corpus Linguistics and Linguistic Theory},
doi = {doi:10.1515/cllt-2018-0086},
year = {2020},
lastchecked = {2024-04-10}
}

@article{tornberg_2024,
author = {Petter Törnberg},
title ={Large Language Models Outperform Expert Coders and Supervised Classifiers at Annotating Political Social Media Messages},
journal = {Social Science Computer Review},
year = {2024},
doi = {10.1177/08944393241286471},
URL = {https://doi.org/10.1177/08944393241286471},
eprint = {https://doi.org/10.1177/08944393241286471},
}

@incollection{santini2010cross,
author="Santini, Marina",
editor="Mehler, Alexander
and Sharoff, Serge
and Santini, Marina",
title="Cross-Testing a Genre Classification Model for the Web",
bookTitle="Genres on the Web: Computational Models and Empirical Studies",
year="2011",
publisher="Springer Netherlands",
address="Dordrecht",
pages="87--128",
isbn="978-90-481-9178-9",
doi="10.1007/978-90-481-9178-9_5",
url="https://doi.org/10.1007/978-90-481-9178-9_5"
}

@article{egbert2015developing,
  title={Developing a bottom-up, user-based method of web register classification},
  author={Egbert, Jesse and Biber, Douglas and Davies, Mark},
  journal={Journal of the Association for Information Science and Technology},
  volume={66},
  number={9},
  pages={1817--1831},
  year={2015}
}

@inproceedings{berninger2008building,
  title={Building a document genre corpus: a profile of the KRYS I corpus},
  author={Berninger, Virginia F. and Kim, Youngju and Ross, Sara},
  booktitle={BCS-IRSG Workshop on Corpus Profiling},
  pages={1--10},
  year={2008}
}

@article{biber2023register-culture,
  title={What is a register? Accounting for linguistic and situational variation within – and outside of – textual varieties},
  author={Biber, Douglas and Egbert, Jesse},
  journal={Register Studies},
  volume={5},
  number={1},
  pages={1--22},
  year={2023},
  month={June}
}

@inproceedings{stubbe2007towards,
  title={Towards a Reference Corpus of Web Genres},
  author={Stubbe, A. and Ringlstetter, C.},
  booktitle={Proceedings of Colloquium held in conjunction with Corpus Linguistics 2007},
  address={Birmingham, UK},
  month={July},
  year={2007}
}

@phdthesis{santini2007automatic,
  title={Automatic identification of genre in web pages},
  author={Santini, Marina},
  year={2007},
  school={University of Brighton},
  note={Unpublished doctoral dissertation}
}

@inproceedings{lepekhin2022estimating,
    title = "Estimating Confidence of Predictions of Individual Classifiers and {T}heir{E}nsembles for the Genre Classification Task",
    author = "Lepekhin, Mikhail  and
      Sharoff, Serge",
    editor = "Calzolari, Nicoletta  and
      B{\'e}chet, Fr{\'e}d{\'e}ric  and
      Blache, Philippe  and
      Choukri, Khalid  and
      Cieri, Christopher  and
      Declerck, Thierry  and
      Goggi, Sara  and
      Isahara, Hitoshi  and
      Maegaard, Bente  and
      Mariani, Joseph  and
      Mazo, H{\'e}l{\`e}ne  and
      Odijk, Jan  and
      Piperidis, Stelios",
    booktitle = "Proceedings of the Thirteenth Language Resources and Evaluation Conference",
    month = jun,
    year = "2022",
    address = "Marseille, France",
    publisher = "European Language Resources Association",
    url = "https://aclanthology.org/2022.lrec-1.642",
    pages = "5974--5982",
}

@article{sharoff-2019-dimensions,
author = {Sharoff, Serge},
title = {Functional Text Dimensions for the annotation of web corpora},
journal = {Corpora},
volume = {13},
number = {1},
pages = {65-95},
year = {2018},
doi = {10.3366/cor.2018.0136},
URL = {         https://doi.org/10.3366/cor.2018.0136
},
eprint = {        https://doi.org/10.3366/cor.2018.0136
 },
}

@inproceedings{eissen_stein2004,
  author       = {Sven Meyer zu Eissen and
                  Benno Stein},
  editor       = {Susanne Biundo and
                  Thom W. Fr{\"{u}}hwirth and
                  G{\"{u}}nther Palm},
  title        = {Genre Classification of Web Pages},
  booktitle    = {{KI} 2004: Advances in Artificial Intelligence, 27th Annual German
                  Conference on AI, {KI} 2004, Ulm, Germany, September 20-24, 2004,
                  Proceedings},
  series       = {Lecture Notes in Computer Science},
  volume       = {3238},
  pages        = {256--269},
  publisher    = {Springer},
  year         = {2004},
  url          = {https://doi.org/10.1007/978-3-540-30221-6\_20},
  doi          = {10.1007/978-3-540-30221-6\_20},
  bibsource    = {dblp computer science bibliography, https://dblp.org}
}

@article{Kuzman.Ljubesic2023,
  title = {Automatic Genre Identification: A Survey},
  shorttitle = {Automatic Genre Identification},
  author = {Kuzman, Taja and Ljube{\v s}i{\'c}, Nikola},
  year = {2023},
  month = nov,
  journal = {Language Resources and Evaluation},
  issn = {1574-0218},
  doi = {10.1007/s10579-023-09695-8},
  url = {https://doi.org/10.1007/s10579-023-09695-8},
  urldate = {2024-03-03},
  langid = {english}
}

@inproceedings{devlin-etal-2019-bert,
    title = "{BERT}: Pre-training of Deep Bidirectional Transformers for Language Understanding",
    author = "Devlin, Jacob  and
      Chang, Ming-Wei  and
      Lee, Kenton  and
      Toutanova, Kristina",
    editor = "Burstein, Jill  and
      Doran, Christy  and
      Solorio, Thamar",
    booktitle = "Proceedings of the 2019 Conference of the North {A}merican Chapter of the Association for Computational Linguistics: Human Language Technologies, Volume 1 (Long and Short Papers)",
    month = jun,
    year = "2019",
    address = "Minneapolis, Minnesota",
    publisher = "Association for Computational Linguistics",
    url = "https://aclanthology.org/N19-1423",
    doi = "10.18653/v1/N19-1423",
    pages = "4171--4186",
}

@inproceedings{kuzman-etal-2022-ginco,
    title = "The {GINCO} Training Dataset for Web Genre Identification of Documents Out in the Wild",
    author = "Kuzman, Taja  and
      Rupnik, Peter  and
      Ljube{\v{s}}i{\'c}, Nikola",
    booktitle = "Proceedings of the Thirteenth Language Resources and Evaluation Conference",
    month = jun,
    year = "2022",
    address = "Marseille, France",
    publisher = "European Language Resources Association",
    url = "https://aclanthology.org/2022.lrec-1.170",
    pages = "1584--1594",
}

@inproceedings{laippala-etal-2022-towards,
    title = "Towards better structured and less noisy Web data: Oscar with Register annotations",
    author = {Laippala, Veronika  and
      Salmela, Anna  and
      R{\"o}nnqvist, Samuel  and
      Aji, Alham Fikri  and
      Chang, Li-Hsin  and
      Dhifallah, Asma  and
      Goulart, Larissa  and
      Kortelainen, Henna  and
      P{\`a}mies, Marc  and
      Prina Dutra, Deise  and
      Skantsi, Valtteri  and
      Sutawika, Lintang  and
      Pyysalo, Sampo},
    booktitle = "Proceedings of the Eighth Workshop on Noisy User-generated Text (W-NUT 2022)",
    month = oct,
    year = "2022",
    address = "Gyeongju, Republic of Korea",
    publisher = "Association for Computational Linguistics",
    url = "https://aclanthology.org/2022.wnut-1.23",
    pages = "215--221",
}

@inproceedings{repo2021zeroshot,
  title={Beyond the English web: Zero-shot cross-lingual and lightweight monolingual classification of registers},
  author={Repo, Liina and Skantsi, Valtteri and R{\"o}nnqvist, Samuel and Hellstr{\"o}m, Saara and Oinonen, Miika and Salmela, Anna and Biber, Douglas and Egbert, Jesse and Pyysalo, Sampo and Laippala, Veronika},
  booktitle={Proceedings of the 16th Conference of the European Chapter of the Association for Computational Linguistics: Student Research Workshop},
  year={2021},
  url={https://aclanthology.org/2021.eacl-srw.24.pdf}
}

@inproceedings{ronnqvist-etal-2021-multilingual,
    title = "Multilingual and Zero-Shot is Closing in on Monolingual Web Register Classification",
    author = {R{\"o}nnqvist, Samuel  and
      Skantsi, Valtteri  and
      Oinonen, Miika  and
      Laippala, Veronika},
    booktitle = "Proceedings of the 23rd Nordic Conference on Computational Linguistics (NoDaLiDa)",
    year = "2021",
    address = "Reykjavik, Iceland (Online)",
    publisher = {Link{\"o}ping University Electronic Press, Sweden},
    url = "https://aclanthology.org/2021.nodalida-main.16",
    pages = "157--165",
}

@book{sogaard-explainable, 
title={Explainable Natural Language Processing}, author={Anders Søgaard},
year={2021},
publisher={Springer}
}

@article{lepekhin-sharoff-adversarial,
  author       = {Mikhail Lepekhin and
                  Serge Sharoff},
  title        = {Experiments with adversarial attacks on text genres},
  journal      = {CoRR},
  volume       = {abs/2107.02246},
  year         = {2021},
  url          = {https://arxiv.org/abs/2107.02246},
  eprinttype    = {arXiv},
  eprint       = {2107.02246},
  bibsource    = {dblp computer science bibliography, https://dblp.org}
}

@article{petrenz-webber,
author = {Petrenz, Philipp and Webber, Bonnie},
year = {2011},
month = {06},
pages = {385-393},
title = {Stable Classification of Text Genres},
volume = {37},
journal = {Computational Linguistics},
doi = {10.1162/COLI_a_00052}
}

@article{skantsi_laippala_2023, title={Analyzing the unrestricted web: The finnish corpus of online registers}, DOI={10.1017/S0332586523000021}, journal={Nordic Journal of Linguistics}, publisher={Cambridge University Press}, author={Skantsi, Valtteri and Laippala, Veronika}, year={2023}, pages={1–31}}

@article{egbert2024_dual,
author = {Egbert, Jesse and Biber, Douglas and Keller, Daniel and Gracheva, Marianna},
year = {2024},
month = {05},
pages = {},
title = {Register and the dual nature of functional correspondence: accounting for text-linguistic variation between registers, within registers, and without registers},
volume = {20},
journal = {Corpus Linguistics and Linguistic Theory},
doi = {10.1515/cllt-2024-0011}
}

@inproceedings{henriksson2025analyzing,
  title={Analyzing register variation in web texts through automatic segmentation},
  author={Henriksson, Erik and Hellstr{\"o}m, Saara and Laippala, Veronika},
  booktitle={Proceedings of the 5th International Conference on Natural Language Processing for Digital Humanities},
  pages={7--19},
  year={2025}
}

@article{laippala_ronnqvist2023,
title = "Register identification from the unrestricted open Web using the Corpus of Online Registers of English",
author = "Veronika Laippala and Samuel R{\"o}nnqvist and Miika Oinonen and Kyr{\"o}l{\"a}inen, {Aki Juhani} and Anna Salmela and Douglas Biber and Jesse Egbert and Sampo Pyysalo",
note = "Funding Information: Open Access funding provided by University of Turku (UTU) including Turku University Central Hospital. Funding was provided by Academy of Finland (Grant No. 331297), Emil Aaltosen s{\"a}{\"a}ti{\"o}, National Science Foundation (Grant No. 1147581). Publisher Copyright: {\textcopyright} 2022, The Author(s).",
year = "2023",
month = sep,
doi = "10.1007/s10579-022-09624-1",
language = "English (US)",
volume = "57",
pages = "1045--1079",
journal = "Language Resources and Evaluation",
issn = "1574-020X",
publisher = "Springer Netherlands",
number = "3",
}

@article{Kuzman-genre-identification-2023, 
    title={Automatic Genre Identification for Robust Enrichment of Massive Text Collections: Investigation of Classification Methods in the Era of Large Language Models}, 
    volume={5}, 
    ISSN={2504-4990}, 
    url={http://dx.doi.org/10.3390/make5030059}, 
    DOI={10.3390/make5030059}, 
    number={3}, 
    journal={Machine Learning and Knowledge Extraction}, 
    publisher={MDPI AG}, 
    author={Kuzman, Taja and Mozetič, Igor and Ljubešić, Nikola}, 
    year={2023}, 
    month={Sep}, 
    pages={1149–1175} 
    }

@inproceedings{Sundararajan2017,
  author    = {Mukund Sundararajan and Ankur Taly and Qiqi Yan},
  title     = {Axiomatic Attribution for Deep Networks},
  booktitle = {International Conference on Machine Learning},
  year      = {2017},
  pages     = {3319--3328},
  publisher = {PMLR},
}

@article{de-marneffe-etal-2021-universal,
    title = "{U}niversal {D}ependencies",
    author = "de Marneffe, Marie-Catherine  and
      Manning, Christopher D.  and
      Nivre, Joakim  and
      Zeman, Daniel",
    journal = "Computational Linguistics",
    volume = "47",
    number = "2",
    month = jun,
    year = "2021",
    address = "Cambridge, MA",
    publisher = "MIT Press",
    url = "https://aclanthology.org/2021.cl-2.11/",
    doi = "10.1162/coli_a_00402",
    pages = "255--308"
}

@inproceedings{
amanda2025register,
title={Register Always Matters: Analysis of {LLM} Pretraining Data Through the Lens of Language Variation},
author={Amanda Myntti and Erik Henriksson and Veronika Laippala and Sampo Pyysalo},
booktitle={Second Conference on Language Modeling},
year={2025}
}

@book{biber1988variation,
  title={Variation across Speech and Writing},
  author={Biber, Douglas},
  year={1988},
  publisher={Cambridge University Press},
  address={Cambridge}
}

@inproceedings{mallen-etal-2023-trust,
    title = "When Not to Trust Language Models: Investigating Effectiveness of Parametric and Non-Parametric Memories",
    author = "Mallen, Alex  and
      Asai, Akari  and
      Zhong, Victor  and
      Das, Rajarshi  and
      Khashabi, Daniel  and
      Hajishirzi, Hannaneh",
    editor = "Rogers, Anna  and
      Boyd-Graber, Jordan  and
      Okazaki, Naoaki",
    booktitle = "Proceedings of the 61st Annual Meeting of the Association for Computational Linguistics (Volume 1: Long Papers)",
    month = jul,
    year = "2023",
    address = "Toronto, Canada",
    publisher = "Association for Computational Linguistics",
    url = "https://aclanthology.org/2023.acl-long.546",
    doi = "10.18653/v1/2023.acl-long.546",
    pages = "9802--9822",
}

@inproceedings{feng-etal-2023-pretraining,
    title = "From Pretraining Data to Language Models to Downstream Tasks: Tracking the Trails of Political Biases Leading to Unfair {NLP} Models",
    author = "Feng, Shangbin  and
      Park, Chan Young  and
      Liu, Yuhan  and
      Tsvetkov, Yulia",
    editor = "Rogers, Anna  and
      Boyd-Graber, Jordan  and
      Okazaki, Naoaki",
    booktitle = "Proceedings of the 61st Annual Meeting of the Association for Computational Linguistics (Volume 1: Long Papers)",
    month = jul,
    year = "2023",
    address = "Toronto, Canada",
    publisher = "Association for Computational Linguistics",
    url = "https://aclanthology.org/2023.acl-long.656",
    doi = "10.18653/v1/2023.acl-long.656",
    pages = "11737--11762",
}

@inproceedings{asheghi-etal-2014-designing,
    title = "Designing and Evaluating a Reliable Corpus of Web Genres via Crowd-Sourcing",
    author = "Asheghi, Noushin Rezapour  and
      Sharoff, Serge  and
      Markert, Katja",
    editor = "Calzolari, Nicoletta  and
      Choukri, Khalid  and
      Declerck, Thierry  and
      Loftsson, Hrafn  and
      Maegaard, Bente  and
      Mariani, Joseph  and
      Moreno, Asuncion  and
      Odijk, Jan  and
      Piperidis, Stelios",
    booktitle = "Proceedings of the Ninth International Conference on Language Resources and Evaluation ({LREC}'14)",
    month = may,
    year = "2014",
    address = "Reykjavik, Iceland",
    publisher = "European Language Resources Association (ELRA)",
    url = "http://www.lrec-conf.org/proceedings/lrec2014/pdf/470_Paper.pdf",
    pages = "1339--1346",
}

@article{doi:10.1177/0075424216628955,
author = {Douglas Biber and Jesse Egbert},
title ={Register Variation on the Searchable Web: A Multi-Dimensional Analysis},

journal = {Journal of English Linguistics},
volume = {44},
number = {2},
pages = {95-137},
year = {2016},
doi = {10.1177/0075424216628955},

URL = { 
    
        https://doi.org/10.1177/0075424216628955
    
    

},
eprint = { 
    
        https://doi.org/10.1177/0075424216628955
    
    

}
}

@article{asheghi2016,
author = {Asheghi, Noushin Rezapour and Sharoff, Serge and Markert, Katja},
year = {2016},
month = {09},
pages = {1-39},
title = {Crowdsourcing for web genre annotation},
volume = {50},
journal = {Language Resources and Evaluation},
doi = {10.1007/s10579-015-9331-6}
}

@inproceedings{kumar-etal-2023-language,
    title = "Language Generation Models Can Cause Harm: So What Can We Do About It? An Actionable Survey",
    author = "Kumar, Sachin  and
      Balachandran, Vidhisha  and
      Njoo, Lucille  and
      Anastasopoulos, Antonios  and
      Tsvetkov, Yulia",
    editor = "Vlachos, Andreas  and
      Augenstein, Isabelle",
    booktitle = "Proceedings of the 17th Conference of the European Chapter of the Association for Computational Linguistics",
    month = may,
    year = "2023",
    address = "Dubrovnik, Croatia",
    publisher = "Association for Computational Linguistics",
    url = "https://aclanthology.org/2023.eacl-main.241",
    doi = "10.18653/v1/2023.eacl-main.241",
    pages = "3299--3321",
}

@article{gururangan2023scaling,
  title={Scaling Expert Language Models with Unsupervised Domain Discovery},
  author={Suchin Gururangan and Margaret Li and Mike Lewis and Weijia Shi and Tim Althoff and Noah A. Smith and Luke Zettlemoyer},
  journal={ArXiv:2303.14177}, 
  year={2023},
  volume={},
  url={https://api.semanticscholar.org/CorpusID:257756896}
}

@book{Biber_Egbert_2018, place={Cambridge}, title={Register Variation Online}, publisher={Cambridge University Press}, author={Biber, Douglas and Egbert, Jesse}, year={2018}}

@article{kilgarriff2003introduction,
  title={Introduction to the Special Issue on the Web as Corpus},
  author={Kilgarriff, Adam and Grefenstette, Gregory},
  journal={Computational Linguistics},
  volume={29},
  number={3},
  pages={333--348},
  year={2003},
  publisher={MIT Press},
}

@article{2020t5,
  author  = {Colin Raffel and Noam Shazeer and Adam Roberts and Katherine Lee and Sharan Narang and Michael Matena and Yanqi Zhou and Wei Li and Peter J. Liu},
  title   = {Exploring the Limits of Transfer Learning with a Unified Text-to-Text Transformer},
  journal = {Journal of Machine Learning Research},
  year    = {2020},
  volume  = {21},
  number  = {140},
  pages   = {1-67},
  url     = {http://jmlr.org/papers/v21/20-074.html}
}

@inproceedings{sharoff-etal-2010-web,
    title = "The Web Library of Babel: evaluating genre collections",
    author = "Sharoff, Serge  and
      Wu, Zhili  and
      Markert, Katja",
    editor = "Calzolari, Nicoletta  and
      Choukri, Khalid  and
      Maegaard, Bente  and
      Mariani, Joseph  and
      Odijk, Jan  and
      Piperidis, Stelios  and
      Rosner, Mike  and
      Tapias, Daniel",
    booktitle = "Proceedings of the Seventh International Conference on Language Resources and Evaluation ({LREC}'10)",
    month = may,
    year = "2010",
    address = "Valletta, Malta",
    publisher = "European Language Resources Association (ELRA)",
    url = "http://www.lrec-conf.org/proceedings/lrec2010/pdf/28_Paper.pdf",
}

@inproceedings{NIPS2017_3f5ee243,
  title = {Attention Is All You Need},
  booktitle = {Advances in Neural Information Processing Systems},
  author = {Vaswani, Ashish and Shazeer, Noam and Parmar, Niki and Uszkoreit, Jakob and Jones, Llion and Gomez, Aidan N and Kaiser, {\L}ukasz and Polosukhin, Illia},
  editor = {Guyon, I. and Luxburg, U. Von and Bengio, S. and Wallach, H. and Fergus, R. and Vishwanathan, S. and Garnett, R.},
  year = {2017},
  volume = {30},
  publisher = {Curran Associates, Inc.},
  url = {https://proceedings.neurips.cc/paper_files/paper/2017/file/3f5ee243547dee91fbd053c1c4a845aa-Paper.pdf}
}

@inproceedings{Wolf.etal2020,
  title = {Transformers: {{State-of-the-Art Natural Language Processing}}},
  shorttitle = {Transformers},
  booktitle = {Proceedings of the 2020 {{Conference}} on {{Empirical Methods}} in {{Natural Language Processing}}: {{System Demonstrations}}},
  author = {Wolf, Thomas and Debut, Lysandre and Sanh, Victor and Chaumond, Julien and Delangue, Clement and Moi, Anthony and Cistac, Pierric and Rault, Tim and Louf, Remi and Funtowicz, Morgan and Davison, Joe and Shleifer, Sam and {von Platen}, Patrick and Ma, Clara and Jernite, Yacine and Plu, Julien and Xu, Canwen and Le Scao, Teven and Gugger, Sylvain and Drame, Mariama and Lhoest, Quentin and Rush, Alexander},
  editor = {Liu, Qun and Schlangen, David},
  year = {2020},
  month = oct,
  pages = {38--45},
  publisher = {Association for Computational Linguistics},
  address = {Online},
  doi = {10.18653/v1/2020.emnlp-demos.6},
  url = {https://aclanthology.org/2020.emnlp-demos.6},
  urldate = {2024-03-15}
}

@inproceedings{xiao2022retromae,
    title = "{R}etro{MAE}: Pre-Training Retrieval-oriented Language Models Via Masked Auto-Encoder",
    author = "Xiao, Shitao  and
      Liu, Zheng  and
      Shao, Yingxia  and
      Cao, Zhao",
    editor = "Goldberg, Yoav  and
      Kozareva, Zornitsa  and
      Zhang, Yue",
    booktitle = "Proceedings of the 2022 Conference on Empirical Methods in Natural Language Processing",
    month = dec,
    year = "2022",
    address = "Abu Dhabi, United Arab Emirates",
    publisher = "Association for Computational Linguistics",
    url = "https://aclanthology.org/2022.emnlp-main.35",
    doi = "10.18653/v1/2022.emnlp-main.35",
    pages = "538--548",
}

@article{Yuan.etal2021,
  title = {{{WuDaoCorpora}}: {{A}} Super Large-Scale {{Chinese}} Corpora for Pre-Training Language Models},
  shorttitle = {{{WuDaoCorpora}}},
  author = {Yuan, Sha and Zhao, Hanyu and Du, Zhengxiao and Ding, Ming and Liu, Xiao and Cen, Yukuo and Zou, Xu and Yang, Zhilin and Tang, Jie},
  year = {2021},
  month = jan,
  journal = {AI Open},
  volume = {2},
  pages = {65--68},
  issn = {2666-6510},
  doi = {10.1016/j.aiopen.2021.06.001},
  url = {https://www.sciencedirect.com/science/article/pii/S2666651021000152},
  urldate = {2024-03-12}
}

@misc{tuningplaybookgithub,
  author = {Varun Godbole and George E. Dahl and Justin Gilmer and Christopher J. Shallue and Zachary Nado},
  title = {Deep Learning Tuning Playbook},
    howpublished={http://github.com/google-research/tuning\_playbook},
  url = {http://github.com/google-research/tuning_playbook},
  year = {2023},
  note = {Version 1.0}
}

@inproceedings{mccoy-etal-2020-berts,
  title = {{{BERTs}} of a Feather Do Not Generalize Together: {{Large}} Variability in Generalization across Models with Similar Test Set Performance},
  booktitle = {Proceedings of the Third {{BlackboxNLP}} Workshop on Analyzing and Interpreting Neural Networks for {{NLP}}},
  author = {McCoy, R. Thomas and Min, Junghyun and Linzen, Tal},
  editor = {Alishahi, Afra and Belinkov, Yonatan and Chrupa{\l}a, Grzegorz and Hupkes, Dieuwke and Pinter, Yuval and Sajjad, Hassan},
  year = {2020},
  month = nov,
  pages = {217--227},
  publisher = {Association for Computational Linguistics},
  address = {Online},
  doi = {10.18653/v1/2020.blackboxnlp-1.21},
  url = {https://aclanthology.org/2020.blackboxnlp-1.21}
}

@article{laippala2021exploring,
  author = {Laippala, Veronika and Egbert, Jesse and Biber, Douglas and Kyröläinen, Akseli J.},
  title = {Exploring the Role of Lexis and Grammar for the Stable Identification of Register in an Unrestricted Corpus of Web Documents},
  journal = {Language Resources and Evaluation},
  volume = {55},
  number = {3},
  pages = {757--788},
  year = {2021},
  doi = {10.1007/s10579-020-09519-z},
  url = {https://doi.org/10.1007/s10579-020-09519-z}
}

@inproceedings{luotolahti-etal-2015-towards,
    title = "Towards Universal Web Parsebanks",
    author = "Luotolahti, Juhani  and
      Kanerva, Jenna  and
      Laippala, Veronika  and
      Pyysalo, Sampo  and
      Ginter, Filip",
    editor = "Nivre, Joakim  and
      Haji{\v{c}}ov{\'a}, Eva",
    booktitle = "Proceedings of the Third International Conference on Dependency Linguistics (Depling 2015)",
    month = aug,
    year = "2015",
    address = "Uppsala, Sweden",
    publisher = "Uppsala University, Uppsala, Sweden",
    url = "https://aclanthology.org/W15-2124",
    pages = "211--220",
}

@article{liaw2018tune,
  title = {Tune: {{A}} Research Platform for Distributed Model Selection and Training},
  author = {Liaw, Richard and Liang, Eric and Nishihara, Robert and Moritz, Philipp and Gonzalez, Joseph E and Stoica, Ion},
  year = {2018},
  journal = {arXiv preprint arXiv:1807.05118},
  eprint = {1807.05118},
  archiveprefix = {arxiv}
}

@article{kuzman2023chatgpt,
      title={ChatGPT: Beginning of an End of Manual Linguistic Data Annotation? Use Case of Automatic Genre Identification}, 
      author={Taja Kuzman and Igor Mozetič and Nikola Ljubešić},
      year={2023},
journal={arXiv:2303.03953},
      eprint={2303.03953},
      archivePrefix={arXiv},
      primaryClass={cs.CL}
}

@inproceedings{pmlr-v28-bergstra13,
  title = {Making a Science of Model Search: {{Hyperparameter}} Optimization in Hundreds of Dimensions for Vision Architectures},
  booktitle = {Proceedings of the 30th International Conference on Machine Learning},
  author = {Bergstra, James and Yamins, Daniel and Cox, David},
  editor = {Dasgupta, Sanjoy and McAllester, David},
  year = {2013},
  month = jun,
  series = {Proceedings of Machine Learning Research},
  volume = {28},
  pages = {115--123},
  publisher = {PMLR},
  address = {Atlanta, Georgia, USA},
  url = {https://proceedings.mlr.press/v28/bergstra13.html}
}

@article{li2020massively,
  title = {A System for Massively Parallel Hyperparameter Tuning},
  author = {Li, Liam and Jamieson, Kevin and Rostamizadeh, Afshin and Gonina, Ekaterina and Hardt, Moritz and Recht, Benjamin and Talwalkar, Ameet},
  year = {2020},
journal={arXiv preprint:1810.05934},
  eprint = {1810.05934},
  primaryclass = {cs.LG},
  archiveprefix = {arxiv}
}

@book{Goodfellow-et-al-2016,
  title = {Deep Learning},
  author = {Goodfellow, Ian and Bengio, Yoshua and Courville, Aaron},
  year = {2016},
  publisher = {MIT Press}
}

@inproceedings{goyal2021largerscale,
    title = "Larger-Scale Transformers for Multilingual Masked Language Modeling",
    author = "Goyal, Naman  and
      Du, Jingfei  and
      Ott, Myle  and
      Anantharaman, Giri  and
      Conneau, Alexis",
    editor = "Rogers, Anna  and
      Calixto, Iacer  and
      Vuli{\'c}, Ivan  and
      Saphra, Naomi  and
      Kassner, Nora  and
      Camburu, Oana-Maria  and
      Bansal, Trapit  and
      Shwartz, Vered",
    booktitle = "Proceedings of the 6th Workshop on Representation Learning for NLP (RepL4NLP-2021)",
    month = aug,
    year = "2021",
    address = "Online",
    publisher = "Association for Computational Linguistics",
    url = "https://aclanthology.org/2021.repl4nlp-1.4",
    doi = "10.18653/v1/2021.repl4nlp-1.4",
    pages = "29--33",
}

@article{liu2019roberta,
      title={RoBERTa: A Robustly Optimized BERT Pretraining Approach}, 
      author={Yinhan Liu and Myle Ott and Naman Goyal and Jingfei Du and Mandar Joshi and Danqi Chen and Omer Levy and Mike Lewis and Luke Zettlemoyer and Veselin Stoyanov},
      year={2019},
        journal={arXiv preprint:1907.11692},
      eprint={1907.11692},
      archivePrefix={arXiv},
      primaryClass={cs.CL}
}

@inproceedings{laippala-etal-2019-toward,
    title = "Toward Multilingual Identification of Online Registers",
    author = {Laippala, Veronika  and
      Kyll{\"o}nen, Roosa  and
      Egbert, Jesse  and
      Biber, Douglas  and
      Pyysalo, Sampo},
    editor = "Hartmann, Mareike  and
      Plank, Barbara",
    booktitle = "Proceedings of the 22nd Nordic Conference on Computational Linguistics",
    month = sep # "{--}" # oct,
    year = "2019",
    address = "Turku, Finland",
    publisher = {Link{\"o}ping University Electronic Press},
    url = "https://aclanthology.org/W19-6130",
    pages = "292--297",
}

@Misc{peft,
  title =        {PEFT: State-of-the-art Parameter-Efficient Fine-Tuning methods},
  author =       {Sourab Mangrulkar and Sylvain Gugger and Lysandre Debut and Younes Belkada and Sayak Paul and Benjamin Bossan},
  howpublished = {https://github.com/huggingface/peft},
  year =         {2022}
}

@incollection{Kuzman.Pollak2022,
  title = {Assessing {{Comparability}} of {{Genre Datasets}} via {{Cross-Lingual}} and {{Cross-Dataset Experiments}}},
  booktitle = {Jezikovne Tehnologije in Digitalna Humanistika: {{Zbornik}} Konference},
  author = {Kuzman, Taja and Pollak, S},
  editor = {Fi{\v s}er, D and Erjavec, T},
  year = {2022},
  pages = {100--107},
  publisher = {Institute of Contemporary History.}
}

@inproceedings{ronnqvist-etal-2022-SACX,
    title = "Explaining Classes through Stable Word Attributions",
    author = {R{\"o}nnqvist, Samuel  and
      Kyr{\"o}l{\"a}inen, Aki-Juhani  and
      Myntti, Amanda  and
      Ginter, Filip  and
      Laippala, Veronika},
    editor = "Muresan, Smaranda  and
      Nakov, Preslav  and
      Villavicencio, Aline",
    booktitle = "Findings of the Association for Computational Linguistics: ACL 2022",
    month = may,
    year = "2022",
    address = "Dublin, Ireland",
    publisher = "Association for Computational Linguistics",
    url = "https://aclanthology.org/2022.findings-acl.85",
    doi = "10.18653/v1/2022.findings-acl.85",
    pages = "1063--1074",
}

@article{northcutt2021confidentlearning,
    title={Confident Learning: Estimating Uncertainty in Dataset Labels},
    author={Curtis G. Northcutt and Lu Jiang and Isaac L. Chuang},
    journal={Journal of Artificial Intelligence Research (JAIR)},
    volume={70},
    pages={1373--1411},
    year={2021}
}

@inproceedings{thyagarajan2023multilabel,
    title={Identifying Incorrect Annotations in Multi-Label Classification Data},
    author={Thyagarajan, Aditya and Snorrason, Elías and Northcutt, Curtis and Mueller, Jonas},
    booktitle={ICLR Workshop on Trustworthy ML},
    year={2023}
}

@article{chen2024automated,
      title={Automated Data Curation for Robust Language Model Fine-Tuning}, 
      author={Jiuhai Chen and Jonas Mueller},
      year={2024},
      eprint={2403.12776},
      archivePrefix={arXiv},
      primaryClass={cs.CL},
journal={arXiv preprint arXiv:2403.12776}
}

@article{goh2022crowdlab,
  title={CROWDLAB: Supervised learning to infer consensus labels and quality scores for data with multiple annotators},
  author={Goh, Hui Wen and Tkachenko, Ulyana and Mueller, Jonas},
  journal={arXiv:2210.06812},
  year={2022}
}

@inproceedings{vidulin2007using,
  title={Using genres to improve search engines},
  author={Vidulin, Vedrana and Lu{\v{s}}trek, Mitja and Gams, Matja{\v{z}}},
  booktitle={1st International Workshop: Towards Genre-Enabled Search Engines: The Impact of Natural Language Processing},
  pages={45--51},
  year={2007}
}

@article{Sharoff2021,
  title = {Genre Annotation for the {{Web}}: {{Text-external}} and Text-Internal Perspectives},
  shorttitle = {Genre Annotation for the {{Web}}},
  author = {Sharoff, Serge},
  year = {2021},
  month = jun,
  journal = {Register Studies},
  volume = {3},
  number = {1},
  pages = {1--32},
  publisher = {John Benjamins Publishing Company},
  issn = {2542-9477},
  doi = {10.1075/rs.19015.sha},
  url = {https://benjamins.com/catalog/rs.19015.sha},
  urldate = {2024-06-04},
  langid = {english}
}

@inproceedings{sun2020finetune,
author="Sun, Chi
and Qiu, Xipeng
and Xu, Yige
and Huang, Xuanjing",
editor="Sun, Maosong
and Huang, Xuanjing
and Ji, Heng
and Liu, Zhiyuan
and Liu, Yang",
title="How to Fine-Tune BERT for Text Classification?",
booktitle="Chinese Computational Linguistics",
year="2019",
publisher="Springer International Publishing",
address="Cham",
pages="194--206",
isbn="978-3-030-32381-3"
}

@article{Argamon2019,
  title = {Register in Computational Language Research},
  author = {Argamon, Shlomo Engelson},
  year = {2019},
  month = apr,
  journal = {Register Studies},
  volume = {1},
  number = {1},
  pages = {100--135},
  publisher = {John Benjamins},
  issn = {2542-9477, 2542-9485},
  doi = {10.1075/rs.18015.arg},
  url = {https://www.jbe-platform.com/content/journals/10.1075/rs.18015.arg},
  urldate = {2024-05-31},
  langid = {english}
}

@inproceedings{kessler-etal-1997-automatic,
    title = "Automatic Detection of Text Genre",
    author = "Kessler, Brett  and
      Nunberg, Geoffrey  and
      Schutze, Hinrich",
    booktitle = "35th Annual Meeting of the Association for Computational Linguistics and 8th Conference of the {E}uropean Chapter of the Association for Computational Linguistics",
    month = jul,
    year = "1997",
    address = "Madrid, Spain",
    publisher = "Association for Computational Linguistics",
    url = "https://aclanthology.org/P97-1005",
    doi = "10.3115/976909.979622",
    pages = "32--38",
}

@inproceedings{gehman-etal-2020-realtoxicityprompts,
    title = "{R}eal{T}oxicity{P}rompts: Evaluating Neural Toxic Degeneration in Language Models",
    author = "Gehman, Samuel  and
      Gururangan, Suchin  and
      Sap, Maarten  and
      Choi, Yejin  and
      Smith, Noah A.",
    editor = "Cohn, Trevor  and
      He, Yulan  and
      Liu, Yang",
    booktitle = "Findings of the Association for Computational Linguistics: EMNLP 2020",
    month = nov,
    year = "2020",
    address = "Online",
    publisher = "Association for Computational Linguistics",
    url = "https://aclanthology.org/2020.findings-emnlp.301",
    doi = "10.18653/v1/2020.findings-emnlp.301",
    pages = "3356--3369",
}

@inproceedings{dodge2021documenting,
  title={Documenting Large Webtext Corpora: A Case Study on the Colossal Clean Crawled Corpus},
  author={Jesse Dodge and Ana Marasovic and Gabriel Ilharco and Dirk Groeneveld and Margaret Mitchell and Matt Gardner},
  booktitle={Conference on Empirical Methods in Natural Language Processing},
  year={2021},
  url={https://api.semanticscholar.org/CorpusID:237568724}
}

@inproceedings{
penedo2023refinedweb,
title={The RefinedWeb Dataset for Falcon {LLM}: Outperforming Curated Corpora with Web Data Only},
author={Guilherme Penedo and Quentin Malartic and Daniel Hesslow and Ruxandra Cojocaru and Hamza Alobeidli and Alessandro Cappelli and Baptiste Pannier and Ebtesam Almazrouei and Julien Launay},
booktitle={Thirty-seventh Conference on Neural Information Processing Systems Datasets and Benchmarks Track},
year={2023},
url={https://openreview.net/forum?id=kM5eGcdCzq}
}

@article{Gries.etal2011,
  title = {N-Grams and the Clustering of Registers},
  author = {Gries, Stefan Th. and Newman, John and Shaoul, Cyrus},
  year = {2011},
  journal = {Empirical Language Research Journal},
  volume = {5},
  number = {11}
}

@inproceedings{Santini2005,
  title = {Genres in Formation? {{An}} Exploratory Study of Web Pages Using Cluster Analysis: {{Proceedings}} of the 8th Annual Colloquium for the {{UK}} Special Interest Group for Computational Linguistics ({{CLUK05}})},
  shorttitle = {Genres in Formation?},
  booktitle = {Proceedings of the 8th Annual Colloquium for the {{UK}} Special Interest Group for Computational Linguistics ({{CLUK05}})},
  author = {Santini, Marina},
  year = {2005},
  address = {Manchester}
}

@inproceedings{
hu2022lora,
title={Lo{RA}: Low-Rank Adaptation of Large Language Models},
author={Edward J Hu and yelong shen and Phillip Wallis and Zeyuan Allen-Zhu and Yuanzhi Li and Shean Wang and Lu Wang and Weizhu Chen},
booktitle={International Conference on Learning Representations},
year={2022},
url={https://openreview.net/forum?id=nZeVKeeFYf9}
}

@inproceedings{lin2018focal,
author = {T. Lin and P. Goyal and R. Girshick and K. He and P. Dollar},
booktitle = {2017 IEEE International Conference on Computer Vision (ICCV)},
title = {Focal Loss for Dense Object Detection},
year = {2017},
volume = {},
issn = {2380-7504},
pages = {2999-3007},
doi = {10.1109/ICCV.2017.324},
url = {https://doi.ieeecomputersociety.org/10.1109/ICCV.2017.324},
publisher = {IEEE Computer Society},
address = {Los Alamitos, CA, USA},
month = {oct}
}

@misc{tokpanov2024zyda,
      title={Zyda: A 1.3T Dataset for Open Language Modeling}, 
      author={Yury Tokpanov and Beren Millidge and Paolo Glorioso and Jonathan Pilault and Adam Ibrahim and James Whittington and Quentin Anthony},
      year={2024},
      eprint={2406.01981},
      archivePrefix={arXiv},
      primaryClass={cs.CL}
}

@inproceedings{10.1007/978-3-642-23808-6_10,
  title = {On the Stratification of Multi-Label Data},
  booktitle = {Machine Learning and Knowledge Discovery in Databases},
  author = {Sechidis, Konstantinos and Tsoumakas, Grigorios and Vlahavas, Ioannis},
  editor = {Gunopulos, Dimitrios and Hofmann, Thomas and Malerba, Donato and Vazirgiannis, Michalis},
  year = {2011},
  pages = {145--158},
  publisher = {Springer Berlin Heidelberg},
  address = {Berlin, Heidelberg},
  isbn = {978-3-642-23808-6}
}

@book{biber2019register,
  title = {Register, Genre, and Style},
  author = {Biber, Douglas and Conrad, Susan},
  year = {2019},
  publisher = {Cambridge University Press},
  address = {Cambridge}
}

@incollection{10.1093/oso/9780195083644.003.0015,
  title = {Register: A Review of Empirical Research},
  booktitle = {Sociolinguistic Perspectives on Register},
  author = {Atkinson, Dwight and Biber, Douglas},
  year = {1994},
  month = jan,
  eprint = {https://academic.oup.com/book/0/chapter/422175006/chapter-pdf/52431717/isbn-9780195083644-book-part-15.pdf},
  publisher = {Oxford University Press},
  doi = {10.1093/oso/9780195083644.003.0015},
  url = {https://doi.org/10.1093/oso/9780195083644.003.0015},
  isbn = {978-0-19-508364-4}
}

@article{Madjarov.etal2019,
  title = {Web Genre Classification with Methods for Structured Output Prediction},
  author = {Madjarov, Gjorgji and Vidulin, Vedrana and Dimitrovski, Ivica and Kocev, Dragi},
  year = {2019},
  month = nov,
  journal = {Information Sciences},
  volume = {503},
  pages = {551--573},
  issn = {0020-0255},
  doi = {10.1016/j.ins.2019.07.009},
  url = {https://www.sciencedirect.com/science/article/pii/S0020025516318965},
  urldate = {2024-06-14}
}

@article{santini2009web,
  title={Web genre benchmark under construction},
  author={Santini, Marina and Sharoff, Serge},
  journal={Journal for Language Technology and Computational Linguistics},
  volume={24},
  number={1},
  pages={129--145},
  year={2009}
}

@article{wang2024multilingual,
  title={Multilingual E5 Text Embeddings: A Technical Report},
  author={Wang, Liang and Yang, Nan and Huang, Xiaolong and Yang, Linjun and Majumder, Rangan and Wei, Furu},
  journal={arXiv preprint arXiv:2402.05672},
  year={2024}
}

@article{jiang2024mixtral,
  title={Mixtral of experts},
  author={Jiang, Albert Q and Sablayrolles, Alexandre and Roux, Antoine and Mensch, Arthur and Savary, Blanche and Bamford, Chris and Chaplot, Devendra Singh and Casas, Diego de las and Hanna, Emma Bou and Bressand, Florian and others},
  journal={arXiv:2401.04088},
  year={2024}
}

@inproceedings{jacob2017quantization,
  author={Jacob, Benoit and Kligys, Skirmantas and Chen, Bo and Zhu, Menglong and Tang, Matthew and Howard, Andrew and Adam, Hartwig and Kalenichenko, Dmitry},
  booktitle={2018 IEEE/CVF Conference on Computer Vision and Pattern Recognition}, 
  title={Quantization and Training of Neural Networks for Efficient Integer-Arithmetic-Only Inference}, 
  year={2018},
  volume={},
  number={},
  pages={2704-2713},
  doi={10.1109/CVPR.2018.00286}}

@inproceedings{dewe-etal-1998-assembling,
    title = "Assembling a Balanced Corpus from the {I}nternet",
    author = "Dewe, Johan  and
      Karlgren, Jussi  and
      Bretan, Ivan",
    editor = "Maegaard, Bente",
    booktitle = "Proceedings of the 11th Nordic Conference of Computational Linguistics ({NODALIDA} 1998)",
    month = mar,
    year = "1998",
    address = "Copenhagen, Denmark",
    publisher = "Center for Sprogteknologi, University of Copenhagen, Denmark",
    url = "https://aclanthology.org/W98-1611",
    pages = "100--108",
}

@incollection{MakingtheWebMoreUsefulasaSourceforLinguisticCorpora,
  title = {Making the Web More Useful as a Source for Linguistic Corpora},
  author = {Fletcher, William H.},
  year = {2004},
  pages = {191--205},
  publisher = {Brill},
  booktitle = {Applied Corpus Linguistics},
  address = {Leiden, The Netherlands},
  doi = {10.1163/9789004333772_011},
  url = {https://brill.com/view/book/9789004333772/B9789004333772-s011.xml},
  isbn = {978-90-04-33377-2}
}

@incollection{FromtheBNCtowardtheCybercorpusAQuantumLeapintoChaos,
  title = {From the {{BNC}} toward the Cybercorpus: A Quantum Leap into Chaos?},
  author = {Brekke, Magnar},
  year = {2000},
  booktitle= {Corpora Galore},
  pages = {227--247},
  publisher = {Brill},
  address = {Leiden, The Netherlands},
  doi = {10.1163/9789004485211_019},
  url = {https://brill.com/view/book/9789004485211/B9789004485211_s019.xml},
  isbn = {978-90-04-48521-1}
}

@incollection{Towardsataxonomyofwebregistersandtexttypesamultidimensionalanalysis,
  title = {Towards a Taxonomy of Web Registers and Text Types: A Multi-Dimensional Analysis},
  author = {Biber, Douglas and Kurjian, Jerry},
  year = {2007},
  pages = {109--131},
  booktitle = {Corpus Linguistics and the Web},
  publisher = {Brill},
  address = {Leiden, The Netherlands},
  doi = {10.1163/9789401203791_008},
  url = {https://brill.com/view/book/9789401203791/B9789401203791-s008.xml},
  isbn = {978-94-012-0379-1}
}

@article{muennighoff2022sgpt,
  title={SGPT: GPT Sentence Embeddings for Semantic Search},
  author={Muennighoff, Niklas},
  journal={arXiv preprint arXiv:2202.08904},
  year={2022}
}

@article{song2022learning,
  author={Song, Hwanjun and Kim, Minseok and Park, Dongmin and Shin, Yooju and Lee, Jae-Gil},
  journal={IEEE Transactions on Neural Networks and Learning Systems}, 
  title={Learning From Noisy Labels With Deep Neural Networks: A Survey}, 
  year={2023},
  volume={34},
  number={11},
  pages={8135-8153},
  doi={10.1109/TNNLS.2022.3152527}}

@inproceedings{kohavi1995study, author = {Kohavi, Ron}, title = {A study of cross-validation and bootstrap for accuracy estimation and model selection}, year = {1995}, isbn = {1558603638}, publisher = {Morgan Kaufmann Publishers Inc.}, address = {San Francisco, CA, USA}, booktitle = {Proceedings of the 14th International Joint Conference on Artificial Intelligence - Volume 2}, pages = {1137–1143}, numpages = {7}, location = {Montreal, Quebec, Canada}, series = {IJCAI'95} }

@misc{Atkins.MacLeod2023,
  title = {How to {{Accurately Time CUDA Kernels}} in {{Pytorch}}},
  author = {Atkins, Lawrence and MacLeod, David},
  year = {2023},
  journal = {Speechmatics},
   howpublished= {https://www.speechmatics.com/company/articles-and-news/timing-operations-in-pytorch} }

@article{Erten-Johansson.etalForthcoming,
  title = {Linguistic Variation beyond the {{Indo-European Web}}: {{Analyzing Turkish Web}} Registers in {{TurCORE}}},
  author = {{Erten-Johansson}, Selcen and Skantsi, Valtteri and Pyysalo, Sampo and Laippala, Veronika},
 year = {2024},
pubstate = {forthcoming},
  journal = {Register {{Studies}}}
}

@misc{törnberg2023chatgpt4outperformsexpertscrowd,
      title={ChatGPT-4 Outperforms Experts and Crowd Workers in Annotating Political Twitter Messages with Zero-Shot Learning}, 
      author={Petter Törnberg},
      year={2023},
      eprint={2304.06588},
      archivePrefix={arXiv},
      primaryClass={cs.CL},
      url={https://arxiv.org/abs/2304.06588}, 
}

@misc{kudo2018sentencepiecesimplelanguageindependent,
      title={SentencePiece: A simple and language independent subword tokenizer and detokenizer for Neural Text Processing}, 
      author={Taku Kudo and John Richardson},
      year={2018},
      eprint={1808.06226},
      archivePrefix={arXiv},
      primaryClass={cs.CL},
      url={https://arxiv.org/abs/1808.06226}, 
}

@inproceedings{henriksson-etal-2024-discrete,
    title = "From Discrete to Continuous Classes: A Situational Analysis of Multilingual Web Registers with {LLM} Annotations",
    author = {Henriksson, Erik  and
      Myntti, Amanda  and
      Hellstr{\"o}m, Saara  and
      Erten-Johansson, Selcen  and
      Eskelinen, Anni  and
      Repo, Liina  and
      Laippala, Veronika},
    editor = {H{\"a}m{\"a}l{\"a}inen, Mika  and
      {\"O}hman, Emily  and
      Miyagawa, So  and
      Alnajjar, Khalid  and
      Bizzoni, Yuri},
    booktitle = "Proceedings of the 4th International Conference on Natural Language Processing for Digital Humanities",
    month = nov,
    year = "2024",
    address = "Miami, USA",
    publisher = "Association for Computational Linguistics",
    url = "https://aclanthology.org/2024.nlp4dh-1.30",
    pages = "308--318",
}

@misc{bermejo2024llms,
  title={LLMs Outperform Outsourced Human Coders on Complex Textual Analysis},
  author={Bermejo, Vicente J. and Gago, Andres and Gálvez, Ramiro H. and Harari, Nicolás},
  year={2024},
  month={nov},
  note= {SSRN Working Paper},
  url= {https://ssrn.com/abstract=5020034},
  doi= {10.2139/ssrn.5020034}
}

@article{Gilardi_2023,
   title={{ChatGPT} outperforms crowd workers for text-annotation tasks},
   volume={120},
   ISSN={1091-6490},
   url={http://dx.doi.org/10.1073/pnas.2305016120},
   DOI={10.1073/pnas.2305016120},
   number={30},
   journal={Proceedings of the National Academy of Sciences},
   publisher={Proceedings of the National Academy of Sciences},
   author={Gilardi, Fabrizio and Alizadeh, Meysam and Kubli, Maël},
   year={2023},
   month={jul}
}

@ARTICLE{2025arXiv250310267B,
       author = {{Burchell}, Laurie and {de Gibert}, Ona and {Arefyev}, Nikolay and {Aulamo}, Mikko and {Ba{\~n}{\'o}n}, Marta and {Chen}, Pinzhen and {Fedorova}, Mariia and {Guillou}, Liane and {Haddow}, Barry and {Haji{\v{c}}}, Jan and {Helcl}, Jind{\v{r}}ich and {Henriksson}, Erik and {Klimaszewski}, Mateusz and {Komulainen}, Ville and {Kutuzov}, Andrey and {Kyt{\"o}niemi}, Joona and {Laippala}, Veronika and {M{\ae}hlum}, Petter and {Malik}, Bhavitvya and {Mehryary}, Farrokh and {Mikhailov}, Vladislav and {Moghe}, Nikita and {Myntti}, Amanda and {O'Brien}, Dayy{\'a}n and {Oepen}, Stephan and {Pal}, Proyag and {Piha}, Jousia and {Pyysalo}, Sampo and {Ram{\'\i}rez-S{\'a}nchez}, Gema and {Samuel}, David and {Stepachev}, Pavel and {Tiedemann}, J{\"o}rg and {Vari{\v{s}}}, Du{\v{s}}an and {Vojt{\v{e}}chov{\'a}}, Tereza and {Zaragoza-Bernabeu}, Jaume},
        title = "{An Expanded Massive Multilingual Dataset for High-Performance Language Technologies (HPLT)}",
      journal = {arXiv e-prints},
         year = 2025,
        month = mar,
          eid = {arXiv:2503.10267},
        pages = {arXiv:2503.10267},
          doi = {10.48550/arXiv.2503.10267},
archivePrefix = {arXiv},
       eprint = {2503.10267},
 primaryClass = {cs.CL},
       adsurl = {https://ui.adsabs.harvard.edu/abs/2025arXiv250310267B}
}

@article{openai2024gpt4o,
  title={{GPT-4o} System Card},
  author={OpenAI},
  journal={arXiv preprint arXiv:2410.21276},
  year={2024}
}

@techreport{anthropic2024claude3,
  title={The {Claude} 3 Model Family: {Opus}, {Sonnet}, {Haiku}},
  author={Anthropic},
  year={2024},
  url={https://assets.anthropic.com/m/61e7d27f8c8f5919/original/Claude-3-Model-Card.pdf}
}

@inproceedings{burchell2025HPLT2,
    title = "An Expanded Massive Multilingual Dataset for High-Performance Language Technologies ({HPLT})",
    author = {Burchell, Laurie  and
      de Gibert, Ona  and
      Arefyev, Nikolay  and
      Aulamo, Mikko  and
      Ba{\~n}{\'o}n, Marta  and
      Chen, Pinzhen  and
      Fedorova, Mariia  and
      Guillou, Liane  and
      Haddow, Barry  and
      Haji{\v{c}}, Jan  and
      Helcl, Jind{\v{r}}ich  and
      Henriksson, Erik  and
      Klimaszewski, Mateusz  and
      Komulainen, Ville  and
      Kutuzov, Andrey  and
      Kyt{\"o}niemi, Joona  and
      Laippala, Veronika  and
      M{\ae}hlum, Petter  and
      Malik, Bhavitvya  and
      Mehryary, Farrokh  and
      Mikhailov, Vladislav  and
      Moghe, Nikita  and
      Myntti, Amanda  and
      O{'}Brien, Dayy{\'a}n  and
      Oepen, Stephan  and
      Pal, Proyag  and
      Piha, Jousia  and
      Pyysalo, Sampo  and
      Ram{\'i}rez-S{\'a}nchez, Gema  and
      Samuel, David  and
      Stepachev, Pavel  and
      Tiedemann, J{\"o}rg  and
      Vari{\v{s}}, Du{\v{s}}an  and
      Vojt{\v{e}}chov{\'a}, Tereza  and
      Zaragoza-Bernabeu, Jaume},
    editor = "Che, Wanxiang  and
      Nabende, Joyce  and
      Shutova, Ekaterina  and
      Pilehvar, Mohammad Taher",
    booktitle = "Proceedings of the 63rd Annual Meeting of the Association for Computational Linguistics (Volume 1: Long Papers)",
    month = jul,
    year = "2025",
    address = "Vienna, Austria",
    publisher = "Association for Computational Linguistics",
    url = "https://aclanthology.org/2025.acl-long.854/",
    doi = "10.18653/v1/2025.acl-long.854",
    pages = "17452--17485",
    ISBN = "979-8-89176-251-0"
}
\newpage
\section*{Appendix}
\label{sec:appendix}

In this section, we present a statistical comparison of different model architectures using a paired bootstrap approach to estimate $F1$-score differences between models with confidence intervals at significance level $\alpha = 0.05$ (95\% confidence). Tables \ref{tab:app-monoling}--\ref{tab:transfer_ur-zh} present the results separated by experiment type: monolingual, zero-shot, multilingual and evaluation-only. For each comparison, we report the estimated micro-$F1$ difference ($\Delta F1$) with confidence interval lower and upper limits. For ease of comparison, we also display the $F1$-scores for each model (in parentheses after the model name) as reported in Tables \ref{tab:main_results_core} and \ref{tab:zeroshot_results}.

For each pair of models and each language subset, we perform paired bootstrap resampling by randomly sampling with replacement from the predictions of each model. For each paired sample, we calculate the micro-averaged $F1$-score for each model, and repeat this process for 1000 iterations. We then analyze the resulting distribution of $F1$-values and estimate the $F1$-score difference between the models as the mean difference of $F1$-scores across iterations, and derive confidence intervals as 2.5th and 97.5th percentiles. We assign significance to the result if the confidence interval excludes zero.

\begin{table}[]

\caption{Comparison between models for monolingual experiments using a paired bootstrap test on micro-averaged $F1$-score. Bold values indicate the better-performing model in each comparison; asterisks (*) indicate significant differences at $\alpha = 0.05$.}
\label{tab:app-monoling}
{\tablefont
\setlength{\tabcolsep}{3.5pt}
\renewcommand{\arraystretch}{0.8}
\begin{tabular}{@{}llcllc@{}}
\toprule
\multicolumn{3}{l}{All}                                                              & \multicolumn{3}{l}{Main}                                                              \\ \midrule
Comparison A                    & Comparison B                    & $\Delta F1$, Conf. Interval    & Comparison A                    & Comparison B                    & $\Delta F1$, Conf. Interval    \\ \midrule \midrule
\multicolumn{3}{l}{Monolingual En} \\ \midrule

\textbf{XLM-R}\scriptsize{(0.74)} & BGE-M3\scriptsize{(0.73)} & 0.004*\scriptsize{(0.002, 0.006)} & \textbf{XLM-R}\scriptsize{(0.76)} & BGE-M3\scriptsize{(0.76)} & 0.004*\scriptsize{(0.002, 0.006)} \\
XLM-R\scriptsize{(0.74)} & \textbf{BGE-M3$^{2048}$}\scriptsize{(0.74)} & -0.003*\scriptsize{(-0.005, -0.000)} & XLM-R\scriptsize{(0.76)} & \textbf{BGE-M3$^{2048}$}\scriptsize{(0.76)} & -0.002\scriptsize{(-0.005, 0.000)} \\
BGE-M3\scriptsize{(0.73)} & \textbf{BGE-M3$^{2048}$}\scriptsize{(0.74)} & -0.007*\scriptsize{(-0.009, -0.005)} & BGE-M3\scriptsize{(0.76)} & \textbf{BGE-M3$^{2048}$}\scriptsize{(0.76)} & -0.006*\scriptsize{(-0.009, -0.004)} \\

 \midrule
\multicolumn{3}{l}{Monolingual Fi} \\ \midrule

\textbf{XLM-R}\scriptsize{(0.79)} & BGE-M3\scriptsize{(0.79)} & 0.005*\scriptsize{(0.000, 0.010)} & \textbf{XLM-R}\scriptsize{(0.82)} & BGE-M3\scriptsize{(0.81)} & 0.006*\scriptsize{(0.001, 0.011)} \\
\bottomrule

\end{tabular}
}

\end{table}

\begin{table}[]

\caption{Comparison between models for zero-shot experiments using a paired bootstrap test on micro-averaged $F1$-score. Bold values indicate the better-performing model in each comparison; asterisks (*) indicate significant differences at $\alpha = 0.05$.}
\label{tab:app-zeroshot}
{\tablefont
\setlength{\tabcolsep}{3.5pt}
\renewcommand{\arraystretch}{0.8}
\begin{tabular}{@{}llcllc@{}}
\toprule
\multicolumn{3}{l}{All}                                                              & \multicolumn{3}{l}{Main}                                                              \\ \midrule
Comparison A                    & Comparison B                    & $\Delta F1$, Conf. Interval    & Comparison A                    & Comparison B                    & $\Delta F1$, Conf. Interval    \\ \midrule \midrule

\multicolumn{3}{l}{Zero-shot En} \\ \midrule

\textbf{XLM-R}\scriptsize{(0.64)} & BGE-M3\scriptsize{(0.62)} & 0.017*\scriptsize{(0.014, 0.020)} & \textbf{XLM-R}\scriptsize{(0.67)} & BGE-M3\scriptsize{(0.66)} & 0.014*\scriptsize{(0.011, 0.017)} \\
\textbf{XLM-R}\scriptsize{(0.64)} & BGE-M3$^{2048}$\scriptsize{(0.63)} & 0.010*\scriptsize{(0.008, 0.013)} & \textbf{XLM-R}\scriptsize{(0.67)} & BGE-M3$^{2048}$\scriptsize{(0.66)} & 0.011*\scriptsize{(0.008, 0.014)} \\
BGE-M3\scriptsize{(0.62)} & \textbf{BGE-M3$^{2048}$}\scriptsize{(0.63)} & -0.007*\scriptsize{(-0.009, -0.004)} & BGE-M3\scriptsize{(0.66)} & \textbf{BGE-M3$^{2048}$}\scriptsize{(0.66)} & -0.003*\scriptsize{(-0.005, -0.000)} \\

 \midrule
\multicolumn{3}{l}{Zero-shot Fi} \\ \midrule

XLM-R\scriptsize{(0.72)} & \textbf{BGE-M3}\scriptsize{(0.72)} & -0.003\scriptsize{(-0.010, 0.004)} & XLM-R\scriptsize{(0.73)} & \textbf{BGE-M3}\scriptsize{(0.74)} & -0.015*\scriptsize{(-0.022, -0.008)} \\
\bottomrule

\end{tabular}
}

\end{table}

\begin{table}[]
\caption{Comparison between models for multilingual experiments on English, Finnish and French using a paired bootstrap test on micro-averaged $F1$-score. Bold values indicate the better-performing model in each comparison; asterisks (*) indicate significant differences at $\alpha = 0.05$.}
\label{tab:multi-en-fi-fr}
{\tablefont
\setlength{\tabcolsep}{3.5pt}
\renewcommand{\arraystretch}{0.8}

\begin{tabular}{@{}llcllc@{}}
\toprule
\multicolumn{3}{l}{All}                                                              & \multicolumn{3}{l}{Main}                                                              \\ \midrule
Comparison A                    & Comparison B                    & $\Delta F1$, Conf. Interval    & Comparison A                    & Comparison B                    & $\Delta F1$, Conf. Interval    \\ \midrule 
 \midrule
\multicolumn{3}{l}{Multilingual En} \\ \midrule

\textbf{XLM-R}\scriptsize{(0.72)} & BGE-M3\scriptsize{(0.71)} & 0.010*\scriptsize{(0.005, 0.015)} & \textbf{XLM-R}\scriptsize{(0.75)} & BGE-M3\scriptsize{(0.74)} & 0.008*\scriptsize{(0.006, 0.011)} \\
\textbf{XLM-R}\scriptsize{(0.72)} & BGE-M3$^{2048}$\scriptsize{(0.72)} & 0.002\scriptsize{(-0.002, 0.007)} & \textbf{XLM-R}\scriptsize{(0.75)} & BGE-M3$^{2048}$\scriptsize{(0.74)} & 0.002\scriptsize{(-0.000, 0.005)} \\
\textbf{XLM-R}\scriptsize{(0.72)} & ME5-L\scriptsize{(0.71)} & 0.005*\scriptsize{(0.000, 0.010)} & \textbf{XLM-R}\scriptsize{(0.75)} & ME5-L\scriptsize{(0.74)} & 0.004*\scriptsize{(0.002, 0.007)} \\
\textbf{XLM-R}\scriptsize{(0.72)} & XLMR-XL\scriptsize{(0.71)} & 0.006*\scriptsize{(0.001, 0.010)} & \textbf{XLM-R}\scriptsize{(0.75)} & XLMR-XL\scriptsize{(0.74)} & 0.003*\scriptsize{(0.000, 0.005)} \\
\textbf{XLM-R}\scriptsize{(0.72)} & Mixtral-8x7B\scriptsize{(0.70)} & 0.019*\scriptsize{(0.014, 0.024)} & \textbf{XLM-R}\scriptsize{(0.75)} & Mixtral-8x7B\scriptsize{(0.73)} & 0.018*\scriptsize{(0.013, 0.023)} \\
\bottomrule
\end{tabular}
}
\end{table}

\begin{table}[]
\caption{Comparison between models for multilingual experiments on Swedish and Turkish using a paired bootstrap test on micro-averaged $F1$-score. Bold values indicate the better-performing model in each comparison; asterisks (*) indicate significant differences at $\alpha = 0.05$.}
\label{tab:multi-sv-tr}
{\tablefont
\setlength{\tabcolsep}{3.5pt}
\renewcommand{\arraystretch}{0.8}

\begin{tabular}{@{}llcllc@{}}
\toprule
\multicolumn{3}{l}{All}                                                              & \multicolumn{3}{l}{Main}                                                              \\ \midrule
Comparison A                    & Comparison B                    & $\Delta F1$, Conf. Interval    & Comparison A                    & Comparison B                    & $\Delta F1$, Conf. Interval    \\ \midrule 
 \midrule
\multicolumn{3}{l}{Multilingual Sv} \\ \midrule

\textbf{XLM-R}\scriptsize{(0.81)} & BGE-M3\scriptsize{(0.80)} & 0.008\scriptsize{(-0.002, 0.019)} & \textbf{XLM-R}\scriptsize{(0.82)} & BGE-M3\scriptsize{(0.81)} & 0.007*\scriptsize{(0.000, 0.014)} \\
XLM-R\scriptsize{(0.81)} & \textbf{BGE-M3$^{2048}$}\scriptsize{(0.81)} & -0.002\scriptsize{(-0.012, 0.007)} & XLM-R\scriptsize{(0.82)} & \textbf{BGE-M3$^{2048}$}\scriptsize{(0.82)} & -0.002\scriptsize{(-0.008, 0.005)} \\
\textbf{XLM-R}\scriptsize{(0.81)} & ME5-L\scriptsize{(0.80)} & 0.006\scriptsize{(-0.005, 0.016)} & \textbf{XLM-R}\scriptsize{(0.82)} & ME5-L\scriptsize{(0.82)} & 0.005\scriptsize{(-0.002, 0.011)} \\
XLM-R\scriptsize{(0.81)} & \textbf{XLMR-XL}\scriptsize{(0.81)} & -0.005\scriptsize{(-0.016, 0.006)} & XLM-R\scriptsize{(0.82)} & \textbf{XLMR-XL}\scriptsize{(0.82)} & -0.003\scriptsize{(-0.010, 0.003)} \\
\textbf{XLM-R}\scriptsize{(0.81)} & Mixtral-8x7B\scriptsize{(0.78)} & 0.024*\scriptsize{(0.014, 0.035)} & \textbf{XLM-R}\scriptsize{(0.82)} & Mixtral-8x7B\scriptsize{(0.80)} & 0.025*\scriptsize{(0.011, 0.037)} \\
\bottomrule
\end{tabular}
}
\end{table}

\begin{table}[]
\caption{Comparison between models for evaluation-only experiments on Arabic, Catalan and Farsi using a paired bootstrap test on micro-averaged $F1$-score. Bold values indicate the better-performing model in each comparison; asterisks (*) indicate significant differences at $\alpha = 0.05$.}
\label{tab:transfer_ar-ca-fa}
{\tablefont
\setlength{\tabcolsep}{3.5pt}
\renewcommand{\arraystretch}{0.8}

\begin{tabular}{@{}llcllc@{}}
\toprule
\multicolumn{3}{l}{All}                                                              & \multicolumn{3}{l}{Main}                                                              \\ \midrule
Comparison A                    & Comparison B                    & $\Delta F1$, Conf. Interval    & Comparison A                    & Comparison B                    & $\Delta F1$, Conf. Interval    \\ \midrule \midrule
\multicolumn{3}{l}{Evaluation-only Ar} \\ \midrule

XLM-R\scriptsize{(0.64)} & \textbf{BGE-M3}\scriptsize{(0.64)} & -0.004\scriptsize{(-0.025, 0.015)} & XLM-R\scriptsize{(0.66)} & \textbf{BGE-M3}\scriptsize{(0.67)} & -0.007\scriptsize{(-0.035, 0.021)} \\
XLM-R\scriptsize{(0.64)} & \textbf{BGE-M3$^{2048}$}\scriptsize{(0.65)} & -0.011\scriptsize{(-0.033, 0.010)} & XLM-R\scriptsize{(0.66)} & \textbf{BGE-M3$^{2048}$}\scriptsize{(0.68)} & -0.015\scriptsize{(-0.044, 0.013)} \\
XLM-R\scriptsize{(0.64)} & \textbf{ME5-L}\scriptsize{(0.65)} & -0.009\scriptsize{(-0.035, 0.015)} & XLM-R\scriptsize{(0.66)} & \textbf{ME5-L}\scriptsize{(0.68)} & -0.021\scriptsize{(-0.049, 0.005)} \\
XLM-R\scriptsize{(0.64)} & \textbf{XLMR-XL}\scriptsize{(0.64)} & -0.008\scriptsize{(-0.030, 0.013)} & XLM-R\scriptsize{(0.66)} & \textbf{XLMR-XL}\scriptsize{(0.68)} & -0.019\scriptsize{(-0.047, 0.008)} \\
\textbf{XLM-R}\scriptsize{(0.64)} & Mixtral-8x7B\scriptsize{(0.59)} & 0.049*\scriptsize{(0.018, 0.081)} & \textbf{XLM-R}\scriptsize{(0.66)} & Mixtral-8x7B\scriptsize{(0.60)} & 0.056*\scriptsize{(0.019, 0.092)} \\
\bottomrule
\end{tabular}
}
\end{table}

\begin{table}[]
\caption{Comparison between models for evaluation-only experiments on Spanish, Hindi and Indonesian using a paired bootstrap test on micro-averaged $F1$-score. Bold values indicate the better-performing model in each comparison; asterisks (*) indicate significant differences at $\alpha = 0.05$.}
\label{tab:transfer_es-hi-id}
{\tablefont
\setlength{\tabcolsep}{3.5pt}
\renewcommand{\arraystretch}{0.8}

\begin{tabular}{@{}llcllc@{}}
\toprule
\multicolumn{3}{l}{All}                                                              & \multicolumn{3}{l}{Main}                                                              \\ \midrule
Comparison A                    & Comparison B                    & $\Delta F1$, Conf. Interval    & Comparison A                    & Comparison B                    & $\Delta F1$, Conf. Interval    \\ \midrule \midrule
\multicolumn{3}{l}{Evaluation-only Es} \\ \midrule

XLM-R\scriptsize{(0.61)} & \textbf{BGE-M3}\scriptsize{(0.62)} & -0.014\scriptsize{(-0.039, 0.013)} & XLM-R\scriptsize{(0.67)} & \textbf{BGE-M3}\scriptsize{(0.69)} & -0.017\scriptsize{(-0.044, 0.010)} \\
XLM-R\scriptsize{(0.61)} & \textbf{BGE-M3$^{2048}$}\scriptsize{(0.62)} & -0.009\scriptsize{(-0.042, 0.024)} & XLM-R\scriptsize{(0.67)} & \textbf{BGE-M3$^{2048}$}\scriptsize{(0.69)} & -0.013\scriptsize{(-0.042, 0.016)} \\
\textbf{XLM-R}\scriptsize{(0.61)} & ME5-L\scriptsize{(0.60)} & 0.007\scriptsize{(-0.018, 0.032)} & \textbf{XLM-R}\scriptsize{(0.67)} & ME5-L\scriptsize{(0.67)} & 0.001\scriptsize{(-0.033, 0.034)} \\
XLM-R\scriptsize{(0.61)} & \textbf{XLMR-XL}\scriptsize{(0.62)} & -0.007\scriptsize{(-0.032, 0.017)} & XLM-R\scriptsize{(0.67)} & \textbf{XLMR-XL}\scriptsize{(0.69)} & -0.012\scriptsize{(-0.039, 0.016)} \\
\textbf{XLM-R}\scriptsize{(0.61)} & Mixtral-8x7B\scriptsize{(0.61)} & 0.001\scriptsize{(-0.031, 0.033)} & \textbf{XLM-R}\scriptsize{(0.67)} & Mixtral-8x7B\scriptsize{(0.67)} & 0.001\scriptsize{(-0.032, 0.033)} \\
\bottomrule
\end{tabular}
}
\end{table}

\begin{table}[]
\caption{Comparison between models for evaluation-only experiments on Japanese, Norwegian and Portuguese using a paired bootstrap test on micro-averaged $F1$-score. Bold values indicate the better-performing model in each comparison; asterisks (*) indicate significant differences at $\alpha = 0.05$.}
\label{tab:transfer_jp-no-pt}
{\tablefont
\setlength{\tabcolsep}{3.5pt}
\renewcommand{\arraystretch}{0.8}

\begin{tabular}{@{}llcllc@{}}
\toprule
\multicolumn{3}{l}{All}                                                              & \multicolumn{3}{l}{Main}                                                              \\ \midrule
Comparison A                    & Comparison B                    & $\Delta F1$, Conf. Interval    & Comparison A                    & Comparison B                    & $\Delta F1$, Conf. Interval    \\ \midrule \midrule
\multicolumn{3}{l}{Evaluation-only Jp} \\ \midrule

\textbf{XLM-R}\scriptsize{(0.51)} & BGE-M3\scriptsize{(0.50)} & 0.008\scriptsize{(-0.023, 0.035)} & XLM-R\scriptsize{(0.61)} & \textbf{BGE-M3}\scriptsize{(0.62)} & -0.005\scriptsize{(-0.037, 0.025)} \\
XLM-R\scriptsize{(0.51)} & \textbf{BGE-M3$^{2048}$}\scriptsize{(0.54)} & -0.035*\scriptsize{(-0.068, -0.005)} & XLM-R\scriptsize{(0.61)} & \textbf{BGE-M3$^{2048}$}\scriptsize{(0.66)} & -0.051*\scriptsize{(-0.084, -0.017)} \\
XLM-R\scriptsize{(0.51)} & \textbf{ME5-L}\scriptsize{(0.51)} & -0.000\scriptsize{(-0.032, 0.031)} & XLM-R\scriptsize{(0.61)} & \textbf{ME5-L}\scriptsize{(0.63)} & -0.014\scriptsize{(-0.053, 0.022)} \\
XLM-R\scriptsize{(0.51)} & \textbf{XLMR-XL}\scriptsize{(0.55)} & -0.038*\scriptsize{(-0.071, -0.003)} & XLM-R\scriptsize{(0.61)} & \textbf{XLMR-XL}\scriptsize{(0.66)} & -0.052*\scriptsize{(-0.087, -0.015)} \\
XLM-R\scriptsize{(0.51)} & \textbf{Mixtral-8x7B}\scriptsize{(0.54)} & -0.023\scriptsize{(-0.059, 0.016)} & XLM-R\scriptsize{(0.61)} & \textbf{Mixtral-8x7B}\scriptsize{(0.64)} & -0.022\scriptsize{(-0.066, 0.022)} \\
\bottomrule
\end{tabular}
}
\end{table}

\begin{table}[]
\caption{Comparison between models for evaluation-only experiments on Urdu and Chinese using a paired bootstrap test on micro-averaged $F1$-score. Bold values indicate the better-performing model in each comparison; asterisks (*) indicate significant differences at $\alpha = 0.05$.}
\label{tab:transfer_ur-zh}
{\tablefont
\setlength{\tabcolsep}{3.5pt}
\renewcommand{\arraystretch}{0.8}

\begin{tabular}{@{}llcllc@{}}
\toprule
\multicolumn{3}{l}{All}                                                              & \multicolumn{3}{l}{Main}                                                              \\ \midrule
Comparison A                    & Comparison B                    & $\Delta F1$, Conf. Interval    & Comparison A                    & Comparison B                    & $\Delta F1$, Conf. Interval    \\ \midrule \midrule
\multicolumn{3}{l}{Evaluation-only Ur} \\ \midrule

\textbf{XLM-R}\scriptsize{(0.82)} & BGE-M3\scriptsize{(0.80)} & 0.019*\scriptsize{(0.006, 0.034)} & \textbf{XLM-R}\scriptsize{(0.84)} & BGE-M3\scriptsize{(0.82)} & 0.020*\scriptsize{(0.002, 0.038)} \\
\textbf{XLM-R}\scriptsize{(0.82)} & BGE-M3$^{2048}$\scriptsize{(0.81)} & 0.011\scriptsize{(-0.005, 0.025)} & \textbf{XLM-R}\scriptsize{(0.84)} & BGE-M3$^{2048}$\scriptsize{(0.82)} & 0.012\scriptsize{(-0.008, 0.030)} \\
XLM-R\scriptsize{(0.82)} & \textbf{ME5-L}\scriptsize{(0.83)} & -0.007\scriptsize{(-0.024, 0.010)} & XLM-R\scriptsize{(0.84)} & \textbf{ME5-L}\scriptsize{(0.84)} & -0.008\scriptsize{(-0.026, 0.011)} \\
\textbf{XLM-R}\scriptsize{(0.82)} & XLMR-XL\scriptsize{(0.81)} & 0.015*\scriptsize{(0.001, 0.028)} & \textbf{XLM-R}\scriptsize{(0.84)} & XLMR-XL\scriptsize{(0.81)} & 0.023*\scriptsize{(0.007, 0.040)} \\
\textbf{XLM-R}\scriptsize{(0.82)} & Mixtral-8x7B\scriptsize{(0.76)} & 0.062*\scriptsize{(0.035, 0.088)} & \textbf{XLM-R}\scriptsize{(0.84)} & Mixtral-8x7B\scriptsize{(0.76)} & 0.071*\scriptsize{(0.042, 0.101)} \\
\bottomrule
\end{tabular}
}
\end{table}

\label{lastpage}
\end{document}